%% file: main.tex
\documentclass{article}

\usepackage{vlrs_preprint}

\usepackage[utf8]{inputenc}
\usepackage[T1]{fontenc}
\usepackage{hyperref}
\usepackage{url}
\usepackage{microtype}
\usepackage{graphicx}
\usepackage{subcaption}
\usepackage{booktabs}
\usepackage{amsmath}
\usepackage{amssymb}
\usepackage{amsfonts}
\usepackage{mathtools}
\usepackage{amsthm}
\usepackage{longtable}
\usepackage{multirow}
\usepackage{pifont}
\usepackage[table]{xcolor}
\usepackage{float}
\usepackage{placeins}
\usepackage{xspace}
\usepackage{caption}
\usepackage[capitalize,noabbrev]{cleveref}
\usepackage[textsize=tiny]{todonotes}

\newcommand{\xmark}{\ding{55}}
\newcommand{\eg}{e.g.,\xspace}

\newcommand{\etc}{etc.\xspace}

\theoremstyle{plain}

\theoremstyle{definition}

\theoremstyle{remark}

\definecolor{wdcolor}{RGB}{128, 0, 255}

\definecolor{wdqcolor}{RGB}{255, 0, 0}

\title{%
  \texorpdfstring{%
    \raisebox{-0.21\height}[0pt][0pt]{\includegraphics[height=1.15em]{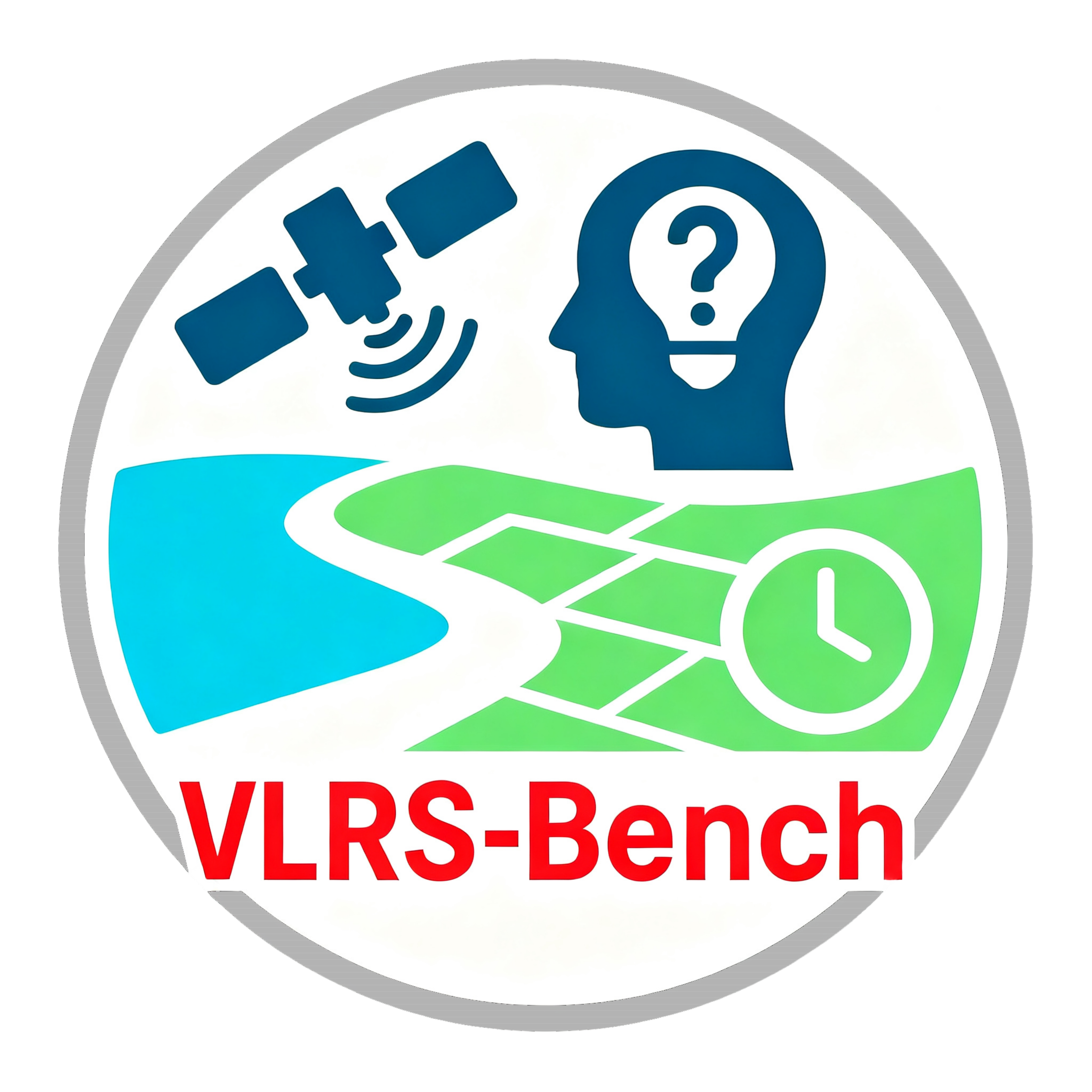}}%
    \hspace{0.27em}%
  }{}%
  VLRS-Bench: A Vision-Language Reasoning Benchmark for Remote Sensing%
}

\author{%
  Zhiming Luo$^{1,2}$, Hebaixu Wang$^{2,3}$, Haonan Guo$^{1,2\dagger}$, Jing Zhang$^{1,2\dagger}$, Bo Du$^{1,2\dagger}$ \\
  $^{1}$School of Computer Science, Wuhan University \\
  $^{2}$Zhongguancun Academy \\
  $^{3}$School of Electronic Information, Wuhan University, Wuhan, China \\
  {\small \href{mailto:thislzm@whu.edu.cn}{thislzm@whu.edu.cn}\quad \href{mailto:wanghebaixu@gmail.com}{wanghebaixu@gmail.com}\quad \href{mailto:haonan.guo@whu.edu.cn}{haonan.guo@whu.edu.cn}} \\
  {\small \href{mailto:jingzhang.cv@gmail.com}{jingzhang.cv@gmail.com}\quad \href{mailto:dubo@whu.edu.cn}{dubo@whu.edu.cn}}
}

\hypersetup{pdftitle={VLRS-Bench: A Vision-Language Reasoning Benchmark for Remote Sensing},pdfauthor={Zhiming Luo, Hebaixu Wang, Haonan Guo, Jing Zhang, Bo Du}}

\begin{document}

\maketitle
\begingroup
\renewcommand{\thefootnote}{\fnsymbol{footnote}}
\footnotetext[2]{Corresponding authors.}
\endgroup

\vspace{-0.9cm}
\begin{center}
  \centering
  \includegraphics[width=\linewidth]{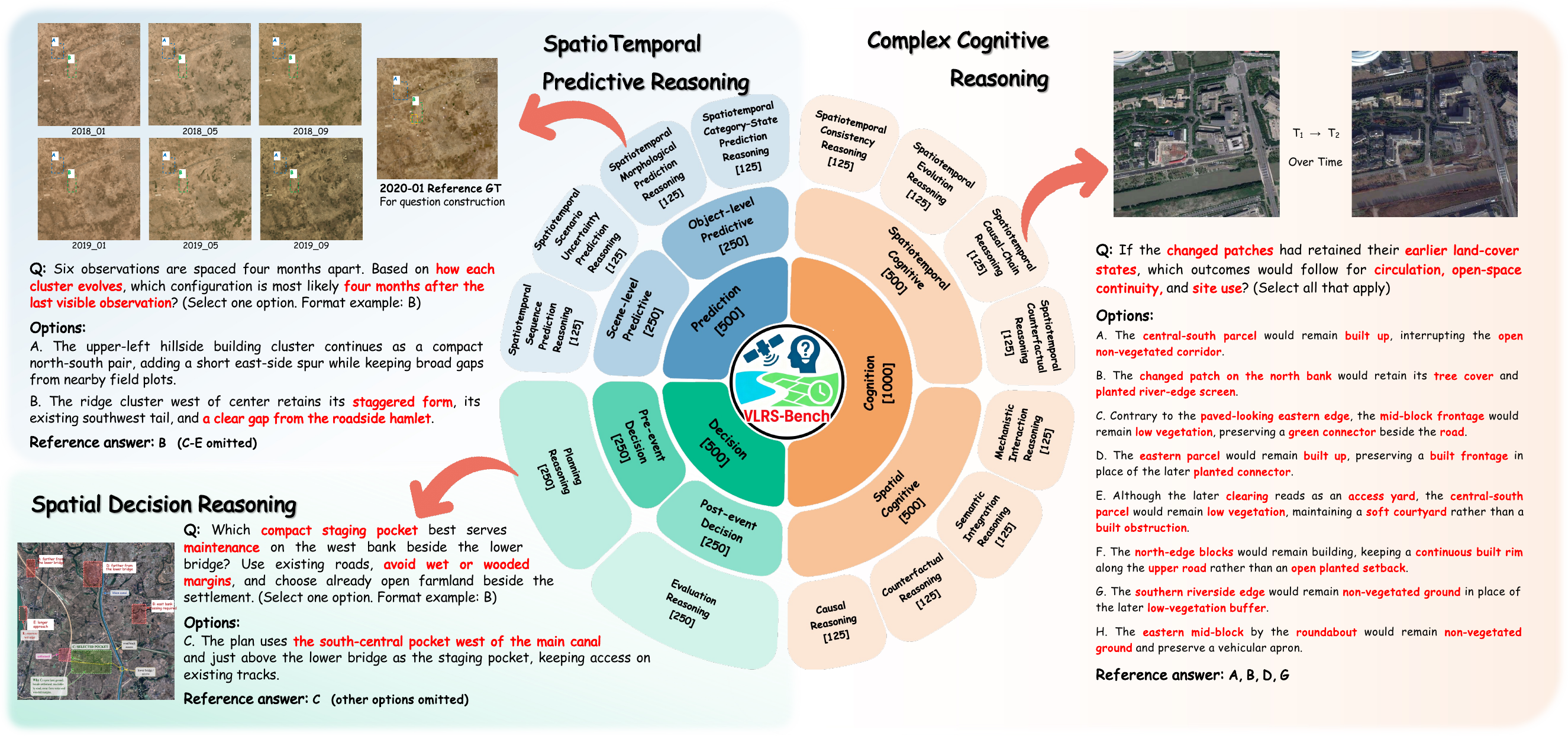}
  \captionof{figure}{Illustrative cases from VLRS-Bench. The benchmark comprises three core reasoning dimensions (Cognition Reasoning: \textbf{\textit{Why is this}}; Decision Reasoning: \textbf{\textit{How to do}}; Prediction Reasoning: \textbf{\textit{What will happen}}), further decomposed into six specific abilities and fourteen fine-grained tasks for comprehensive evaluation of MLLM reasoning.}
  \label{fig:intro}
\end{center}

\begin{abstract}
Recent advancements in Multimodal Large Language Models (MLLMs) have enabled complex reasoning. However, existing remote sensing (RS) benchmarks remain heavily biased toward perception tasks, such as object recognition and scene classification. This limitation hinders the development of MLLMs for cognitively demanding RS applications. To address this, we propose a \textbf{V}ision \textbf{L}anguage \textbf{R}ea\textbf{S}oning Benchmark (VLRS-Bench), which is the first benchmark exclusively dedicated to complex RS reasoning. Structured across the three core dimensions of Cognition, Decision, and Prediction, VLRS-Bench comprises 2,000 question-answer pairs with an average question length of 129.87 words, spanning 14 tasks and up to eight temporal phases. VLRS-Bench is constructed via a specialized pipeline that integrates RS-specific priors and expert knowledge to ensure geospatial realism and reasoning complexity. Experimental results reveal significant bottlenecks in existing state-of-the-art MLLMs, providing critical insights for advancing multimodal reasoning within the remote sensing community.

\end{abstract}

\vspace{-0.4cm}
\section{Introduction}
\label{sec:intro}

Multimodal Large Language Models (MLLMs) \cite{wang2025geollava,hu2025rsgpt,luo2024skysensegpt} have substantially advanced visual understanding and complex reasoning by integrating powerful visual encoders with large language models. There is now broad agreement in the research community that well-designed benchmarks are essential for guiding model development and advancing model capabilities. This consensus has motivated extensive benchmarking efforts across diverse domains, spanning computer vision and medical diagnosis \cite{medical1, medical2}, autonomous driving \cite{drive1, drive2}, and emerging areas such as remote sensing \cite{rs1, rs2}.

Remote sensing imagery presents a unique evaluation landscape for MLLMs, characterized by intricate spatial dependencies and long-term dynamic evolution. Previous benchmarks \cite{rsvqa, zhan2023rsvg} have predominantly focused on perception tasks, such as object recognition and basic relational understanding. However, the fundamental scientific value of remote sensing lies in deciphering causal mechanisms and evolutionary patterns. Comprehending these dynamics demands complex reasoning capabilities that transcend the scope of perception-oriented frameworks.
 
Although recent benchmarks such as CHOICE \cite{choice} have incorporated several reasoning dimensions, they remain largely perception- or format-driven, with limited temporal depth and restricted use of RS-specific priors, as summarized in Table~\ref{tab:benchmark_comparison}. They therefore still exhibit notable limitations in evaluating genuine reasoning ability: \textbf{(1) Reasoning tasks lack a clear classification system and hierarchical structure.}
Most existing benchmarks \cite{rsvqa, zhan2023rsvg} define task categories according to application scenarios (\eg environmental assessment or path planning) or task formats (\eg attribute judgment or area counting), resulting in task-driven rather than cognition-driven designs. Such categorization lacks a coherent cognitive progression from basic to advanced abilities, making it difficult to systematically assess whether a model possesses the underlying cognitive skills required for higher-level reasoning.
\textbf{(2) Reasoning forms are limited to simple relational or descriptive tasks.}
Previous benchmarks \cite{hu2025rsgpt, LHRS_Bot} have focused primarily on perception tasks such as object recognition and basic relational understanding, which assess only interpretative abilities. Yet remote sensing imagery, with its complex semantics and explicit spatiotemporal structure, demands complex reasoning capable of uncovering causal mechanisms and evolutionary patterns.
\textbf{(3) Current benchmarks underutilize remote sensing–specific priors.}
Existing benchmarks \cite{wang2025xlrs, wang2024earthvqa} primarily rely on standard RGB imagery and largely overlook remote sensing–specific priors such as elevation data (DSM), non-visible spectra (NIR), and expert annotations (masks, bounding boxes). Derived from specialized sensors and expert analysis, these multi-modal data are crucial for designing realistic and complex reasoning tasks. Their integration would enable benchmarks to support more challenging, authentic reasoning and yield more robust evaluations of remote sensing MLLMs.

To address these limitations, we introduce VLRS-Bench, the first benchmark designed to systematically evaluate complex reasoning in remote sensing. As shown in Figure~\ref{fig:intro}, inspired by neuroscientific research \cite{neuroscience,Neuroscience1}, we structure VLRS-Bench around three L-1 dimensions: (1) Cognition Reasoning (\textbf{\textit{Why is this}}), representing causal understanding; (2) Decision Reasoning (\textbf{\textit{How to do}}), representing strategic execution; and (3) Prediction Reasoning (\textbf{\textit{What will happen}}), temporal extrapolation of geospatial states. These dimensions are further organized into six L-2 specific abilities and fourteen L-3 tasks. To construct VLRS-Bench, we develop a highly automated pipeline that explicitly integrates multi-modal priors, such as DSM and NIR imagery, and RS expert pixel-level annotations to generate evaluation scenarios grounded in authentic physical constraints and complex geospatial logic. Extensive experiments validate the rigor of this design, demonstrating that VLRS-Bench poses significant challenges to current MLLMs while effectively highlighting the distinctive reasoning demands of the remote sensing domain. In summary, our main contributions are as follows:

\begin{itemize}
\vspace{-0.2cm}

\item We propose VLRS-Bench, the first benchmark tailored for remote sensing multimodal reasoning tasks. It systematically decomposes complex reasoning into three core dimensions: Cognition, Decision, and Prediction, encompassing 14 fine-grained tasks, which facilitates a comprehensive and holistic quantitative assessment of MLLMs' reasoning capabilities.
\vspace{-0.1cm}
\item We introduce a highly automated pipeline for constructing VLRS-Bench. It explicitly leverages RS-specific priors such as DSM, NIR imagery, and expert-annotated pixel-level masks to generate evaluation tasks with enhanced geospatial realism and reasoning complexity.
\vspace{-0.1cm}
\item Experimental results demonstrate that generalist MLLMs exhibit significant deficiencies in geospatial reasoning. Although RS-specific MLLMs achieve higher average scores, they still face critical limitations in complex decision-making and prediction tasks, highlighting the challenging nature of VLRS-Bench and the need for further architectural advances.

\end{itemize}

\begin{table*}[!t]
\centering
\footnotesize
\vspace{-0.2cm}
\caption{Comparison of multimodal benchmarks in general and RS domains.}
\vspace{-0.2cm}
\label{tab:benchmark_comparison}

\renewcommand{\arraystretch}{1.2}
\resizebox{\linewidth}{!}{
\begin{tabular}{lcccccc}
\toprule
\textbf{Benchmark} & \textbf{Data Source} & \textbf{Avg. Q Len.} & \textbf{Type} & \textbf{Reason. Dim.} & \textbf{Temporals} & \textbf{RS Priors} \\
\midrule
\rowcolor{orange!25}\multicolumn{7}{l}{\textit{\textbf{General Benchmarks}}} \\
MMBench \cite{liu2024mmbench} & 10 Public Datasets & \xmark & MCQ & 8 & \xmark & \xmark \\
MMStar \cite{chen2024we} & 6 Public Benchmarks & \xmark & MCQ & 6 & \xmark & \xmark \\
SEED-Bench-2 \cite{Li_2024_CVPR} & 5 Public Datasets & \xmark & MCQ & 7 & \xmark & \xmark \\
\midrule
\rowcolor{orange!25}\multicolumn{7}{l}{\textit{\textbf{Remote Sensing Benchmarks}}} \\
LEVIR-CC \cite{Liu_2024ca} & LEVIR-CD & 7.99 & FF & \xmark & 2 & \xmark \\
RSVGD \cite{zhan2023rsvg} & DIOR Dataset & 7.47 & FF & \xmark & \xmark & BBox \\
RSVQA \cite{rsvqa} & HR \& LR Datasets & \xmark & FF & \xmark & \xmark & \xmark \\
RSIEval \cite{hu2025rsgpt} & DOTA Val. Dataset & \xmark & FF & 1 & \xmark & BBox \\
LHRS-Bench \cite{LHRS_Bot} & Google Earth & \xmark & MCQ & 1 & \xmark & \xmark \\
GeoChat \cite{kuckreja2024geochat} & SAMRS, LRBEN & \xmark & FF & \xmark & \xmark & BBox, Mask \\
EarthVQA \cite{wang2024earthvqa} & LoveDA Dataset & \xmark & FF & 3 & \xmark & Mask \\
VRSBench \cite{li2024vrsbench} & DOTA-v2, DIOR & 52 & FF & 1 & \xmark & BBox \\
RSVLM-QA \cite{RSVLM-QA} & WHU, LoveDA, iSAID & 9.23 & FF & 2 & \xmark & Mask \\
GEOBench-VLM \cite{Geobench-vlm} & 8 Public Datasets & \xmark & MCQ, BBox & 4 & \xmark & BBox, Mask \\
XLRS-Bench \cite{wang2025xlrs} & 6 Public Datasets & \xmark & MCQ, TF & 6 & 2 & BBox \\
CHOICE \cite{choice} & Multi-sat. (non-public) & \xmark & MCQ & 7 & 4 & BBox, Mask \\
VLRS-Bench & 10 Public Datasets & 129.87 & MCQ, FF, TF & 14 & 8 & BBox, Mask, DSM, NIR \\
\bottomrule
\multicolumn{7}{l}{\footnotesize \textit{Note:} MCQ = Multiple/Single Choice Questions; TF = True/False Questions; FF = Free-Form Questions. VLRS length is measured in question words.}
\end{tabular}
}

\end{table*}

\section{Related Work}
\label{sec:related}

\noindent\textbf{MLLMs.} Significant advances in vision-language understanding are largely driven by MLLMs: contrastive models such as CLIP \cite{radford2021learning} and ALIGN \cite{jia2021scaling} established scalable pretraining, while instruction-tuned systems such as LLaVA \cite{liu2023llava} and GPT-4V \cite{gpt4v} enabled stronger open-ended reasoning. However, general-domain models remain constrained by object-centric natural-image data such as COCO \cite{lin2014microsoft}, motivating RS-oriented MLLMs such as GeoChat \cite{kuckreja2024geochat}, LHRS-Bot \cite{LHRS_Bot}, VHM \cite{pang2025vhm}, SkySenseGPT \cite{luo2024skysensegpt}, GeoPixel \cite{GeoPixel}, GeoLLaVA-8K \cite{wang2025geollava}, and EarthDial \cite{Earthdial}; despite these adaptations, systematic evaluation of complex reasoning in specialized RS MLLMs remains limited.

\noindent {\textbf{Benchmarks for MLLMs.}} Comprehensive benchmarks in the general domain, such as MMBench \cite{liu2024mmbench} and SEED-Bench \cite{Li_2024_CVPR}, have been established to evaluate MLLMs. However, their focus on tasks with limited spatial and temporal complexity renders them insufficient for the unique challenges of Earth observation. This gap prompted the development of domain-specific benchmarks. Early efforts like RSVQA \cite{rsvqa} introduced visual question answering to satellite imagery but primarily assessed recognition and basic relational understanding. Subsequent benchmarks expanded this foundation by incorporating higher-resolution imagery \cite{li2023hrvqavisualquestionanswering} and addressing specific technical challenges such as reference grounding \cite{zhan2023rsvg}. Recent contributions such as RSVLM-QA \cite{RSVLM-QA} and GEOBench-VLM \cite{Geobench-vlm} further evaluate capabilities ranging from object counting to fine-grained categorization. CHOICE \cite{choice} and XLRS-Bench \cite{wang2025xlrs} include several reasoning-oriented dimensions, but, as shown in Table~\ref{tab:benchmark_comparison}, their coverage remains limited in temporal depth, RS priors, and cognition-driven task organization. Thus, existing benchmarks still leave room for a systematic evaluation framework centered on complex RS reasoning.

\section{VLRS-Bench}
\label{vlrs}

This section outlines the hierarchical reasoning taxonomy defining VLRS-Bench and describes the automated pipeline implemented for its construction. Specific details regarding the benchmark configuration and the statistical distribution of question types are presented in \textbf{Appendix~\ref{vlrs_configuration}.}

\subsection{Reasoning Dimension of VLRS-Bench}
\label{Reason}

As illustrated in Figure~\ref{fig:intro}, we structure VLRS-Bench around three L-1 reasoning levels:
\textbf{(1) Cognition.} Interpreting a scene’s current state through retrospective reasoning;
\textbf{(2) Decision.} Determining what actions should be taken based on the current observation; and
\textbf{(3) Prediction.} Forecasting what will happen next.
In the following subsections, we elaborate on the L-2 dimensions and briefly outline their associated L-3 capabilities, while comprehensive definitions for each L-3 capability are provided in \textbf{Appendix~\ref{L_3_detail}} as detailed reference material.

\subsubsection{Cognition}
This dimension evaluates the model's ability to perform deep reasoning into the causality and mechanisms of geospatial phenomena. We assess this across two complementary L-2 dimensions: static spatial relationships and dynamic spatiotemporal evolutions.


\noindent\textbf{Spatial Cognitive (SC).}
This dimension evaluates whether a model can move beyond surface-level object recognition to capture the intrinsic causal logic underlying a static scene. Its core objective is to assess whether the model can synthesize isolated visual cues into a coherent causal understanding, bridging the gap between ``what is present'' and ``why it is present.''  To this end, we incorporate four interconnected L-3 reasoning capabilities: 
\textit{1) Causal Reasoning (CR)}, which identifies latent etiological factors driving observed phenomena; 
\textit{2) Counterfactual Reasoning (CFR)}, which examines the consequences of hypothetical interventions under alternative scenarios; 
\textit{3) Mechanistic Interaction Reasoning (MIR)}, which infers implicit physical interactions and feedback mechanisms among spatial elements; and 
\textit{4) Semantic Integration Reasoning (SIR)}, which integrates low-level visual primitives into coherent high-level regional semantics.


\noindent\textbf{Spatiotemporal Cognitive (ST-C).}
Extending reasoning into the temporal domain, this dimension evaluates a model’s ability to uncover the mechanisms governing geospatial evolution. The core objective of ST-C is to assess whether the model perceives spatiotemporal change as a structured and rule-driven process rather than a random sequence, requiring an understanding of how past events shape present states. Accordingly, we assess four interconnected L-3 reasoning capabilities:
\textit{1) Spatiotemporal Causal-Chain Reasoning (ST-CCR)}, which infers and explains causal event chains across time;
\textit{2) Spatiotemporal Counterfactual Reasoning (ST-CFR)}, which explores alternative evolutionary trajectories by hypothetically modifying critical past events;
\textit{3) Spatiotemporal Evolution Reasoning (ST-ER)}, which captures functional transformations and semantic shifts of regions over time; and
\textit{4) Spatiotemporal Consistency Reasoning (ST-CR)}, which verifies the logical coherence of temporal changes under geospatial constraints.


\subsubsection{Decision}
This dimension evaluates a model’s capacity for spatial decision-making. It shifts the focus from passive observation to active strategy formulation, requiring the model to reason about how to achieve specific objectives within complex geospatial environments. This dimension is further categorized into generative planning and critical evaluation.


\noindent\textbf{Pre-event Decision (PRE-D).}
This dimension focuses on the proactive generation of spatial strategies. Its objective is to assess whether a model can synthesize environmental constraints into actionable plans, thereby moving beyond scene understanding toward goal-driven intervention. To this end, we define \textit{Planning Reasoning (PR)}, which examines the model’s ability to formulate spatially optimized solutions that satisfy implicit logistical constraints for achieving predefined objectives, such as site selection or practical route planning.


\noindent\textbf{Post-event Decision (POST-D).}
This dimension complements proactive planning with retrospective evaluation. Its objective is to assess whether a model can critically examine proposed interventions and their potential outcomes, ensuring that theoretical strategies are consistent with physical and environmental constraints. To this end, we define \textit{Evaluation Reasoning (ER)}, which examines the model’s ability to assess the feasibility and robustness of candidate plans by analyzing their compatibility with the scene context and identifying potential risks.


\subsubsection{Prediction}
This dimension evaluates a model’s capacity for spatiotemporal forecasting. It requires the model to extrapolate from historical observations to infer future states, reasoning about how observed patterns and constraints evolve over time. Predictions are assessed at both the discrete object level and the broader scene level as a whole.


\noindent\textbf{Object-level Predictive (OP).}
This dimension focuses on forecasting the future evolution of individual entities. It assesses whether a model can reason about object-specific trajectories driven by observable pressures, rather than treating the scene as a static background. Accordingly, we evaluate two L-3 reasoning capabilities:
\textit{1) Spatiotemporal Category–State Prediction Reasoning (ST-CS-PR)}, which predicts semantic state transitions by inferring future object identities from developmental trends; and
\textit{2) Spatiotemporal Morphological Prediction Reasoning (ST-M-PR)}, which extrapolates geometric evolution by modeling the physical dynamics governing changes in object shape or extent.


\noindent\textbf{Scene-level Predictive (SP).}
This dimension extends prediction to the macro-scale evolution of entire landscapes. It evaluates whether a model can reason about the inherently uncertain dynamics of complex geospatial systems, where future states are often non-deterministic. 
Accordingly, we assess two L-3 reasoning capabilities:
\textit{1) Spatiotemporal Scenario Uncertainty Prediction Reasoning (ST-SU-PR)}, which performs probabilistic forecasting by modeling multiple plausible evolutionary trajectories; and
\textit{2) Spatiotemporal Sequence Prediction Reasoning (ST-SQ-PR)}, which requires predicting scene-level spatial states by reasoning over temporal dependencies across multiple sequential observations in temporally ordered scenes.

\begin{figure}[!t]
  \centering
  \vspace{-0.1cm}
  \includegraphics[width=1\linewidth]{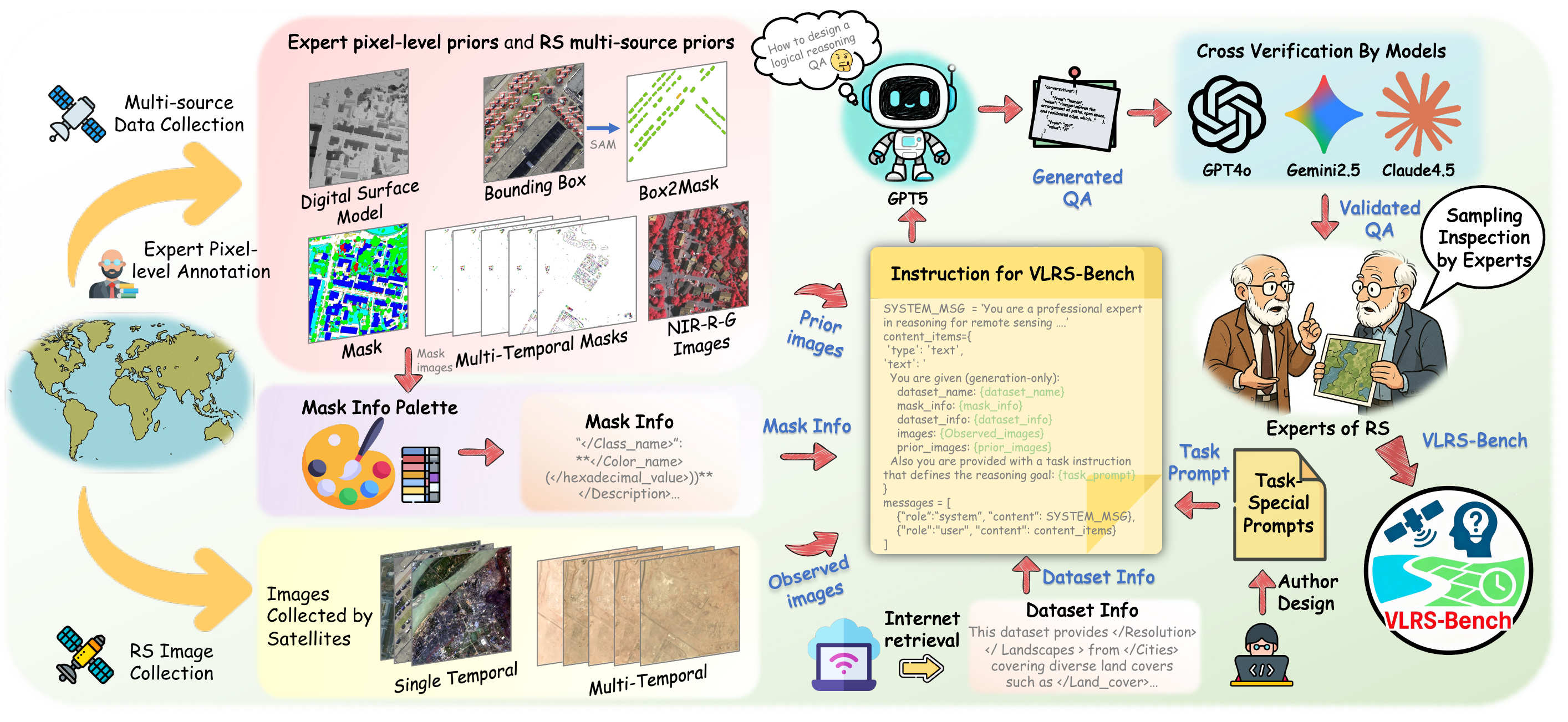}

\vspace{-0.2cm}
   \caption{Pipeline for constructing VLRS-Bench. The process integrates the target RGB image with multi-source remote sensing priors (e.g., DSM and expert masks) to form a structured multimodal instruction, which guides GPT-5-chat to produce reasoning tasks across cognitive dimensions. Each generated item is then verified through a three-stage protocol, including automated filtering, multi-MLLM cross-validation, and human expert review for quality control.}
    \vspace{-0.2cm}
   \label{figpipl}
\end{figure}

\subsection{Pipeline of VLRS-Bench}
\label{pipelines}

This section details the automated construction pipeline of VLRS-Bench, encompassing data curation, instruction synthesis, and rigorous quality verification. Further technical specifications regarding the pipeline implementation are provided in \textbf{Appendix~\ref{sec:pipeline}}.

\subsubsection{Data preparation of VLRS-Bench}

To construct a comprehensive remote sensing (RS) reasoning benchmark, we curate a diverse data foundation comprising publicly available RS imagery spanning multiple sources, time periods, and diverse scene types and contexts.

\noindent\textbf{Single-temporal datasets:} We select several public semantic segmentation and object detection datasets, including LoveDA \cite{wang2loveda}, Potsdam \cite{rottensteiner2012isprs}, Vaihingen \cite{rottensteiner2012isprs}, GID-15 \cite{GID2020}, DIOR \cite{li2020object}, and FAIR1M \cite{sun2022fair1m}. These datasets span native ground sampling distances from 0.05 m to 30 m and cover a wide range of land-cover categories, providing a robust multi-scale and semantically diverse visual foundation for reasoning.

\noindent\textbf{Multi-temporal datasets:} We introduce typical change detection and temporal inference datasets such as xView2 \cite{Gupta_2019_CVPR_Workshops}, SECOND \cite{yang2021semanticchangedetectionasymmetric}, miniUCD \cite{tian2020hiucdlargescaledataseturban}, and SpaceNet7 \cite{vanetten2019spacenetremotesensingdataset}. These datasets provide paired multi-temporal image sequences capturing explicit temporal changes and corresponding expert pixel-level annotated masks, covering dynamic scenes such as urban expansion and land surface changes.


To further expand pixel-level annotations, we employ the SAMRS framework \cite{wang2023samrs} to convert bounding-box annotations from public datasets into high-quality segmentation masks, unifying the annotation format at scale.

\subsubsection{Instruction of Pipeline}

As illustrated in Figure~\ref{figpipl}, VLRS-Bench is constructed via a highly automated generation pipeline. For each RS scene, the pipeline fuses the observed RGB image with auxiliary RS priors (e.g., DSM and NIR), selected multi-temporal references, and expert-provided pixel-level masks (used as conditioning priors rather than prediction targets). It then packages these inputs—together with mask definitions, dataset metadata, and a task-specific prompt—into a unified multimodal instruction. Specifically, the instruction consists of the following five complementary components:


\noindent\textbf{Observed Image.} The original RGB RS image provided as the primary visual input for the task.

\noindent\textbf{Prior Images.} This component incorporates multidimensional cross-modal contextual information through three complementary elements:
\textit{(1) Multi-source RS Priors}, which load DSM and NIR data to provide structural and spectral cues beyond standard RGB perception;
\textit{(2) Expert Pixel-Level Priors}, which use predefined masks as conditioning inputs rather than prediction targets to constrain reasoning within expert-defined contexts; and
\textit{(3) Multi-temporal Reference Priors}, which leverage selected temporal observations as causal anchors to support logically consistent multi-temporal reasoning under explicit temporal constraints.


\noindent\textbf{Mask Info.} It defines a dedicated palette for each mask type, mapping pixel colors to corresponding land-cover categories. Each category is accompanied by a textual description that specifies its semantic meaning and functional role, providing the model with structured semantic priors.


\noindent\textbf{Dataset Info.} It provides metadata describing the dataset’s composition, source, and characteristics, offering high-level contextual information for task interpretation.


\noindent\textbf{Task Special Prompt.} This component provides a task-specific prompt tailored to each subtask (Section~\ref{Reason}), defining the underlying reasoning logic and problem orientation. It guides the model toward the intended analytical objectives and multi-step reasoning processes, with illustrative task-specific examples provided as reference.

Together, these components enforce task alignment through RS priors, format constraints, and capability-specific prompts. The subsequent cross-model verification and expert review further screen whether each item matches its intended reasoning category, is grounded in visual evidence, and uses standard RS terminology throughout the benchmark.

After instruction assembly, the pipeline iteratively generates aligned question--answer (QA) pairs using GPT-5-chat. These QA pairs are then organized into multiple-choice, single-choice, fill-in-the-blank, and true/false formats, covering all reasoning dimensions systematically.

\subsubsection{Verification of Pipeline}
Following the automatic generation of QA pairs, we apply a three-stage verification pipeline to ensure the quality and reliability of VLRS-Bench.

First, an automated filtering stage quantitatively evaluates each item based on fundamental criteria, including clarity, image relevance, and ambiguity, removing low-quality samples. Second, to mitigate single-model bias, the remaining items undergo multi-model cross-validation: a proposer model generates a design justification for each QA pair, which is then examined for logical consistency and factual correctness by a panel of independent verification models.

Finally, items that pass automated verification proceed to a full human expert review. Starting from more than 6,500 generated candidates, automated filtering and cross-model verification retain 2,694 items. A panel of nine Ph.D.-level RS experts then reviews all retained candidates for professional relevance, logical rigor, visual grounding, and ground-truth correctness, removing 694 items and yielding the final 2,000-item benchmark. This verification pipeline ensures that only high-quality and reliable items are included in VLRS-Bench.

\begin{table*}[!t]
\centering
\caption{Performance evaluation of different MLLMs on the L-3 subtasks of VLRS-Bench. The models are categorized into general MLLMs and RS MLLMs, with average scores provided for each category to support direct comparison. Overall scores are averaged over supported tasks.}
\vspace{-0.1cm}
\renewcommand{\arraystretch}{1.2}
\footnotesize
\resizebox{\textwidth}{!}{
\begin{tabular}{l | cccccccc | cc | cccc | c}
\toprule
\multirow{2}{*}{\textbf{Models}} & \multicolumn{8}{c}{\textbf{Cognition}} & \multicolumn{2}{c}{\textbf{Decision}} & \multicolumn{4}{c}{\textbf{Prediction}} & \multirow{2}{*}{\textbf{Avg. Score.}} \\
\cmidrule(lr){2-9} \cmidrule(lr){10-11} \cmidrule(lr){12-15}
& \textbf{CR} & \textbf{CFR} & \textbf{SIR} & \textbf{MIR} & \textbf{ST-CFR} & \textbf{ST-CCR} & \textbf{ST-ER} & \textbf{ST-CR} & \textbf{PR} & \textbf{ER} & \textbf{ST-CS-PR} & \textbf{ST-M-PR} & \textbf{ST-SU-PR} & \textbf{ST-SQ-PR} & \\
\midrule
\rowcolor{orange!25}\multicolumn{16}{l}{\textit{\textbf{General MLLMs}}} \\
GPT-5.4 & \textbf{0.528} & \textbf{0.556} & \textbf{0.600} & \textbf{0.448} & \textbf{0.480} & \textbf{0.488} & \textbf{0.460} & 0.400 & 0.314 & 0.402 & \textbf{0.508} & 0.448 & \textbf{0.460} & \textbf{0.512} & \textbf{0.458} \\
GPT-5-chat & 0.424 & 0.400 & 0.472 & 0.316 & 0.276 & 0.352 & 0.380 & 0.368 & 0.388 & 0.388 & 0.388 & 0.276 & 0.280 & 0.292 & 0.361 \\
GPT-4o-2024-11-20 & 0.376 & 0.432 & 0.420 & 0.332 & 0.360 & 0.352 & 0.400 & 0.364 & 0.286 & 0.334 & 0.416 & 0.352 & 0.284 & 0.340 & 0.354 \\
GPT-4o-mini & 0.428 & 0.416 & 0.428 & 0.328 & 0.400 & 0.356 & 0.352 & 0.336 & 0.248 & 0.304 & 0.424 & 0.372 & 0.280 & 0.304 & 0.346 \\
Gemini-3.1-Pro-Preview & 0.420 & 0.448 & 0.496 & 0.396 & 0.476 & 0.436 & \textbf{0.460} & \textbf{0.456} & \textbf{0.428} & \textbf{0.458} & 0.400 & 0.360 & 0.428 & 0.432 & 0.436 \\
Gemini-2.5-flash & 0.200 & 0.188 & 0.264 & 0.240 & 0.188 & 0.160 & 0.116 & 0.160 & 0.232 & 0.240 & 0.164 & 0.168 & 0.116 & 0.152 & 0.191 \\
Claude-3.5-haiku & 0.308 & 0.316 & 0.304 & 0.360 & 0.192 & 0.208 & 0.232 & 0.200 & 0.372 & 0.370 & 0.208 & 0.224 & 0.248 & 0.168 & 0.278 \\
Claude-Opus-4.6 & 0.272 & 0.264 & 0.372 & 0.408 & 0.452 & 0.392 & 0.336 & 0.300 & 0.398 & 0.348 & 0.264 & 0.320 & 0.396 & 0.340 & 0.350 \\
Grok-2-vision & 0.188 & 0.216 & 0.288 & 0.368 & 0.232 & 0.240 & 0.240 & 0.280 & 0.252 & 0.300 & 0.220 & 0.196 & 0.172 & 0.148 & 0.243 \\
\cmidrule[0.4pt](lr){1-1} \cmidrule(lr){2-9} \cmidrule(lr){10-11} \cmidrule(lr){12-15}  \cmidrule(lr){16-16}
Deepseek-vl2 & 0.372 & 0.392 & 0.452 & 0.344 & 0.144 & 0.216 & 0.200 & 0.148 & 0.416 & 0.446 & 0.096 & 0.128 & 0.064 & 0.080 & 0.273 \\
GLM-4.5v & 0.268 & 0.136 & 0.248 & 0.312 & 0.152 & 0.084 & 0.084 & 0.132 & 0.312 & 0.340 & 0.132 & 0.172 & 0.180 & 0.112 & 0.207 \\
LLama-3.2-11B & 0.232 & 0.228 & 0.244 & 0.264 & \xmark & \xmark & \xmark & \xmark & 0.292 & 0.286 & \xmark & \xmark & \xmark & \xmark & 0.266 \\
LLama-3.2-90B & 0.368 & 0.364 & 0.356 & 0.308 & 0.236 & 0.268 & 0.300 & 0.268 & 0.408 & 0.430 & 0.308 & 0.352 & 0.268 & 0.212 & 0.330 \\
Qwen2.5-VL-7B & 0.256 & 0.172 & 0.384 & 0.176 & 0.224 & 0.324 & 0.248 & 0.216 & 0.198 & 0.238 & 0.328 & 0.280 & 0.232 & 0.212 & 0.245 \\
Qwen2.5-VL-32B & 0.292 & 0.312 & 0.368 & 0.296 & 0.244 & 0.300 & 0.256 & 0.236 & 0.370 & 0.308 & 0.276 & 0.284 & 0.236 & 0.156 & 0.288 \\
Qwen2.5-VL-72B & 0.216 & 0.300 & 0.296 & 0.392 & 0.316 & 0.240 & 0.216 & 0.204 & 0.402 & 0.370 & 0.172 & 0.180 & 0.204 & 0.188 & 0.279 \\
Qwen3-VL-2B & 0.350 & 0.361 & 0.468 & 0.185 & 0.204 & 0.428 & 0.412 & 0.336 & 0.287 & 0.260 & 0.380 & 0.312 & 0.304 & 0.312 & 0.322 \\
Qwen3-VL-8B & 0.341 & 0.363 & 0.472 & 0.211 & 0.388 & 0.405 & 0.398 & 0.371 & 0.233 & 0.275 & 0.381 & 0.401 & 0.401 & 0.384 & 0.346 \\
Qwen3-VL-32B & 0.372 & 0.416 & 0.476 & 0.316 & 0.408 & 0.428 & 0.388 & 0.392 & 0.416 & 0.364 & 0.388 & \textbf{0.456} & 0.336 & 0.372 & 0.394 \\
\rowcolor{gray!30} \textit{Gen. MLLMs Avg.} & \textit{0.327} & \textit{0.331} & \textit{0.390} & \textit{0.316} & \textit{0.298} & \textit{0.315} & \textit{0.304} & \textit{0.287} & \textit{0.329} & \textit{0.340} & \textit{0.303} & \textit{0.293} & \textit{0.272} & \textit{0.262} & \textit{0.314} \\
\midrule
\rowcolor{orange!25}\multicolumn{16}{l}{\textit{\textbf{Remote Sensing MLLMs}}} \\
GeoChat & 0.280 & 0.332 & 0.360 & 0.308 & \xmark & \xmark & \xmark & \xmark & 0.352 & 0.356 & \xmark & \xmark & \xmark & \xmark & 0.337 \\
VHM & 0.302 & 0.297 & 0.308 & 0.210 & \xmark & \xmark & \xmark & \xmark & 0.324 & 0.332 & \xmark & \xmark & \xmark & \xmark & 0.304 \\
ScoreRS w/ SFT & 0.403 & 0.367 & 0.421 & 0.345 & 0.294 & 0.310 & 0.288 & 0.284 & 0.382 & 0.419 & 0.341 & 0.320 & 0.313 & 0.295 & 0.349 \\
ScoreRS w/ RL & 0.313 & 0.338 & 0.382 & 0.295 & 0.399 & 0.335 & 0.367 & 0.392 & 0.409 & 0.371 & 0.338 & 0.313 & 0.382 & 0.342 & 0.360 \\
\rowcolor{gray!30} \textit{RS MLLMs Avg.} & \textit{0.325} & \textit{0.334} & \textit{0.368} & \textit{0.290} & \textit{0.347} & \textit{0.323} & \textit{0.328} & \textit{0.338} & \textit{0.367} & \textit{0.370} & \textit{0.340} & \textit{0.317} & \textit{0.348} & \textit{0.319} & \textit{0.337} \\

\bottomrule
\end{tabular}
}
\label{tab:model_performance}
\vspace{-0.1cm}
\end{table*}

\section{Experiment}
\label{sec:experiment}

We evaluate the reasoning capabilities of current MLLMs on VLRS-Bench. This section presents the experimental setup, including the evaluated models, evaluation protocols, and comprehensive experimental results for systematic analysis.

\subsection{Experimental Setup}
\label{subsec:setup}

We evaluate a broad range of state-of-the-art MLLMs, grouped into three categories. 
The first category includes closed-source proprietary models, such as the GPT-5 family (GPT-5.4 and GPT-5-chat) \cite{openai_gpt5_2025}, the GPT-4o series \cite{hurst2024gpt}, Claude-3.5-haiku, Claude-Opus-4.6, the Gemini family (Gemini-3.1-Pro-Preview and Gemini-2.5-flash) \cite{comanici2025gemini}, and Grok-2-Vision \cite{xai_grok2vision_2024}. 
The second category comprises open-source general-purpose models, including DeepSeek-VL \cite{lu2024deepseek}, GLM-4.5V \cite{glm2024chatglm}, multiple sizes of the Llama-3.2 series \cite{dubey2024llama}, and the Qwen2.5-VL \cite{bai2025qwen2} and Qwen3-VL series \cite{qwen3}. 
The third category focuses on domain-specialized models, including ScoreRS \cite{muhtar2025qualitydriven} under both SFT and RL training paradigms. To ensure fair comparison, all models are evaluated under a zero-shot setting using a standardized prompt, the question text, and the observed RGB frames.

\subsection{Evaluation Strategy}
\label{subsec:eval}

To evaluate reasoning performance in VLRS-Bench, we adopt a tiered scoring scheme tailored to different question formats. 
For 8-select-N multiple-choice questions (N $\geq$ 2), partial credit is assigned: 1.0 for perfect selections, 0.5 for incomplete but error-free answers, and 0 for any incorrect choice. 
Other question formats, including 5-select-1 single-choice and binary true/false questions, are scored with one point per correct response. The fill-in-the-blank questions are evaluated using a semantic similarity-based criterion rather than exact string matching. We compute embedding similarity using the all-MiniLM-L6-v2 model \cite{all_minilm} and consider an answer correct if its similarity to the ground truth exceeds an 80\% threshold. This threshold was selected through expert calibration to accept valid paraphrases while suppressing semantically incorrect answers. Partial credit is supported: responses receive 0.5 points when the answer is partially correct, and 1.0 point only when all blanks are filled correctly. Finally, the overall performance is reported as the mean percentage score across questions in the tasks supported by each model, providing a holistic measure of high-level reasoning performance in complex remote sensing.



\subsection{Main Results}

This section presents a fine-grained, detailed analysis of L-3 capabilities to identify specific cognitive nuances and reasoning bottlenecks. A complementary macro-level analysis at the L-2 dimension level, along with qualitative visualizations for each individual L-3 task, is provided in \textbf{Appendix~\ref{sec:analysis_l1l2}} and \textbf{Appendix~\ref{sec:visualizations}}, respectively, for completeness.

\subsubsection{Results on Cognition Ability}

Table~\ref{tab:model_performance} reveals distinct performance patterns under the Cognition dimension, which evaluates causal, relational, and process-oriented reasoning beyond visual recognition. Two consistent trends can be observed. \textbf{(1) Degradation from single-state cognition to spatiotemporal reasoning.} General MLLMs demonstrate noticeably weaker performance when reasoning requires modeling temporal evolution and change attribution, compared with reasoning from a single observation. This suggests that their reasoning is largely anchored in static semantic cues, with limited capacity for modeling temporal dependencies. In contrast, RS MLLMs show more similar group-average scores across single-state and spatiotemporal cognition tasks, indicating stronger alignment with geospatial change semantics and improved ability to associate observed variations with underlying causes. \textbf{(2) Disparity between semantic integration and mechanistic inference.} At the aggregate level, Semantic Integration Reasoning remains higher than Mechanistic Interaction Reasoning for both General and RS MLLMs, suggesting that current models are more proficient at organizing observable semantic elements than at inferring latent interaction mechanisms. Mechanistic reasoning requires abstraction over indirect, non-visual dependencies, which remains a formidable challenge even for specialized domain-adapted models. Collectively, these results suggest that cognition in remote sensing extends well beyond simple semantic recognition and that VLRS-Bench provides a structured evaluation setting that exposes such complex reasoning limitations in a clear and systematic manner across cognition tasks.

\subsubsection{Results on Decision Ability}

The Decision dimension assesses a model’s capability to perform spatial planning and outcome assessment under explicit constraints. Analysis of the results in Table~\ref{tab:model_performance} reveals two consistent patterns. \textbf{(1) Decision reasoning is not determined by model scale alone.} Frontier general models often achieve strong scores, but their gains are uneven across PR and ER, while domain-specialized RS models remain competitive on several decision subtasks. For example, ScoreRS w/ SFT exceeds Qwen2.5-VL-72B on ER (0.419 vs.\ 0.370). This indicates that planning and evaluation reasoning require not only general logical capacity but also grounded geospatial constraints and operational criteria. \textbf{(2) Non-equivalence between PR and ER.} PR and ER do not form a single, monotonic decision capability. Across both General and RS MLLMs, the relative performance between PR and ER varies substantially, indicating that success in one subtask does not guarantee robust competence in the other. Some models achieve comparable scores on PR and ER, while others exhibit clear asymmetry, regardless of model scale or domain specialization. This variability suggests that decision-making in VLRS-Bench involves at least two decoupled processes: generating plausible candidate actions under spatial constraints and assessing their potential outcomes. By explicitly separating PR and ER, VLRS-Bench avoids collapsing decision reasoning into a single aggregate score and instead reveals meaningful structural differences in how models plan actions and judge their practical consequences in context.

\subsubsection{Results on Prediction Ability}

The Prediction dimension evaluates a model’s ability to anticipate future geospatial states under varying degrees of temporal dependency and uncertainty. Results in Table~\ref{tab:model_performance} also reveal two notable patterns. \textbf{(1) Increasing difficulty from local state prediction to global system evolution.} Model performance tends to decline as prediction tasks shift from localized object changes to integrated scene-level evolution. General MLLMs perform relatively better when future states can be inferred from limited and localized cues, but struggle when predictions depend on the coordinated evolution of multiple entities over time, highlighting limitations in modeling long-range temporal and structural dependencies. \textbf{(2) Sensitivity to uncertainty and temporal complexity.} RS MLLMs demonstrate more differentiated behavior across prediction settings, showing stronger performance when future outcomes follow clearer physical or semantic regularities, while exhibiting increased variability under prediction scenarios with higher uncertainty and longer temporal horizons. This contrast indicates that prediction in remote sensing is not a uniform capability, but depends critically on how well models internalize domain-specific evolution patterns and uncertainty structures. Overall, these results indicate that the Prediction dimension imposes distinct challenges beyond cognition and decision-making, and they illustrate how VLRS-Bench systematically exposes differences in models’ ability to reason about future geospatial dynamics rather than single static observations.

\begin{figure}[!t]
    \centering
    \includegraphics[width=1\linewidth]{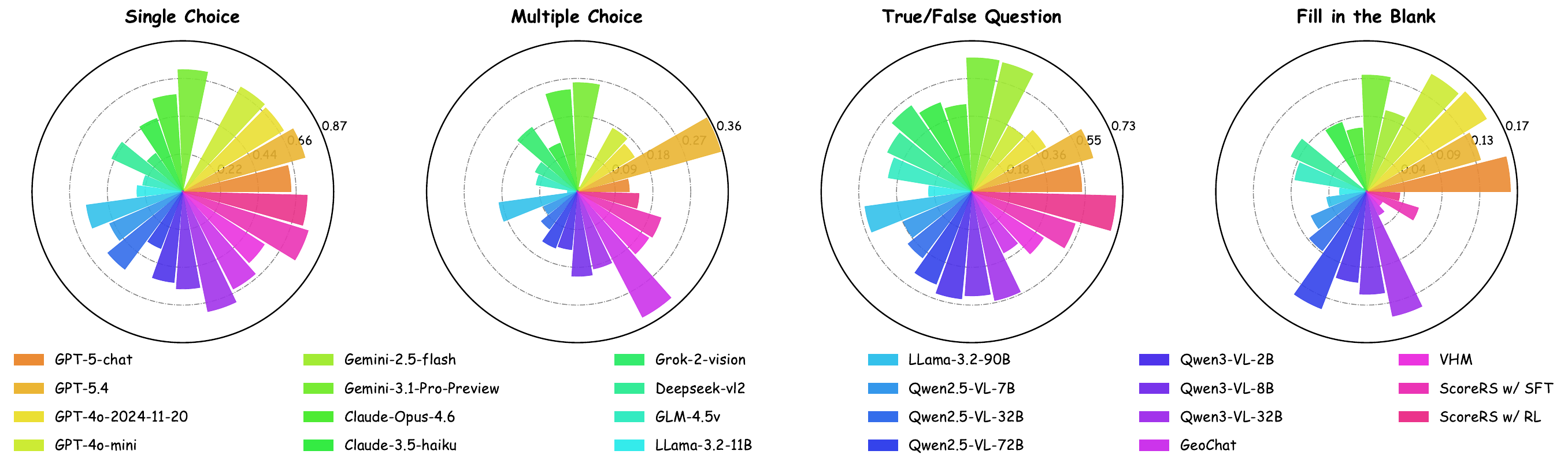}
    \caption{QA-format performance on VLRS-Bench. The results show consistently higher scores on single-choice and true/false questions than on multi-choice and fill-in-the-blank questions, indicating that VLRS-Bench exposes model differences in option selection, exhaustive verification, and concise answer generation in RS reasoning.}
    \vspace{-0.2cm}
    \label{fig:qa_type_analysis}
\end{figure}

\subsubsection{Results on different QA-type tasks}

Figure~\ref{fig:qa_type_analysis} summarizes the mean scores of evaluated MLLMs with available QA-format breakdowns across four QA-types, revealing clear disparities in how models respond to different answer spaces and scoring mechanisms. Two consistent patterns can be observed. \textbf{(1) Sensitivity to answer-space complexity.} Across models in Figure~\ref{fig:qa_type_analysis}, Single-Choice and True/False questions obtain substantially higher average scores (52.6\% and 47.5\%) than Multi-Choice and Fill-in-the-Blank questions (16.6\% and 9.0\%). The same pattern holds for both General MLLMs and RS MLLMs, indicating that current models remain more reliable under tightly constrained answer spaces, while performance drops when they must identify multiple valid options or generate concise semantic answers. \textbf{(2) Partial-credit behavior under uncertainty.} Multi-Choice scores are non-zero for most models but remain far below Single-Choice and True/False scores, suggesting that models often recover partial option sets without performing exhaustive verification. Under partial-credit evaluation, the score depends on both coverage and selection accuracy: conservative models may accumulate partial scores by selecting few options, whereas models attempting broader coverage risk zero credit when a single unsupported option is included. Fill-in-the-Blank is the lowest-scoring format for most models, further exposing the difficulty of open-form answer generation even when semantic matching is used.

\section{Conclusion}
\label{sec:conclusion}

This study introduces VLRS-Bench, the first benchmark systematically evaluating the complex reasoning capabilities of Multimodal Large Language Models (MLLMs) in remote sensing (RS) across dimensions of Cognition, Decision, and Prediction. VLRS-Bench reveals critical limitations of current MLLMs: in Cognition, they struggle to bridge visual patterns with deep RS physical causal mechanisms and counterfactual possibilities; in Decision, general models score lower on spatial planning than on retrospective evaluation; and in Prediction, they lack the temporal scalability to grasp the logic of dynamic landscape evolution. Furthermore, RS-specialized models often outperform larger generalist models, showing the value of domain specialization for reasoning about spatial constraints and temporal change. We anticipate VLRS-Bench will catalyze a shift toward MLLMs capable of substantive cognition and decision-making within Earth's complex systems.

\paragraph{Data and code availability.}
The dataset is available at \url{https://huggingface.co/datasets/thislzm/VLRS-Bench}. The project page is \url{https://github.com/MiliLab/VLRS-Bench}.

\begin{ack}
\input{funding_disclosure.tex}
\end{ack}

\bibliographystyle{plainnat}
\bibliography{main}

\newpage
\appendix

\setcounter{page}{1}
\renewcommand{\thepage}{A\arabic{page}}
\setcounter{section}{0}
\setcounter{table}{0}
\setcounter{figure}{0}
\renewcommand{\thetable}{A\arabic{table}}
\renewcommand{\thefigure}{A\arabic{figure}}

\section{Overview of the Appendix}\label{sec:overview}
This appendix supplements the proposed VLRS-Bench with additional experimental results and details excluded from the main paper due to space constraints.
The appendix is organized as follows:
\begin{itemize}
    \item Sec.~\ref{sec:analysis_l1l2}: More Analysis of L-2 Tasks in VLRS-Bench.
    \item Sec.~\ref{sec:vlrs_details}: More Details of VLRS-Bench.
    \item Sec.~\ref{sec:pipeline}: VLRS-Bench Pipeline Details
    \item Sec.~\ref{sec:visualizations}: Visualizations of Random Sampling Cases.
    \item Sec.~\ref{app-datasheets}: Datasheets for the VLRS-Bench.
    \item Sec.~\ref{app-limitation}: Discussion on Limitations and Societal Impact.
\end{itemize}

\section{More Analysis of L-1 and L-2 Dimensions in VLRS-Bench.}
\label{sec:analysis_l1l2}

To comprehensively evaluate the remote sensing reasoning capabilities of various MLLMs within the VLRS-Bench framework, we extend our analysis to the aggregated L-1 and L-2 dimensions, uncovering macroscopic performance patterns that reveal the fundamental cognitive strengths and limitations of current models in the geospatial domain.

\subsection{Results on Cognition Ability}

The Cognition dimension serves as the foundational layer of geospatial intelligence, structured into Spatial Cognitive and Spatiotemporal Cognitive tasks. The performance patterns observed here provide compelling evidence for the structural rationality of the VLRS-Bench design. We summarize two key insights: 

\textbf{(1) Differentiation of Static and Dynamic Reasoning.} A noticeable performance divergence is observed in general MLLMs, which achieve respectable scores on Spatial Cognitive tasks but decline on Spatiotemporal Cognitive tasks. This phenomenon validates the structural necessity of distinguishing between spatial and spatiotemporal cognition in VLRS-Bench. It demonstrates that static visual perception and dynamic causal reasoning are fundamentally distinct cognitive capabilities. General models, while capable of identifying objects in a single frame, lack the temporal logic required to interpret how geospatial features evolve. By decoupling these dimensions, VLRS-Bench successfully isolates the temporal bottleneck inherent in current multimodal systems, showing that high perceptual accuracy does not automatically translate to an understanding of dynamic geospatial processes.

\textbf{(2) Hierarchy of Cognitive Depth across Dimensions.} Expanding the analysis to the broader L-1 dimension, the results establish a clear difficulty gradient: among General MLLMs, Decision has the highest group mean, while Prediction has the lowest. This hierarchy validates the rationality of the L-1 design, confirming that the three dimensions capture distinct levels of cognitive complexity. Decision represents constraint-based logical execution, Cognition bridges perception and physical causality, and Prediction requires extrapolating unseen states. This distinct stratification shows that VLRS-Bench covers the full spectrum of geospatial intelligence, effectively distinguishing between models that can merely follow rules and those that possess a grounded understanding of the Earth's physical evolution.

\begin{table*}[!t]
\centering
\caption{Performance evaluation of different MLLMs on the L-1 and L-2 dimension of VLRS-Bench. The models are categorized into General Models and Remote Sensing (RS) Models, with average scores provided for each category. Dimension scores are averaged over supported tasks; group means average the available model scores in each column.}

\renewcommand{\arraystretch}{1}
\footnotesize
\resizebox{\textwidth}{!}{%
\begin{tabular}{l | ccc | ccc | ccc}
\toprule
 \multirow{2}{*}{\textbf{Model}} & \multicolumn{3}{c}{\textbf{Cognition}} & \multicolumn{3}{c}{\textbf{Decision}} & \multicolumn{3}{c}{\textbf{Prediction}} \\
\cmidrule(lr){2-4} \cmidrule(lr){5-7} \cmidrule(lr){8-10}
 & \textbf{SC} & \textbf{ST-C} & \textbf{Avg.Score} & \textbf{Pre-D} & \textbf{Post-D} & \textbf{Avg.Score} & \textbf{OP} & \textbf{SP} & \textbf{Avg.Score} \\
\midrule
\rowcolor{orange!25}\multicolumn{10}{l}{\textit{\textbf{General MLLMs}}} \\
GPT-5.4 & \textbf{0.533} & \textbf{0.457} & \textbf{0.495} & 0.314 & 0.402 & 0.358 & \textbf{0.478} & \textbf{0.486} & \textbf{0.482} \\
GPT-5-chat & 0.403 & 0.344 & 0.374 & 0.388 & 0.388 & 0.388 & 0.332 & 0.286 & 0.309 \\
GPT-4o-2024-11-20 & 0.390 & 0.369 & 0.380 & 0.286 & 0.334 & 0.310 & 0.384 & 0.312 & 0.348 \\
GPT-4o-mini & 0.400 & 0.361 & 0.381 & 0.248 & 0.304 & 0.276 & 0.398 & 0.292 & 0.345 \\
Gemini-3.1-Pro-Preview & 0.440 & \textbf{0.457} & 0.449 & \textbf{0.428} & \textbf{0.458} & \textbf{0.443} & 0.380 & 0.430 & 0.405 \\
Gemini-2.5-flash & 0.223 & 0.156 & 0.190 & 0.232 & 0.240 & 0.236 & 0.166 & 0.134 & 0.150 \\
Claude-3.5-haiku & 0.322 & 0.208 & 0.265 & 0.372 & 0.370 & 0.371 & 0.216 & 0.208 & 0.212 \\
Claude-Opus-4.6 & 0.329 & 0.370 & 0.350 & 0.398 & 0.348 & 0.373 & 0.292 & 0.368 & 0.330 \\
Grok-2-vision & 0.265 & 0.248 & 0.257 & 0.252 & 0.300 & 0.276 & 0.208 & 0.160 & 0.184 \\
\cmidrule[0.4pt](lr){1-1} \cmidrule(lr){2-4} \cmidrule(lr){5-7} \cmidrule(lr){8-10}
Deepseek-vl2 & 0.390 & 0.177 & 0.284 & 0.416 & 0.446 & 0.431 & 0.112 & 0.072 & 0.092 \\
GLM-4.5v & 0.241 & 0.113 & 0.177 & 0.312 & 0.340 & 0.326 & 0.152 & 0.146 & 0.149 \\
LLama-3.2-11B & 0.242 & \xmark & 0.242 & 0.292 & 0.286 & 0.289 & \xmark & \xmark & \xmark \\
LLama-3.2-90B & 0.349 & 0.268 & 0.309 & 0.408 & 0.430 & 0.419 & 0.330 & 0.240 & 0.285 \\
Qwen2.5-VL-7B & 0.247 & 0.253 & 0.250 & 0.198 & 0.238 & 0.218 & 0.304 & 0.222 & 0.263 \\
Qwen2.5-VL-32B & 0.317 & 0.259 & 0.288 & 0.370 & 0.308 & 0.339 & 0.280 & 0.196 & 0.238 \\
Qwen2.5-VL-72B & 0.301 & 0.244 & 0.273 & 0.402 & 0.370 & 0.386 & 0.176 & 0.196 & 0.186 \\
Qwen3-VL-2B & 0.341 & 0.345 & 0.343 & 0.287 & 0.260 & 0.273 & 0.346 & 0.308 & 0.327 \\
Qwen3-VL-8B & 0.347 & 0.391 & 0.369 & 0.233 & 0.275 & 0.254 & 0.391 & 0.393 & 0.392 \\
Qwen3-VL-32B & 0.395 & 0.404 & 0.400 & 0.416 & 0.364 & 0.390 & 0.422 & 0.354 & 0.388 \\
\rowcolor{gray!30} \textit{Gen. MLLMs Avg.} & \textit{0.341} & \textit{0.301} & \textit{0.319} & \textit{0.329} & \textit{0.340} & \textit{0.335} & \textit{0.298} & \textit{0.267} & \textit{0.282} \\
\midrule
\rowcolor{orange!25}\multicolumn{10}{l}{\textit{\textbf{Remote Sensing MLLMs}}} \\
GeoChat & 0.320 & \xmark & 0.320 & 0.352 & 0.356 & 0.354 & \xmark & \xmark & \xmark \\
VHM & 0.279 & \xmark & 0.279 & 0.324 & 0.332 & 0.328 & \xmark & \xmark & \xmark \\
ScoreRS w/ SFT & 0.384 & 0.294 & 0.339 & 0.382 & 0.419 & 0.401 & 0.331 & 0.304 & 0.317 \\
ScoreRS w/ RL & 0.332 & 0.373 & 0.353 & 0.409 & 0.371 & 0.390 & 0.326 & 0.362 & 0.344 \\
\rowcolor{gray!30} \textit{RS MLLMs Avg.} & \textit{0.329} & \textit{0.334} & \textit{0.323} & \textit{0.367} & \textit{0.370} & \textit{0.368} & \textit{0.328} & \textit{0.333} & \textit{0.331} \\

\bottomrule
\end{tabular}%

}
\label{tab:agg_capabilities}

\end{table*}

\subsubsection{Results on Decision Ability}

The Decision dimension examines a model’s ability to select, compare, and evaluate actions under explicit geospatial constraints, and is decomposed into Pre-Decision (Pre-D) and Post-Decision (Post-D) tasks. The results in Table~\ref{tab:agg_capabilities} reveal that decision-making in remote sensing is not a monolithic capability, but a staged reasoning process with distinct cognitive demands.

\textbf{(1) Asymmetric Difficulty between Pre-Decision and Post-Decision Reasoning.} 
Across General MLLMs, performance on Pre-D and Post-D tasks exhibits noticeable asymmetry. While several models maintain comparable scores between the two stages, others show a clear decline in Post-D performance, indicating difficulty in evaluating downstream consequences after a decision has been made. This suggests that Pre-D tasks primarily assess feasibility filtering under spatial constraints, whereas Post-D tasks require counterfactual reasoning and outcome-aware comparison. RS MLLMs are more balanced across the two stages on average (0.367 on Pre-D and 0.370 on Post-D). Within this group, \textit{ScoreRS w/ SFT} is stronger on Post-D (0.419 vs.\ 0.382), while its RL counterpart is stronger on Pre-D (0.409 vs.\ 0.371). This staged decomposition confirms that VLRS-Bench effectively separates early constraint satisfaction from later consequence evaluation.

\textbf{(2) Domain-grounded Decision Criteria beyond Model Scale.}
Although larger models generally achieve higher Decision scores, scale alone does not fully account for decision performance. Domain-aligned models remain competitive with several substantially larger General MLLMs, particularly in tasks requiring adherence to geospatial norms and operational criteria. For example, \textit{ScoreRS w/ SFT} surpasses Qwen2.5-VL-72B in average Decision score despite a significant parameter gap. This indicates that the Decision dimension emphasizes normative judgment and structured comparison rather than abstract reasoning depth alone. By embedding domain-specific constraints into both Pre-D and Post-D tasks, VLRS-Bench distinguishes generic logical competence from grounded decision-making ability, highlighting a critical gap between general-purpose reasoning and operationally valid geospatial decisions.

\subsubsection{Results on Prediction Ability}

The Prediction dimension evaluates a model’s capacity to extrapolate future geospatial states from observed dynamics, and is decomposed into Object-level Prediction (OP) and Scene-level Prediction (SP) tasks. Unlike Cognition and Decision, Prediction requires models to reason beyond observable evidence and to internalize assumptions about temporal continuity, interaction, and uncertainty.

\textbf{(1) Distinct Predictive Assumptions at Object and Scene Levels.}
A consistent performance gap is observed between OP and SP tasks across General MLLMs, with average scores decreasing from 0.298 to 0.267. This gap reflects not merely a difference in spatial scale, but a shift in predictive assumptions. OP tasks primarily rely on localized temporal continuity, where future states can be inferred from object-level inertia and short-term trends. In contrast, SP tasks require models to anticipate coordinated evolution across multiple entities, enforcing global consistency under shared physical and semantic constraints. The pronounced degradation in SP performance indicates that current general models lack an internal mechanism to maintain system-level coherence when extrapolating complex geospatial processes.

\textbf{(2) Domain-informed Modeling of Temporal Uncertainty.}
RS MLLMs exhibit more stable performance across OP and SP tasks, with \textit{ScoreRS w/ RL} achieving 0.326 on OP and 0.362 on SP. This relative robustness suggests that domain-aligned models encode stronger priors about plausible geospatial evolution, reducing structurally invalid predictions even under increased uncertainty. At the group level, Prediction scores are lower than Decision scores for both model groups and lower than Cognition scores for General MLLMs. The different patterns across OP and SP suggest that prediction difficulty depends on how uncertainty carries from local changes to scene-wide evolution. By explicitly separating object-level extrapolation from scene-level evolution, VLRS-Bench exposes a critical limitation of existing MLLMs: the inability to consistently model long-term, system-level geospatial change.

\section{More details of VLRS-Bench}\label{sec:vlrs_details}
This appendix section presents the statistical specifications of VLRS-Bench and provides comprehensive definitions for the fine-grained L-3 reasoning capabilities.
\subsection{Configuration and Statistics of VLRS-Bench}
\label{vlrs_configuration}
\begin{table*}[!t]
\centering
\caption{Characteristics and vision-language formats of VLRS-Bench}
\label{tab:task_distribution}
\footnotesize
\renewcommand{\arraystretch}{1.2}
\resizebox{\linewidth}{!}{%
\begin{tabular}{@{}lllccccccccc@{}}
\toprule
\textbf{L-1 Tasks} & \textbf{L-2 Tasks} & \textbf{L-3 Tasks} & \textbf{Abbr.} & \textbf{Samples} & \textbf{SC} & \textbf{MC} & \textbf{FB} & \textbf{TF} & \textbf{Avg Q} & \textbf{Avg Img} & \textbf{Max Img} \\
\midrule
\multirow{8}{*}{Cognition} & \multirow{4}{*}{Spatial Cognitive (SC)} & Causal Reasoning & CR & 125 & 31 & 32 & 31 & 31 & 129.4 & 1.00 & 1 \\
& & Counterfactual Reasoning & CFR & 125 & 32 & 31 & 31 & 31 & 134.2 & 1.00 & 1 \\
& & Semantic Integration Reasoning & SIR & 125 & 31 & 31 & 31 & 32 & 127.1 & 1.00 & 1 \\
& & Mechanistic Interaction Reasoning & MIR & 125 & 32 & 31 & 31 & 31 & 128.7 & 1.00 & 1 \\
\cmidrule(l){2-12}
& \multirow{4}{*}{Spatiotemporal Cognitive (ST-C)} & Spatiotemporal Counterfactual Reasoning & ST-CFR & 125 & 33 & 29 & 31 & 32 & 123.2 & 2.00 & 2 \\
& & Spatiotemporal Causal-Chain Reasoning & ST-CCR & 125 & 32 & 30 & 30 & 33 & 129.9 & 2.00 & 2 \\
& & Spatiotemporal Evolution Reasoning & ST-ER & 125 & 34 & 28 & 32 & 31 & 127.3 & 2.00 & 2 \\
& & Spatiotemporal Consistency Reasoning & ST-CR & 125 & 33 & 29 & 31 & 32 & 130.4 & 2.00 & 2 \\
\midrule
\multirow{2}{*}{Decision} & Pre-event Decision (Pre-D) & Planning Reasoning & PR & 250 & 69 & 56 & 63 & 62 & 132.3 & 1.00 & 1 \\
& Post-event Decision (Post-D) & Evaluation Reasoning & ER & 250 & 67 & 58 & 62 & 63 & 133.1 & 1.00 & 1 \\
\midrule
\multirow{4}{*}{Prediction} & \multirow{2}{*}{Object-level Predictive (OP)} & Spatiotemporal Category-State Prediction Reasoning & ST-CS-PR & 125 & 27 & 35 & 32 & 31 & 125.3 & 2.34 & 8 \\
& & Spatiotemporal Morphological Prediction Reasoning & ST-M-PR & 125 & 26 & 37 & 31 & 31 & 136.8 & 2.31 & 8 \\
\cmidrule(l){2-12}
& \multirow{2}{*}{Scene-level Predictive (SP)} & Spatiotemporal Scenario Uncertainty Prediction Reasoning & ST-SU-PR & 125 & 27 & 36 & 32 & 30 & 128.9 & 2.42 & 8 \\
& & Spatiotemporal Sequence Prediction Reasoning & ST-SQ-PR & 125 & 26 & 37 & 32 & 30 & 125.9 & 2.37 & 8 \\
\bottomrule
\multicolumn{12}{l}{\footnotesize SC: Single Choice, MC: Multi Choice, FB: Fill Blank, TF: True/False, Avg Q: mean words in the question, including options and instructions, Avg Img: Avg Image Count, Max Img: Max Image Count}
\end{tabular}
}
\end{table*}

Table~\ref{tab:task_distribution} provides a detailed overview of our proposed VLRS-Bench, the first benchmark specifically designed to evaluate the advanced reasoning capabilities of Multimodal Large Language Models (MLLMs) in the remote sensing domain. The benchmark is meticulously structured into a three-level hierarchy (L-1, L-2, L-3) to comprehensively assess a wide spectrum of MLLMs' abilities.

A key feature of VLRS-Bench is its high linguistic complexity, which is directly linked to the reasoning complexity of the tasks. As shown in the \textquotedblleft Avg Q\textquotedblright{} (Average Question Words) column, the average question length is 129.87 words overall and exceeds 120 words for every task. This deliberate design choice requires MLLMs to process and understand long, detailed contextual information, thereby testing their capacity for complex reasoning rather than simple pattern matching. For instance, tasks like Counterfactual Reasoning (CFR) and Mechanistic Interaction Reasoning (MIR) present intricate scenarios that demand a deep comprehension of the provided text to arrive at a correct conclusion.

Furthermore, VLRS-Bench is uniquely designed to evaluate spatiotemporal reasoning through its multi-temporal questions. The \textquotedblleft Avg Img\textquotedblright{} and \textquotedblleft Max Img\textquotedblright{} columns highlight this characteristic. While foundational tasks like Causal Reasoning (CR) are based on single images, more advanced tasks, particularly in the Prediction category, leverage multi-temporal data. In our benchmark, this is exemplified by tasks such as Spatiotemporal Morphological Prediction (ST-M-PR) and Spatiotemporal Sequence Prediction (ST-SQ-PR), which require models to process up to 8 images (\textquotedblleft Max Img\textquotedblright). This feature compels MLLMs to not only understand individual scenes but also to reason about complex changes, trends, and future states over extended time periods, a critical capability for real-world remote sensing applications.

The table also details the distribution of annotation formats, including Single-Choice (SC), Multi-Choice (MC), Fill in Blank (FB), and True/False (TF). This diversity ensures a multifaceted evaluation, preventing models from overfitting to a single question style. In summary, the combination of high linguistic complexity and multi-temporal data makes VLRS-Bench a robust and comprehensive tool for assessing the advanced reasoning capabilities of MLLMs in the remote sensing domain.

\subsection{More Detailed Definitions of L-3 Capabilities}
\label{L_3_detail}
\noindent\textbf{Spatial Cognitive (SC).} This L-2 dimension focuses on evaluating the model's ability to comprehend the formative causes and intrinsic relationships of geospatial patterns at a single point in time.
\begin{itemize}
    \item \textbf{Causal Reasoning (CR).} This capability evaluates the aptitude for etiological inference. It requires the model to transcend surface-level observables to identify the latent drivers, including natural forces or anthropogenic activities, that precipitated a specific spatial phenomenon. The objective is to determine the provenance of the current state by explaining why a phenomenon emerged from underlying environmental conditions.

    \item \textbf{Counterfactual Reasoning (CFR).} This assesses the capacity for abstract simulation via hypothetical intervention. The model must mentally alter specific antecedent variables to deduce divergent spatial trajectories. This tests the understanding of causal dependencies by examining why an outcome is contingent upon specific conditions, thereby exploring the logical consequences of alternative realities.

    \item \textbf{Mechanistic Interaction Reasoning (MIR).} This focuses on the dynamic coupling of geospatial variables. Unlike identifying a single cause, this requires deciphering the invisible physical or logical feedback loops between elements, such as how topography modulates hydrology. It addresses the process-oriented rationale of why the synergistic interplay of these factors physically precipitates the observed risk or phenomenon.

    \item \textbf{Semantic Integration Reasoning (SIR).} This focuses on the holistic synthesis of regional identity. It requires the model to aggregate disparate low-level visual primitives, including land cover distribution and infrastructure density, into a coherent high-level concept. It addresses the structural rationale of why the specific spatial configuration of these elements collectively constitutes the functional character of the region.
\end{itemize}

\noindent\textbf{Spatiotemporal Cognitive (ST-C).}
This L-2 dimension focuses on evaluating the model's ability to understand how and why geospatial patterns evolve over time.
\begin{itemize}
    \item \textbf{Spatiotemporal Causal-Chain Reasoning (ST-CCR).} This capability evaluates the aptitude for constructing diachronic causal chains. It requires the model to elucidate the logical progression of events, explaining why a specific sequence of temporal antecedents inevitably led to the current spatial state. The objective is to decode the causal lineage of a landscape, linking discrete temporal snapshots into a coherent narrative of evolution.

    \item \textbf{Spatiotemporal Counterfactual Reasoning (ST-CFR).} This assesses the capacity for historical simulation via hypothetical intervention. The model must mentally alter a critical past event to derive a divergent evolutionary trajectory. This tests the understanding of temporal dependencies by examining why the present reality is contingent upon specific historical conditions, thereby exploring the logical consequences of alternative developmental paths.

 \item \textbf{Spatiotemporal Evolution Reasoning (ST-ER).} This evaluates the model’s comprehension of long-term functional metamorphosis over time. It requires the model to interpret the latent semantic shift of a region, explaining why the accumulation of incremental physical changes constitutes a fundamental transition in land-use identity. For instance, the model must deduce why the emergence of infrastructure transforms a region's functional definition from agricultural to residential, rather than merely noting the visual difference.

\item \textbf{Spatiotemporal Consistency Reasoning (ST-CR).} This tests the ability to verify strict temporal coherence through rigorous quantitative logic. Rather than merely identifying change, the model must assess why the observed magnitude or rate of resource evolution aligns or conflicts with expected patterns. This involves validating the logical consistency of the timeline by quantifying key metrics such as loss or gain rates against the inherent physical constraints of the scene.
\end{itemize}

\subsubsection{Decision}
The Decision dimension evaluates the capacity for operational spatial problem-solving. It shifts the focus from passive observation to active strategy, requiring the model to determine \textit{how to execute} specific objectives within complex geospatial environments. We categorize this into generative planning and critical evaluation.

\noindent\textbf{Pre-event Decision (PRE-D).}
This category focuses on the proactive generation of spatial strategies.
\begin{itemize}
    \item \textbf{Planning Reasoning (PR).} This capability tests the aptitude for spatial optimization. The model must formulate an optimal solution to achieve a defined objective, such as site selection or route optimization. This requires synthesizing the visual evidence to propose a plan that satisfies implicit environmental and logistical constraints, effectively answering \textit{how to configure} the space to meet the goal.
\end{itemize}

\noindent\textbf{Post-event Decision (POST-D).}
This category focuses on the retrospective assessment of proposed interventions.
\begin{itemize}
\item \textbf{Evaluation Reasoning (ER).} This capability tests the practical aptitude for rigorous feasibility analysis. The model must scrutinize a proposed spatial plan against the actual physical reality of the scene to validate its overall robustness. It involves identifying potential risks or latent inefficiencies to determine how viable the plan is given the realistic contextual limitations, ensuring consistent alignment between the proposal and the surrounding environment.

\end{itemize}

\subsubsection{Prediction}
The Prediction dimension evaluates the capacity for temporal extrapolation. It requires the model to project historical observations into the unobserved future, forecasting spatiotemporal states at both the discrete object level and the holistic scene level.

\noindent\textbf{Object-level Predictive (OP).}
This category focuses on forecasting the specific evolutionary trajectories of individual entities.
\begin{itemize}
    \item \textbf{Spatiotemporal Category–State Prediction Reasoning (ST-CS-PR).} This evaluates the aptitude for forecasting semantic transitions. The model must infer the developmental trajectory of an entity to determine its future identity, such as predicting the inevitable conversion of an agricultural plot into commercial infrastructure based on visible urbanization pressures.

    \item \textbf{Spatiotemporal Morphological Prediction Reasoning (ST-M-PR).} This tests a model’s fine-grained capability to extrapolate continuous geometric evolution. It requires modeling physical dynamics to forecast progressive changes in shape or extent. For instance, the model must analyze historical erosion rates to project the future morphology of a receding coastline, deriving the shoreline's position five years hence.
\end{itemize}

\noindent\textbf{Scene-level Predictive (SP).}
This category focuses on reasoning about the macro-scale evolution and stochastic nature of entire landscapes.
\begin{itemize}
    \item \textbf{Spatiotemporal Scenario Uncertainty Prediction Reasoning (ST-SU-PR).} This assesses the capacity for probabilistic forecasting amidst uncertainty. Recognizing that geospatial evolution is non-deterministic, the model must delineate multiple plausible trajectories. For example, it might derive divergent futures for a peri-urban zone, contrasting a low-density sprawl scenario against a high-density mixed-use development based on potential planning policies.

    \item \textbf{Spatiotemporal Sequence Prediction Reasoning (ST-SQ-PR).} This evaluates the understanding of temporal coherence and logical succession. The model must determine the next phase in a multi-step evolutionary process, validating whether a proposed future state represents a causally consistent continuation of the observed historical sequence.
\end{itemize}




\section{VLRS-Bench Pipeline Details}\label{sec:pipeline}

\subsection{Data collection}
\textbf{FAIR1M.}
FAIR1M is designed for fine-grained object recognition in high-resolution remote sensing imagery. It integrates data from the Gaofen satellite series and Google Earth, with a spatial resolution ranging from 0.3\,m to 0.8\,m. The dataset contains approximately 15,000 images and more than one million annotated instances spanning 5 major categories and 37 subcategories. All objects are labeled using oriented bounding boxes, and metadata such as geographic coordinates and image resolution are provided. Due to its high intra-class variability, wide geographic coverage, large scale changes, and complex backgrounds, FAIR1M is considered substantially more challenging than previous public datasets, offering a realistic and demanding benchmark for remote sensing object detection.

\textbf{DIOR.}
DIOR is a large-scale optical remote sensing benchmark for object detection, covering more than 80 countries. It includes 23,463 images and around 192,472 object instances belonging to 20 categories. The dataset explicitly accounts for variations in scale, imaging conditions (weather, season, viewpoint), object morphology, and background complexity, highlighting challenges such as intra-class diversity and semantic ambiguity between similar categories (\eg \textquotedblleft bridge\textquotedblright{} or \textquotedblleft overpass\textquotedblright). Although DIOR mainly provides RGB imagery without infrared or DSM channels, its rich diversity and substantial scale variations make it an important benchmark for evaluating detection algorithms in complex environments.

\textbf{LoveDA.}
LoveDA focuses on semantic segmentation and domain adaptation in high-resolution remote sensing imagery. It consists of images collected from three Chinese cities, covering both urban and rural scenes, with a spatial resolution of approximately 0.3\,m. The dataset includes 5,987 images and about 166,768 annotated objects across seven semantic classes: building, road, water, farmland, forest, barren, and background. LoveDA explicitly targets domain discrepancies between urban and rural environments, making it highly suitable for evaluating semantic segmentation and cross-domain learning methods. No infrared or DSM modalities are specified in the released data.

\textbf{Potsdam.}
The Potsdam dataset, released as part of the ISPRS \textquotedblleft 2D Semantic Labeling Contest\textquotedblright, contains high resolution urban imagery with original image sizes of $6000\times6000$ pixels. It includes RGB, infrared (IR), and DSM (digital surface model) channels, with a typical ground sampling distance of 5\,cm. The dataset provides pixel-level annotations for six semantic classes including building, low vegetation, tree, road, car, and clutter. Owing to its extremely high resolution and multi-modal characteristics, Potsdam is widely used for semantic segmentation and urban change detection tasks.

\textbf{Vaihingen.}
Vaihingen is another ISPRS benchmark dataset containing high-resolution orthophotos of the Vaihingen region in Germany, accompanied by DSM and infrared imagery. The average image size is approximately 2494$\times$2064 pixels, with a native ground sampling distance of 0.09\,m. The annotated classes match those in the Potsdam dataset, including building, low vegetation, tree, road, car, and clutter. The dataset is recognized for its challenging scenes characterized by complex backgrounds, occlusions, and strong shadows, making it an important benchmark for advanced semantic segmentation methods.

\textbf{GID-15.}
The Gaofen Image Dataset with 15 categories is primarily constructed from GF-2 satellite imagery. It comprises around 150 large-scale images with 15 annotated land-cover categories. The GF-2 sensor provides 1\,m panchromatic and 4\,m multispectral imagery. GID-15 is widely adopted in land-cover classification, semantic segmentation, and cross-region generalization research, serving as a major resource in the remote sensing community.

\textbf{miniUCD.}
miniUCD is a subset of the Hi-UCD series designed for urban semantic change detection. It is based on ultra–high-resolution (0.1\,m) aerial orthophotos from Tallinn, Estonia, covering three temporal snapshots (2017, 2018, 2019). Images are cropped into 1024$\times$1024 paired patches, with pixel-level annotations for nine land-cover classes and semantic change types (\eg building, road, bare soil, vegetation). Emphasizing fine-grained semantics, multi-temporal observations, and detailed change annotations, miniUCD provides a rigorous benchmark for evaluating semantic change detection models in complex urban environments.

\textbf{SpaceNet7.}
SpaceNet7, also known as the Multi-Temporal Urban Development (MUDS) dataset, focuses on building footprint detection, tracking, and temporal change analysis using satellite image time series. It contains monthly composite imagery over 24 months, covering more than 100 geographic regions and approximately 40,000\,km$^2$. The dataset includes over 11 million building annotations, each assigned a unique identifier to facilitate long-term tracking. Tasks supported by SpaceNet7 include building detection, temporal tracking, and construction/demolition change analysis. Its temporal continuity, large geographic coverage, and object-level tracking make it a key dataset for studying urbanization and spatiotemporal remote sensing.

\textbf{xView2.}
xView2 focuses on post-disaster building damage assessment, developed from the xView and xBD initiatives. It uses high-resolution satellite imagery (\eg WorldView-3 with $\sim$0.3\,m GSD) and provides paired pre-disaster and post-disaster images, multi-level building damage labels (0--3), and polygon-based building annotations. The combination of fine spatial resolution, detailed damage categorization, and diverse disaster scenarios makes xView2 an essential benchmark for change detection and automated disaster response.

\textbf{SECOND.}
The Semantic Change Detection Dataset (SECOND) is a large-scale benchmark specifically developed for semantic change detection in high-resolution remote sensing imagery. Released by Wuhan University, it contains approximately 4,662 pairs of multi-temporal aerial/remote sensing images covering urban regions such as Hangzhou, Chengdu, and Shanghai. Each image pair has a size of 512$\times$512 pixels and provides pixel-level semantic change labels covering classes such as water, bare soil, low vegetation, tree, building, and playground. SECOND targets not only binary change detection but also \textquotedblleft from-to\textquotedblright{} semantic transitions, enabling fine-grained modeling of land-cover evolution. Its clear category design, detailed annotations, and complex urban scenes make it a strong benchmark for advancing semantic change detection research.

\subsection{Prior Images}

\begin{table*}[!t]
\centering
\caption{Distribution of reasoning tasks across different datasets in VLRS-Bench.}
\label{tab:dataset_distribution}
\footnotesize
\resizebox{\linewidth}{!}{%
\begin{tabular}{@{}l|cc|cccccccc|cc|cccc|c@{}}
\toprule
\multirow{2}{*}{\textbf{Dataset}} & \multirow{2}{*}{\textbf{Prior}} & \multirow{2}{*}{\begin{tabular}[c]{@{}c@{}}\textbf{Max. observed} \\ \textbf{frames}\end{tabular}} & \multicolumn{8}{c|}{\textbf{Cognition}} & \multicolumn{2}{c|}{\textbf{Decision}} & \multicolumn{4}{c|}{\textbf{Prediction}} & \multirow{2}{*}{\textbf{Total}} \\
\cmidrule(lr){4-11} \cmidrule(lr){12-13} \cmidrule(lr){14-17}
& & & \textbf{CR} & \textbf{CFR} & \textbf{SIR} & \textbf{MIR} & \textbf{ST-CFR} & \textbf{ST-CCR} & \textbf{ST-ER} & \textbf{ST-CR} & \textbf{PR} & \textbf{ER} & \textbf{ST-CS-PR} & \textbf{ST-M-PR} & \textbf{ST-SU-PR} & \textbf{ST-SQ-PR} & \\
\midrule
FAIR1M & Box2Mask & 1 & 26 & 8 & 33 & 14 & 0 & 0 & 0 & 0 & 41 & 23 & 0 & 0 & 0 & 0 & 145 \\
DIOR & Box2Mask & 1 & 14 & 18 & 35 & 27 & 0 & 0 & 0 & 0 & 74 & 66 & 0 & 0 & 0 & 0 & 234 \\
\midrule
LoveDA & Mask & 1 & 18 & 19 & 20 & 14 & 0 & 0 & 0 & 0 & 29 & 28 & 0 & 0 & 0 & 0 & 128 \\
Potsdam & Mask, DSM & 1 & 17 & 15 & 14 & 28 & 0 & 0 & 0 & 0 & 40 & 58 & 0 & 0 & 0 & 0 & 172 \\
Vaihingen & NIR, Mask, DSM & 1 & 11 & 10 & 1 & 9 & 0 & 0 & 0 & 0 & 8 & 11 & 0 & 0 & 0 & 0 & 50 \\
GID-15 & Mask & 1 & 20 & 24 & 22 & 33 & 0 & 0 & 0 & 0 & 58 & 64 & 0 & 0 & 0 & 0 & 221 \\
\midrule
xView2 & Mask, temp-img & 2 & 11 & 19 & 0 & 0 & 12 & 37 & 36 & 34 & 0 & 0 & 0 & 0 & 0 & 0 & 149 \\
SECOND & Mask, temp-img & 2 & 8 & 12 & 0 & 0 & 9 & 42 & 46 & 25 & 0 & 0 & 0 & 0 & 0 & 0 & 142 \\
\midrule
miniUCD & Mask, temp-img & 3 & 0 & 0 & 0 & 0 & 0 & 0 & 0 & 0 & 0 & 0 & 23 & 0 & 0 & 0 & 23 \\
SpaceNet7 & Mask, temp-img & 8 & 0 & 0 & 0 & 0 & 104 & 46 & 43 & 66 & 0 & 0 & 102 & 125 & 125 & 125 & 736 \\
\midrule
\textbf{Total} & - & - & \textbf{125} & \textbf{125} & \textbf{125} & \textbf{125} & \textbf{125} & \textbf{125} & \textbf{125} & \textbf{125} & \textbf{250} & \textbf{250} & \textbf{125} & \textbf{125} & \textbf{125} & \textbf{125} & \textbf{2000} \\
\bottomrule
\end{tabular}
}
\end{table*}

A critical stage in the VLRS-Bench generation pipeline is the integration of Prior Images. This step is designed to systematically elevate the complexity of the generated tasks beyond standard visual perception. By treating these priors as privileged information, our pipeline constructs a comprehensive multimodal instruction that provides the MLLMs with the necessary multi-dimensional context to address the advanced reasoning tasks defined in our benchmark. 

The pipeline purposefully integrates three distinct categories of prior images, each serving as a mechanism to generate questions targeting specific cognitive, decision-making, and predictive abilities:

\begin{enumerate}
    \item \textbf{Multi-source Remote Sensing Priors.} The pipeline automatically ingests multi-source data such as Digital Surface Models (DSM) and Near-Infrared (NIR) imagery from datasets like Potsdam and Vaihingen (see Table~\ref{tab:dataset_distribution}). This integration is a deliberate step to generate questions for \textbf{Spatial Cognitive (SC)} tasks. For instance, by providing a DSM, the pipeline can formulate a Causal Reasoning (CR) question that requires the model to infer a building's collapse was caused by its location on a steep slope. Similarly, this data allows the pipeline to construct \textbf{Decision} tasks like Planning Reasoning (PR), where a question might ask the model to identify suitable locations for new infrastructure based on ground elevation.

    \item \textbf{Expert Pixel-Level Priors.} A unique procedure within our pipeline is the strategic use of expert-annotated masks as input conditions rather than as prediction targets. This procedural choice is specifically designed to generate high-level \textbf{Decision} tasks. For example, the pipeline can generate a Planning Reasoning (PR) task by providing a risk-zone mask and asking the model to chart an evacuation route that avoids these areas. It can also construct an Evaluation Reasoning (ER) question by supplying a \textquotedblleft protected zone\textquotedblright{} mask and prompting the model to verify post-event policy compliance. This step in the pipeline transforms the nature of the generated task from simple recognition to constrained, goal-oriented problem-solving.

    \item \textbf{Multi-temporal Reference Priors.} For spatio-temporal analysis, the pipeline incorporates reference images from different time points (denoted as temp-img) from datasets like xView2 and SpaceNet7. These priors function as \textquotedblleft causal anchors\textquotedblright{} within the generated instruction. \textbf{Crucially, for each temporal image, the pipeline embeds a precise timestamp (accurate to the month) into the instruction.} This grounds the reasoning tasks in physical reality, compelling the model to move beyond simple sequential ordering. It must now consider plausible rates of change—for example, distinguishing between rapid post-disaster reconstruction and slow-paced urban growth over several years. This ensures that the generated \textbf{Spatiotemporal Cognitive (ST-C)} and \textbf{Prediction} tasks adhere to plausible real-world dynamics. This pipeline step is the explicit mechanism for generating all such tasks, from Spatiotemporal Causal-Chain Reasoning (ST-CCR) to Spatiotemporal Sequence Prediction Reasoning (ST-SQ-PR).
\end{enumerate}

By orchestrating the integration of these three prior types, our pipeline systematically constructs a sophisticated evaluation landscape. This methodological step ensures that VLRS-Bench moves beyond assessing simple pattern recognition to evaluating a model's capacity for true cognitive synthesis: analyzing multi-source geospatial data, making decisions under expert constraints, and comprehending complex spatio-temporal dynamics that are grounded in the physical laws of the real world.

\subsection{Observed Images}
\label{sec:observed_images}


\begin{table*}[!t]
\centering
\caption{Summary of datasets comprising the visual foundation of VLRS-Bench. Image size refers to the benchmark images. Ground sampling distance refers to the source imagery.}
\label{tab:dataset_summary}
\resizebox{\linewidth}{!}{%
\begin{tabular}{@{}llcll@{}}
\toprule
\textbf{Dataset} & \textbf{Source Type} & \textbf{Image Size} & \textbf{Classes} & \textbf{Native GSD} \\
\midrule
FAIR1M & \multirow{2}{*}{Object Detection} & 600$\times$600 & 37 (\eg Airplane, Baseball field, Bridge, ...) & 0.3 m to 0.8 m \\
DIOR & & 800$\times$800 & 20 (\eg airplane, airport, harbor, ...) & 0.5 m to 30 m \\
\midrule
LoveDA & \multirow{4}{*}{Semantic Segmentation} & 1024$\times$1024 & 7 (\eg Buildings, Roads, Water bodies, ...) & 0.3 m \\
Potsdam & & 1024$\times$1024 & 6 (\eg Impervious surfaces, Buildings, Trees, ...) & 0.05 m \\
Vaihingen & & 1024$\times$1024 & 6 (\eg Building roofs, Low vegetation, Cars, ...) & 0.09 m \\
GID-15 & & 1024$\times$1024 & 15 (\eg Industrial land, Paddy field, River, ...) & 1 m pan / 4 m multispectral \\
\midrule
xView2 & \multirow{4}{*}{Change Detection} & 1024$\times$1024 & 4 (\eg No damage, Minor damage, Destroyed, ...) & $\leq$ 0.8 m \\
SECOND & & 512$\times$512 & 7 (\eg Non-change, Low vegetation, Building, ...) & -- \\
miniUCD & & 1024$\times$1024 & 9 land-cover classes plus no change & 0.1 m \\
SpaceNet7 & & 1024$\times$1024 & 1 (Building Footprints) & 4.0 m \\
\bottomrule
\end{tabular}%
}
\end{table*}

The Observed Image serves as the primary visual anchor for every task within the VLRS-Bench pipeline. It is the original RGB image that provides the foundational visual information upon which all subsequent reasoning is built. To ensure our benchmark is comprehensive and robust, we have curated a highly diverse collection of observed images from a wide array of public remote sensing datasets, as summarized in Table~\ref{tab:dataset_summary}.

This collection is characterized by its multi-scale nature and semantic richness. The ground resolution spans from very-high-resolution (VHR) imagery, such as miniUCD (0.1 m) and Potsdam (0.05 m), which allows for the identification of fine-grained details like individual cars, to medium-resolution data like DIOR (up to 30 m), suitable for analyzing large-scale land-use patterns. This vast range in scale ensures that VLRS-Bench can evaluate a model's reasoning capabilities across different levels of abstraction.

Furthermore, the semantic diversity is extensive, drawing from both object detection and semantic segmentation datasets. Datasets like FAIR1M and DIOR provide a rich vocabulary of specific object categories (\eg \textquotedblleft Airplane,\textquotedblright{} \textquotedblleft Storage tank\textquotedblright), enabling tasks that require fine-grained recognition and spatial relationship analysis. In contrast, datasets such as LoveDA and GID-15 offer broad land-cover categories (\eg \textquotedblleft Forests,\textquotedblright{} \textquotedblleft Industrial land,\textquotedblright{} \textquotedblleft Paddy field\textquotedblright), facilitating reasoning about complex scenes, ecological systems, and urban functionality. The scenes themselves are equally varied, encompassing dense urban centers (Potsdam), rural and agricultural landscapes (LoveDA), and specialized infrastructure like airports and harbors (DIOR).

By constructing this varied and challenging visual foundation, we ensure that the Observed Images provide a robust starting point for the generation of complex, multi-faceted reasoning problems across all dimensions of VLRS-Bench.

\subsection{Mask Info}

\begin{figure}[!htbp]
    \centering
    \includegraphics[width=\textwidth]{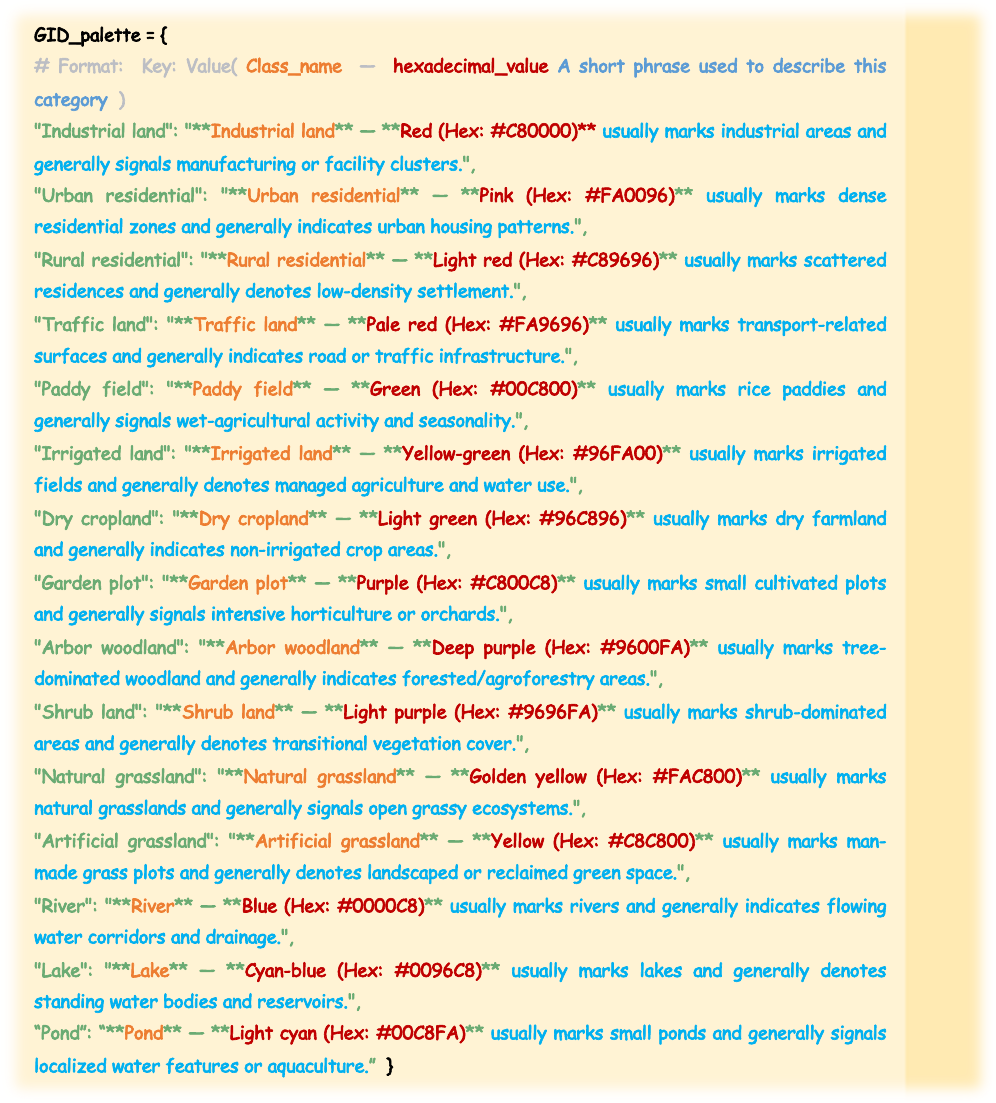}
    \caption{An illustration of the Mask\_Info stage using the GID-15 dataset's mask palette as an example. This process acts as a \textquotedblleft semantic bridge,\textquotedblright{} transforming a raw segmentation mask into a structured, semantically rich instruction. For each class in the GID-15 palette, the pipeline maps the pixel color to a standardized hexadecimal code and appends a functional description, bridging the semantic gap and enabling expert-level reasoning about land use.}
    \label{fig:semantic_bridge}
\end{figure}

A fundamental challenge in applying Multimodal Large Language Models (MLLMs) to specialized domains like remote sensing is their inability to comprehend the intrinsic meaning of expert annotations. A segmentation mask, to an MLLM, is merely a collection of colored pixels. To address this, as illustrated in Figure~\ref{fig:semantic_bridge}, our pipeline incorporates a novel Mask Info stage, which acts as a powerful semantic bridge, transforming inert pixel data into a rich, interpretable layer of domain knowledge. This step is a cornerstone of our methodology and represents a significant departure from previous approaches.

The figure depicts how this stage deconstructs a raw segmentation mask and reassembles it as a structured, multi-part semantic dictionary. The design of this dictionary, shown in the central part of Figure~\ref{fig:semantic_bridge}, is our key innovation:

\begin{enumerate}
    \item \textbf{Semantic Bridge.} The process first establishes a direct link between a class label (\eg \textquotedblleft Industrial land\textquotedblright) and its precise visual representation using a standardized hexadecimal color code (\eg \textquotedblleft \#C80000\textquotedblright). This provides a consistent, machine-readable anchor to the image pixels.

    \item \textbf{Category Semantic Mapping.} This is a crucial design element. The pipeline does not stop at generating simple category labels; instead, it further appends concise explanations of the functional significance of each category as well as its typical practical implications in real-world scenarios. This descriptive content is crafted to provide the model with a context-aware, expert-level cognitive perspective.
\end{enumerate}

The innovation, visually captured in the figure, lies in shifting the paradigm from simple labeling to in-context knowledge injection. Prior work typically uses masks as ground truth for segmentation tasks or employs simplistic prompts like \textquotedblleft the red area is a building.\textquotedblright{} Such methods only test the model's ability to associate a shape with a name. Our approach, by contrast, compels the model to understand the implications of a category. It teaches the model not just what something is, but what it means in a broader context.

Figure~\ref{fig:semantic_bridge} provides a concrete example using the GID-15 dataset. The color Red (Hex: \textquotedblleft \#C80000\textquotedblright) is not merely labeled \textquotedblleft Industrial land\textquotedblright; the pipeline enriches it with the functional context: \textquotedblleft Industrial land... usually marks industrial areas and generally signals manufacturing or facility clusters.\textquotedblright{} This enrichment, central to the process shown in the figure, equips the MLLMs to answer complex causal reasoning questions, such as: \textquotedblleft Given the proximity of the industrial land (red) to the river (blue), what is a potential environmental risk that should be monitored?\textquotedblright

This is a level of reasoning that is impossible without the functional semantic context provided by our Mask Info stage. By systematically converting abstract color codes into a rich, interpretable knowledge base, this step transforms expert priors from passive constraints into active components of a complex reasoning problem. It is this mechanism that fundamentally enables VLRS-Bench to evaluate the deep, domain-specific cognitive abilities that are essential for real-world remote sensing applications.

\subsection{Dataset Info}

\begin{figure}[!t]
    \centering
    \includegraphics[width=0.5\linewidth]{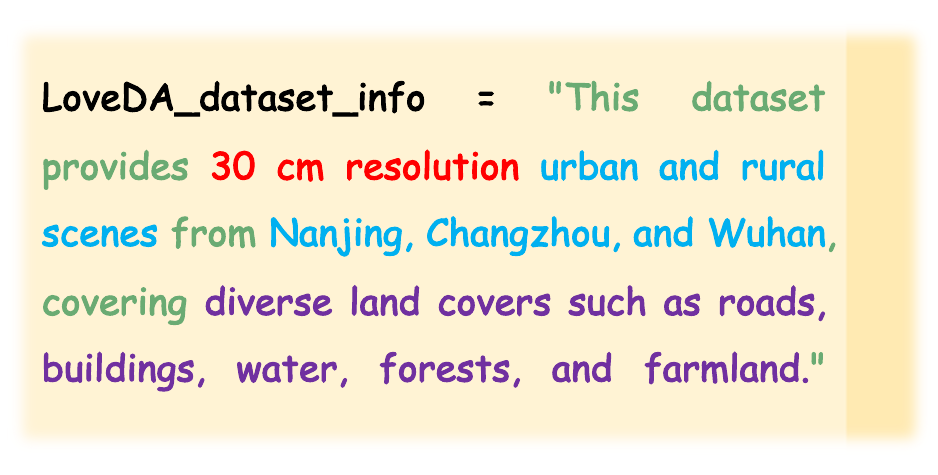}
    \caption{An example of the Dataset Info component for the LoveDA dataset. This text provides the MLLMs with high-level contextual information, including geographic origin, ground resolution, and typical scene types. This meta-level knowledge is crucial for enabling more plausible and context-aware reasoning.}
    \label{fig:dataset_info}
\end{figure}

To provide essential context for the visual data, our pipeline incorporates a Dataset Info component. An MLLM typically processes an image in isolation, unaware of its real-world origin or technical specifications, information that is fundamental to any professional analysis. The Dataset Info stage is designed to inject this crucial metadata directly into the instruction, thereby simulating a more realistic analytical scenario.

As illustrated in Figure~\ref{fig:dataset_info}, the pipeline appends a concise, descriptive summary for each source dataset. This summary provides the MLLMs with a multi-layered contextual scaffold. For instance, the example for the LoveDA dataset shown in the figure informs the model about the image's \textquotedblleft 30 cm resolution,\textquotedblright{} its geographic origin in \textquotedblleft Nanjing, Changzhou, and Wuhan,\textquotedblright{} and the typical land covers it contains. This information allows the MLLMs to make more plausible, context-aware inferences by setting its expectations regarding scale, regional characteristics, and semantic content.

By providing this high-level metadata, the Dataset Info component prevents the model from making ungrounded or generic assumptions. It ensures that the reasoning tasks are not performed in a vacuum but are instead anchored in a specific geographic and technical context, enabling a more realistic and challenging evaluation of the MLLMs' geospatial reasoning capabilities.

\subsection{Task Special Prompt}
\begin{table*}[!t]
\centering
\caption{Temporal frames and auxiliary priors used to construct VLRS-Bench reasoning tasks. The reference frame (marked \textquotedblleft Withheld\textquotedblright) anchors the causal sequence or establishes the outcome, while expert masks and auxiliary DSM and NIR imagery enrich the context for question design.}
\label{tab:task_data_mapping_final}
\resizebox{\linewidth}{!}{%
\begin{tabular}{@{}llccccccccp{4.5cm}@{}}
\toprule
\multirow{2}{*}{\textbf{Category}} & \multirow{2}{*}{\textbf{Reasoning Task}} & \multirow{2}{*}{\textbf{Abbr.}} & \multicolumn{3}{c}{\textbf{Source Frames in Construction Template}} & \multicolumn{4}{c}{\textbf{Auxiliary Construction Priors}} & \multirow{2}{*}{\textbf{Applicable Dataset Category}} \\
\cmidrule(lr){4-6} \cmidrule(lr){7-10}
& & & \textbf{Single (1 img)} & \textbf{Dual (2 imgs)} & \textbf{Tri/Multi ($\geq$ 3 imgs)} & \textbf{Box2Mask} & \textbf{MASK} & \textbf{DSM} & \textbf{NIR} & \\
\midrule
\multirow{8}{*}{\textbf{Cognition}} & Causal Reasoning & CR & \checkmark & \checkmark (Withheld: $im_1$) & \ding{55} & \checkmark & \checkmark & \checkmark & \checkmark & Single: All; Dual: All \\
& Counterfactual Reasoning & CFR & \checkmark & \checkmark (Withheld: $im_1$) & \ding{55} & \checkmark & \checkmark & \checkmark & \checkmark & Single: All; Dual: All \\
& Semantic Integration Reasoning & SIR & \checkmark & \ding{55} & \ding{55} & \checkmark & \checkmark & \checkmark & \checkmark & Single: All \\
& Mechanistic Interaction Reasoning & MIR & \checkmark & \ding{55} & \ding{55} & \checkmark & \checkmark & \checkmark & \checkmark & Single: All \\
\cmidrule(l){2-11} 
& Spatiotemporal Causal-Chain Reasoning & ST-CCR & \ding{55} & \ding{55} & \checkmark (Withheld: $im_2$) & \ding{55} & \checkmark & \ding{55} & \ding{55} & All multi-temporal datasets \\
& Spatiotemporal Counterfactual Reasoning & ST-CFR & \ding{55} & \ding{55} & \checkmark (Withheld: $im_2$) & \ding{55} & \checkmark & \ding{55} & \ding{55} & All multi-temporal datasets \\
& Spatiotemporal Evolution Reasoning & ST-ER & \ding{55} & \ding{55} & \checkmark (Withheld: $im_2$) & \ding{55} & \checkmark & \ding{55} & \ding{55} & All multi-temporal datasets \\
& Spatiotemporal Consistency Reasoning & ST-CR & \ding{55} & \ding{55} & \checkmark (Withheld: $im_2$) & \ding{55} & \checkmark & \ding{55} & \ding{55} & All multi-temporal datasets \\
\midrule
\multirow{2}{*}{\textbf{Decision}} & Planning Reasoning & PR & \checkmark & \ding{55} & \ding{55} & \checkmark & \checkmark & \checkmark & \checkmark & Single: All \\
& Evaluation Reasoning & ER & \checkmark & \ding{55} & \ding{55} & \checkmark & \checkmark & \checkmark & \checkmark & Single: All \\
\midrule
\multirow{4}{*}{\textbf{Prediction}} & Spatiotemporal Category–State Prediction & ST-CS-PR & \ding{55} & \ding{55} & \checkmark (Withheld: $im_{-1}$) & \ding{55} & \checkmark & \ding{55} & \ding{55} & All multi-temporal datasets \\
& Spatiotemporal Morphological Prediction & ST-M-PR & \ding{55} & \ding{55} & \checkmark (Withheld: $im_{-1}$) & \ding{55} & \checkmark & \ding{55} & \ding{55} & All multi-temporal datasets \\
& Spatiotemporal Scenario Uncertainty Prediction & ST-SU-PR & \ding{55} & \ding{55} & \checkmark (Withheld: $im_{-1}$) & \ding{55} & \checkmark & \ding{55} & \ding{55} & All multi-temporal datasets \\
& Spatiotemporal Sequence Prediction & ST-SQ-PR & \ding{55} & \ding{55} & \checkmark (Withheld: $im_{-1}$) & \ding{55} & \checkmark & \ding{55} & \ding{55} & All multi-temporal datasets \\
\bottomrule
\end{tabular}%
}
\end{table*}

Having established the specific composition of data and priors for each reasoning task in Table~\ref{tab:task_data_mapping_final}, the final step in our instruction assembly is to translate this structural template into an executable directive for the MLLMs. This is the role of the Task Special Prompt. Therefore, for each subtask in VLRS-Bench, we engineer a highly-constrained prompt that meticulously guides the model's generation process. This prompt is multi-faceted, imposing several layers of constraints to ensure the output aligns perfectly with our evaluation objectives, focusing on the desired format, question type, and target capability, as detailed in the following subsections.

\begin{figure}[!htbp]
    \centering
    \includegraphics[width=\textwidth]{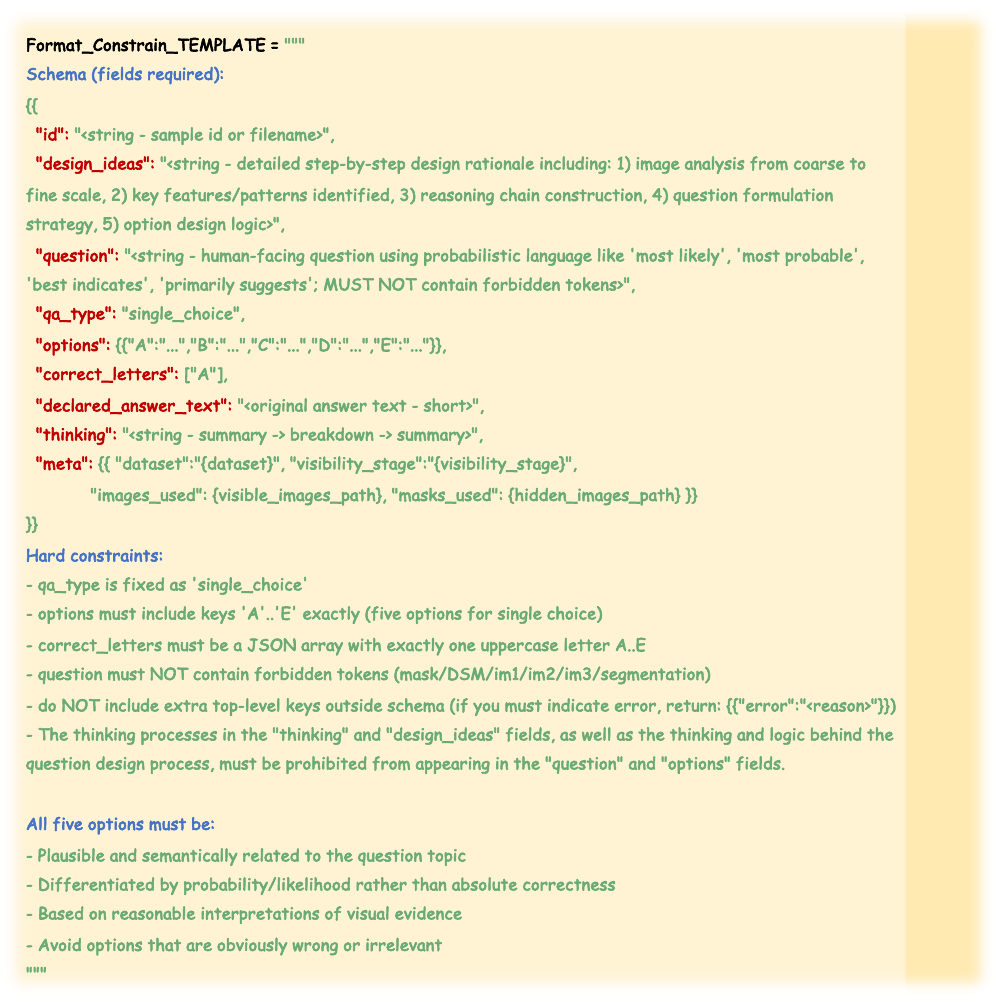}
    \caption{The template for the Format-Constrained Prompt. This template enforces a strict JSON schema and a set of validation rules, ensuring that all generated evaluation items are structurally consistent, machine-readable, and adhere to predefined quality standards.}
    \label{fig:format_prompt}
\end{figure}

\subsubsection{Format Constrained Prompt}

The first and most fundamental layer of constraint imposed by the Task Special Prompt is the Format-Constrained Prompt. Its primary function is to enforce a rigid output structure, ensuring that every generated item is standardized, machine-readable, and contains all necessary metadata for evaluation and analysis. As detailed in the template shown in Figure~\ref{fig:format_prompt}, this is achieved through a comprehensive JSON schema that specifies all required fields, from the question text to the internal design rationale.

Crucially, the schema enforces a strict separation of concerns. It mandates fields like \textquotedblleft design ideas\textquotedblright{} and \textquotedblleft thinking\textquotedblright{} to capture the generative model's reasoning process, while explicitly prohibiting this internal logic from appearing in the user-facing \textquotedblleft question\textquotedblright{} and \textquotedblleft options\textquotedblright. This prevents the model from creating \textquotedblleft shortcut\textquotedblright{} questions that are easily solved by pattern matching. Furthermore, the hard constraints explicitly forbid the leakage of privileged information into the final question by blacklisting tokens such as \textquotedblleft mask\textquotedblright, \textquotedblleft DSM\textquotedblright, or \textquotedblleft im1\textquotedblright. The prompt also governs the qualitative aspects of the multiple-choice options, mandating that all distractors be plausible and semantically related to the topic.

By imposing this rigorous format, the Format-Constrained Prompt acts as the foundational quality gate in our pipeline. It guarantees that the output is structurally sound and that the generated questions are self-contained, fairly testing MLLM reasoning abilities without providing unintended clues.

\begin{figure*}[!t]
    \centering
    \includegraphics[width=\textwidth]{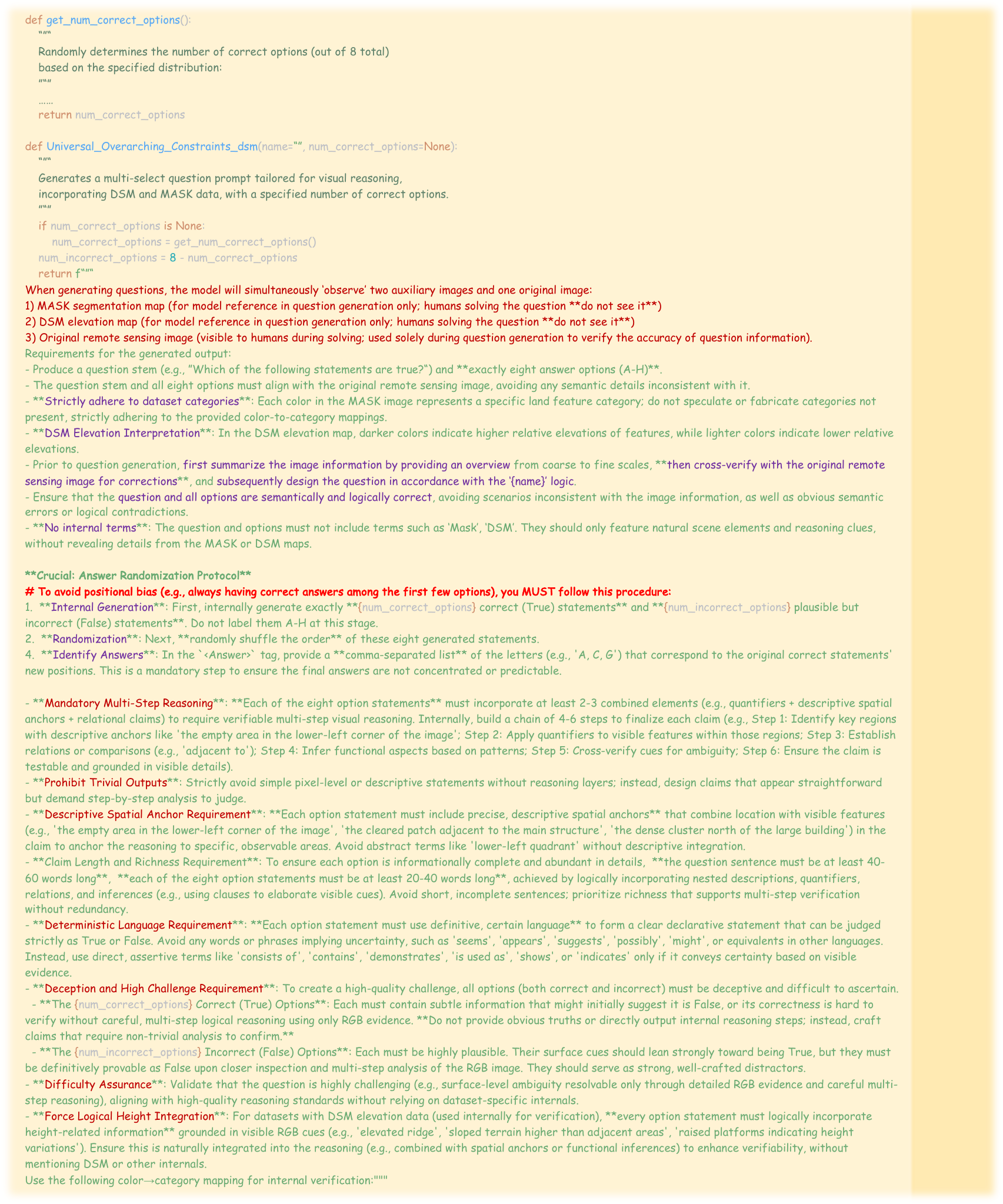}
    \caption{An example of the Reasoning Overarching Constraints, specifically the template for a multi-select question utilizing DSM as a prior. This prompt layer governs the substantive quality of the generated content by enforcing rules for reasoning depth, mitigating MLLMs' biases through a mandatory randomization protocol, and ensuring a high degree of challenge.}
    \label{fig:universal_constraints}
\end{figure*}

\subsubsection{Reasoning Overarching Constraints Prompt}

Beyond the structural integrity enforced by the Format Constrained Prompt, we introduce a more sophisticated layer of control: the Reasoning Overarching Constraints. This prompt layer is not about the output schema, but about the intrinsic quality and cognitive complexity of the reasoning task itself. As exemplified by the multi-select prompt for DSM-enabled datasets shown in Figure~\ref{fig:universal_constraints}, its primary purpose is to guide the generative model to produce questions that are deep, challenging, and test specific reasoning pathways.
The core of this prompt is dedicated to designing for reasoning depth. It mandates that every option be a complex declarative statement requiring multi-step verification. This is achieved by enforcing the use of \textquotedblleft Descriptive Spatial Anchors\textquotedblright{} (\eg \textquotedblleft the cleared patch adjacent to the main structure\textquotedblright) to ground the reasoning in specific image regions, and \textquotedblleft Deterministic Language\textquotedblright{} to eliminate ambiguity, making each option definitively true or false. For datasets with auxiliary data, the prompt even forces the logical integration of concepts from privileged information (such as height from a DSM) into an RGB-solvable question, compelling the model to translate abstract data into visual cues (\eg shadows, perspective) and thus increasing the reasoning complexity.

Furthermore, a critical function of these constraints is to ensure the robustness of the evaluation by actively mitigating known MLLMs' biases. We observed that models often exhibit positional preference, favoring earlier options (\eg A, B, C) when selecting from a large set. To counteract this, the prompt enforces a strict \textquotedblleft Answer Randomization Protocol\textquotedblright: a predefined number of correct and incorrect statements are generated internally and then shuffled before being assigned letters. This procedure ensures that the correct answers are distributed randomly, preventing models from succeeding based on lazy heuristics rather than genuine reasoning. Furthermore, the prompt's \textquotedblleft Deception and High Challenge Requirement\textquotedblright{} ensures that all options are plausible and difficult, forcing a thorough analysis for both true and false statements.
In essence, the Reasoning Overarching Constraints act as a cognitive blueprint, compelling the generative model to function less like a data labeler and more like an expert question designer. This ensures the high difficulty and validity of the VLRS-Bench evaluation items by simultaneously demanding deep reasoning and neutralizing predictable model behaviors.

\subsubsection{Temporal Prior Prompt}

\begin{figure*}[!t]
    \centering
    \includegraphics[width=\textwidth]{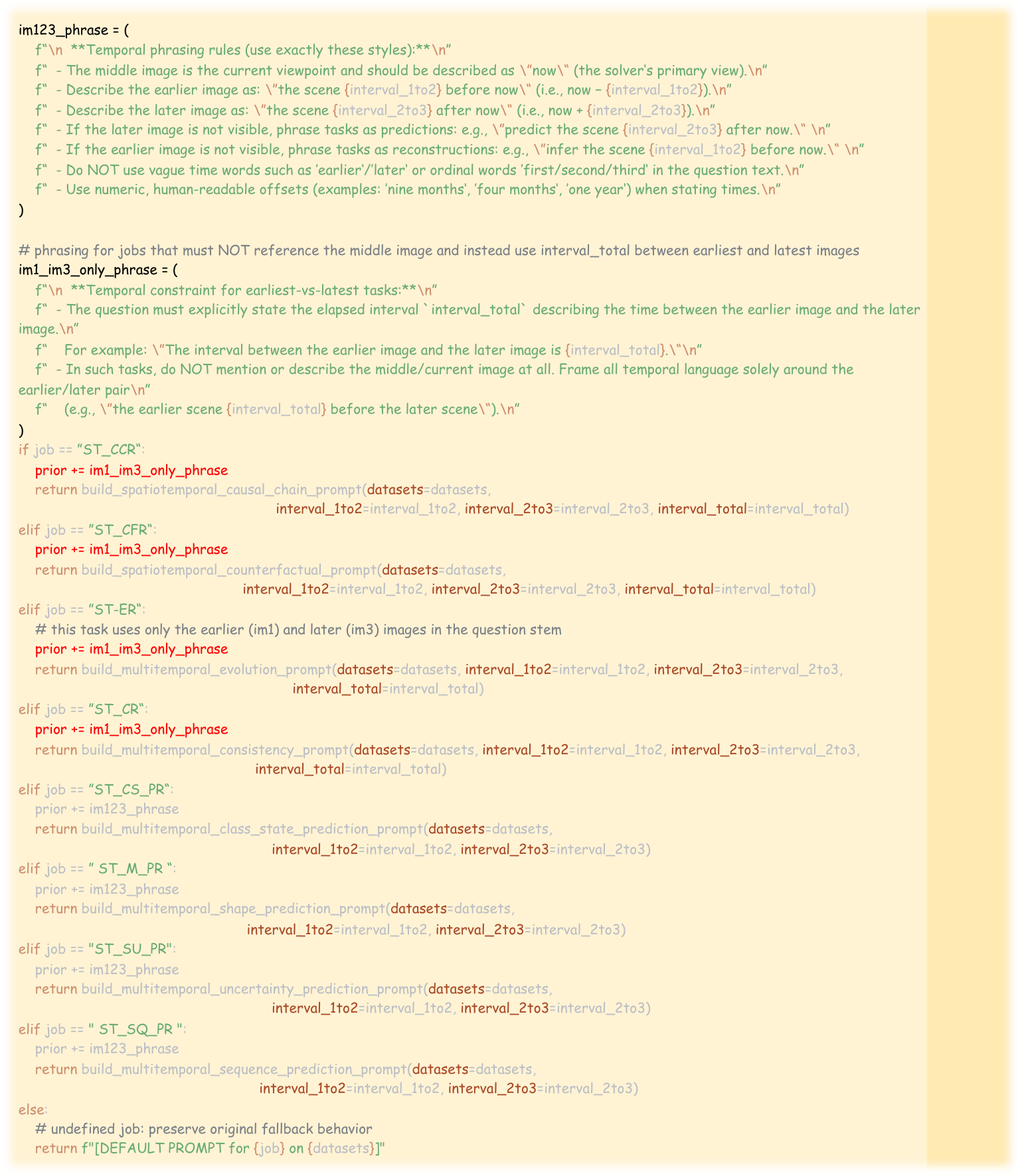}
     \caption{An example of the Temporal Prior Prompt, demonstrating how three-phase temporal data is used to construct diverse reasoning tasks. The code illustrates how different images are strategically employed as hidden reference priors. By dynamically selecting phrasing rules, this mechanism significantly enhances the logical realism and reasoning depth of the generated questions.}
    \label{fig:temporal_prompt}
\end{figure*}

To operationalize the diverse temporal data structures defined in Table~\ref{tab:task_data_mapping_final}, our pipeline employs a dynamic Temporal Prior Prompt. This component translates the abstract concept of an image reference prior (\eg \textquotedblleft Withheld: $im_2$\textquotedblright) into a concrete, linguistically framed reasoning problem. As shown in Figure~\ref{fig:temporal_prompt}, this is not a single static prompt, but a rule-based system that constructs a specific temporal narrative tailored to the target reasoning task.

The system uses two primary phrasing strategies to control the generation model's temporal perspective. For tasks like Spatiotemporal Prediction (\textquotedblleft ST-CS-PR\textquotedblright, \textquotedblleft ST-M-PR\textquotedblright, \etc), it uses a \textquotedblleft relative-to-now\textquotedblright{} framing (\textquotedblleft im123\_phrase\textquotedblright). This establishes the middle image as the present viewpoint (\textquotedblleft now\textquotedblright) and describes the past and future with precise, human-readable offsets (\eg \textquotedblleft the scene nine months before now\textquotedblright). This structure forces the model to reason forward or backward from a known anchor point.

In contrast, for tasks requiring reasoning about a hidden intermediate state, such as Spatiotemporal Causal-Chain Reasoning (\textquotedblleft ST-CCR\textquotedblright), the prompt uses an \textquotedblleft earliest-latest\textquotedblright{} framing (\lq im1\_im3\_only\_phrase\rq). This strategy deliberately omits any mention of the middle image, forcing the model to reason about the causal gap between the first and last visible images across a total elapsed interval. This directly implements the \textquotedblleft Withheld: $im_2$\textquotedblright{} logic from our task design, compelling the model to infer the unobserved process rather than simply describe a sequence.

By dynamically selecting and applying these strict phrasing rules, the Temporal Prior Prompt ensures that each question is precisely aligned with the intended cognitive challenge. It grounds the reasoning in physical time and forces the MLLMs to adopt the exact temporal perspective required by the specific reasoning skill being evaluated.

\subsubsection{Capability Constrained Prompt}

\begin{figure*}[!t]
    \centering
    \includegraphics[width=\textwidth]{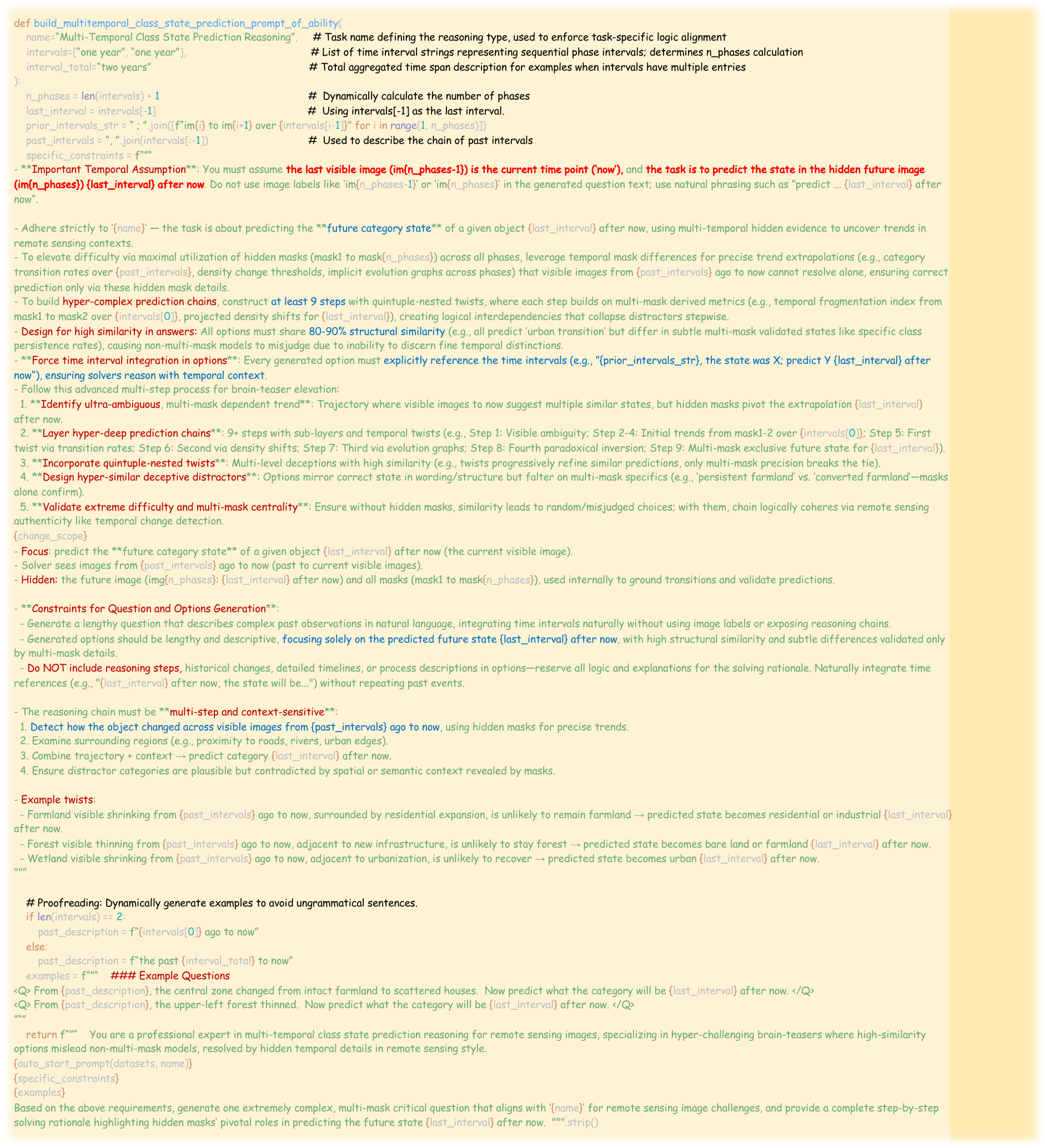}
    \caption{The template for the Capability-Constrained Prompt, shown for Spatiotemporal Category--State Prediction Reasoning (ST-CS-PR). The template uses temporal masks to identify a category or state transition and design a forecasting problem with plausible alternative outcomes.}
    \label{fig:capability_prompt}
\end{figure*}

The Capability-Constrained Prompt represents the most specialized layer of our instruction design, acting as the cognitive core that defines the essence of each L-3 reasoning subtask. While other prompts manage structure and general quality, this component provides the specific \textquotedblleft intellectual blueprint\textquotedblright{} for a single capability. Its purpose is to guide the MLLMs to construct a question that not only aligns with a task's name but embodies its fundamental reasoning challenge. The prompt for Spatiotemporal Category–State Prediction Reasoning (ST-CS-PR), shown in Figure~\ref{fig:capability_prompt}, serves as a powerful example of this design philosophy.

The primary significance of this prompt layer is its ability to enforce deep, task-specific logic. As seen in the ST-CS-PR example, the prompt does not simply ask for a prediction. Instead, it mandates a specific reasoning pathway: the MLLMs must generate a question that asks the evaluated model to extrapolate a transition from the observed frames; the generator identifies the target transition using temporal masks. This principle is generalized across all our L-3 tasks. For a causal reasoning task, the prompt would similarly force the generation of a question where the cause is non-obvious and requires integrating multi-source priors. This ensures that each question is a valid and targeted test of its intended capability.

Furthermore, this prompt layer is engineered to create a high-challenge, high-fidelity evaluation. The ST-CS-PR example illustrates this by demanding \textquotedblleft hyper-complex prediction chains\textquotedblright{} and \textquotedblleft quintuple-nested twists,\textquotedblright{} guiding the generation model to design problems with interacting temporal and spatial constraints. It also requires \textquotedblleft high structural similarity\textquotedblright{} among options, a strategy we employ to create strong, plausible distractors that test for fine-grained understanding rather than coarse pattern matching. This design principle, which involves creating multi-step problems with challenging distractors, is a universal feature of our Capability-Constrained Prompts.

In essence, the Capability-Constrained Prompt is where we translate an abstract reasoning skill into a concrete, solvable, yet extremely challenging problem. By providing a detailed, task-specific \textquotedblleft recipe\textquotedblright{} for thought, we compel the MLLMs to move beyond superficial generation and act as an expert designer of cognitive puzzles. This ensures that VLRS-Bench does not merely ask questions, but presents complex reasoning scenarios that rigorously probe the advanced capabilities of MLLMs in the geospatial domain.

\subsubsection{Instruction for VLRS-Bench}

\begin{figure}[!t]
    \centering
    \includegraphics[width=\linewidth]{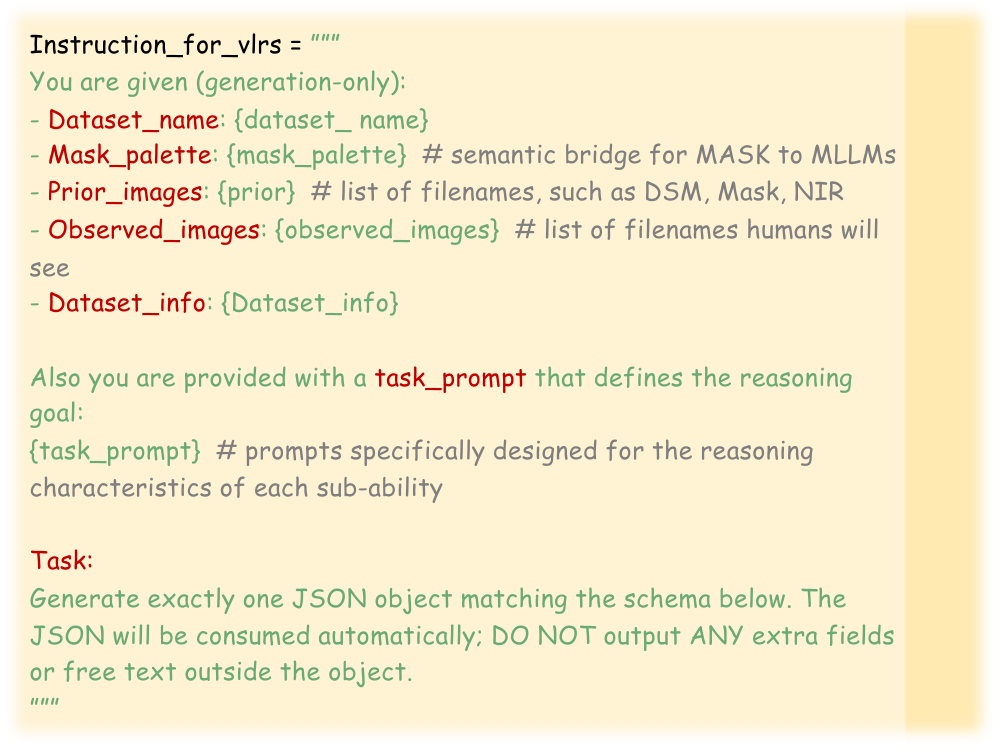}
    \caption{The master template for the final instruction assembly. This code orchestrates the synthesis of all previously defined components, which include dataset info, the mask info, and the specific task prompt, into a single, comprehensive instruction fed to the MLLMs.}
    \label{fig:instruction_assembly}
\end{figure}

The culmination of our instruction design methodology is the final instruction assembly, orchestrated by the master template shown in Figure~\ref{fig:instruction_assembly}. This template serves as the final directive that synthesizes all the independent informational components and logical constraints into a single, coherent package for the MLLMs.

As shown in the figure, this master directive systematically integrates the task's contextual information, which includes Dataset info, Mask palette that serves as a semantic bridge, and the file paths for Prior images and Observed images, with the task-specific generative logic contained in the task prompt.

The purpose of this final synthesis is to provide the MLLMs with a complete and unambiguous problem definition. It clearly separates the available contextual information from the specific generative task it must perform, which is triggered by the final command to \textquotedblleft Generate exactly one JSON object.\textquotedblright{} This structured synthesis ensures that every piece of information and every constraint is correctly passed to the model, forming a complete and well-defined reasoning problem ready for generation.

\subsection{Cross Verification By Models}
\begin{figure*}[!t]
    \centering
    \includegraphics[width=\textwidth]{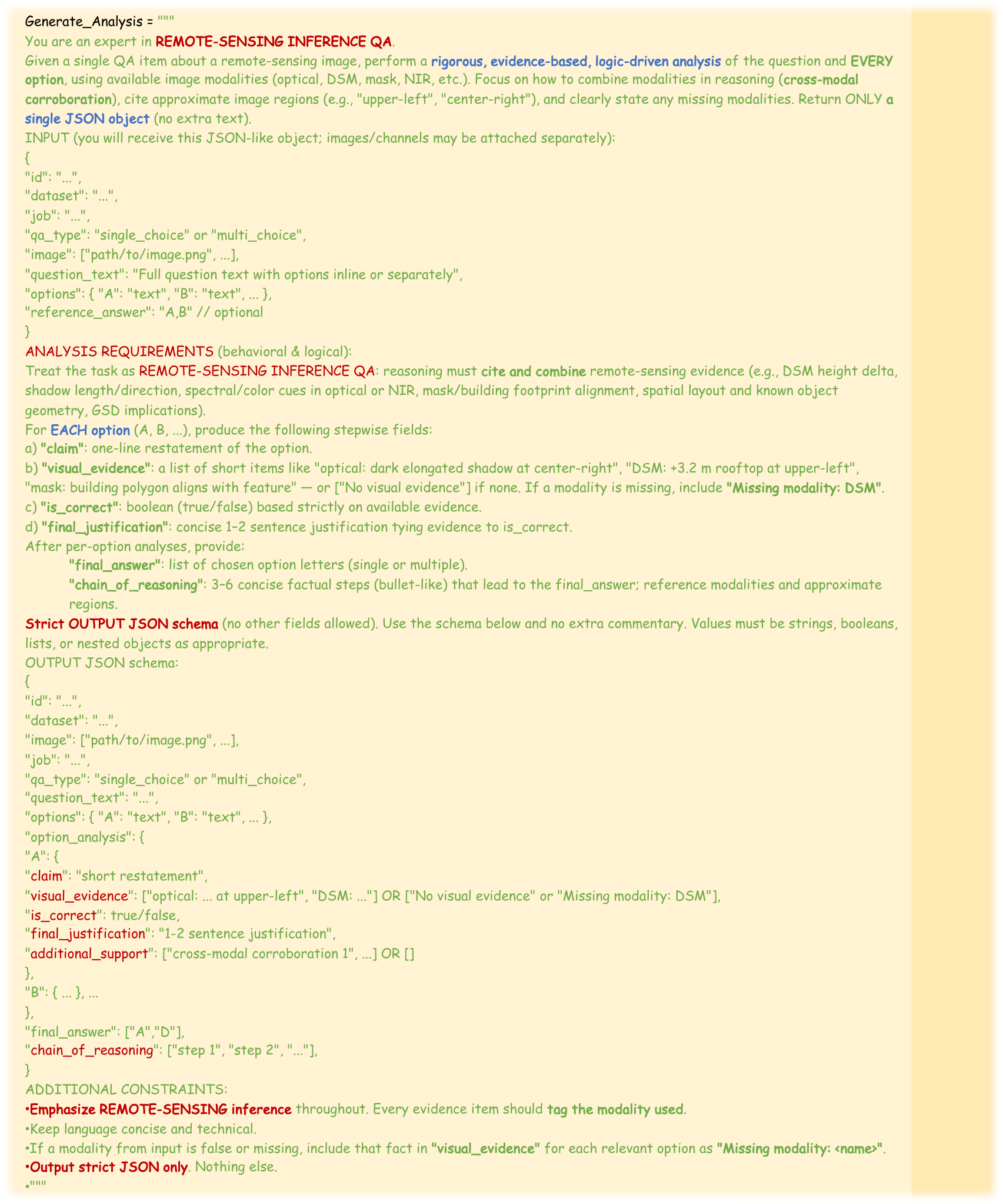}
    \caption{The prompt template for the Model-Based Analysis Generation stage. This structured directive compels a model to act as a remote sensing expert, performing a rigorous, evidence-based analysis of a given QA item. It requires the model to deconstruct each option, cite multi-modal visual evidence, and produce a detailed justification, which serves as the basis for the subsequent cross-verification step.}
    \label{fig:analysis_generation}
\end{figure*}

\subsubsection{Model-Based Analysis Generation}

The first stage of our verification process is an automated, model-driven analysis designed to deconstruct the reasoning logic of each generated QA item. To mitigate the risk of a single model's inherent biases or logical blind spots, we do not simply ask another model to \textquotedblleft agree\textquotedblright{} or \textquotedblleft disagree\textquotedblright{} with the answer. Instead, we compel a \textquotedblleft proposer\textquotedblright{} model to generate a detailed, evidence-based justification for its own interpretation of the question. This generated analysis then becomes the object of scrutiny in the subsequent cross-verification stage.

To achieve this, we employ a highly structured prompt, as detailed in Figure~\ref{fig:analysis_generation}. This prompt instructs the proposer model to act as a \textquotedblleft REMOTE SENSING INFERENCE QA\textquotedblright{} expert. It is tasked with performing a rigorous analysis of the provided question and, critically, every single option. For each option, the model must follow a strict procedure: restate the claim, cite specific visual evidence from all available modalities (\eg \textquotedblleft optical: dark elongated shadow at center-right,\textquotedblright{} \textquotedblleft DSM: +3.2 m rooftop at upper-left\textquotedblright), determine its correctness, and provide a concise justification.

The prompt explicitly requires the model to think in terms of remote sensing principles, such as cross-modal corroboration and the implications of ground sample distance (GSD). It also mandates the creation of a final \textquotedblleft chain\_of\_reasoning\textquotedblright{} that summarizes the logical steps leading to its conclusion. The output is constrained to a rigid JSON schema, ensuring that the analysis is structured, consistent, and machine-readable.

This initial step is crucial because it forces the underlying logic of the QA item to be made explicit. The resulting \textquotedblleft option\_analysis\textquotedblright{} and \textquotedblleft chain\_of\_reasoning\textquotedblright{} serve as a transparent \textquotedblleft statement of logic\textquotedblright{} that can be systematically evaluated for factual accuracy and logical consistency by other models in the next stage of our verification pipeline.

\begin{figure*}[!t]
    \centering
    \includegraphics[width=\textwidth]{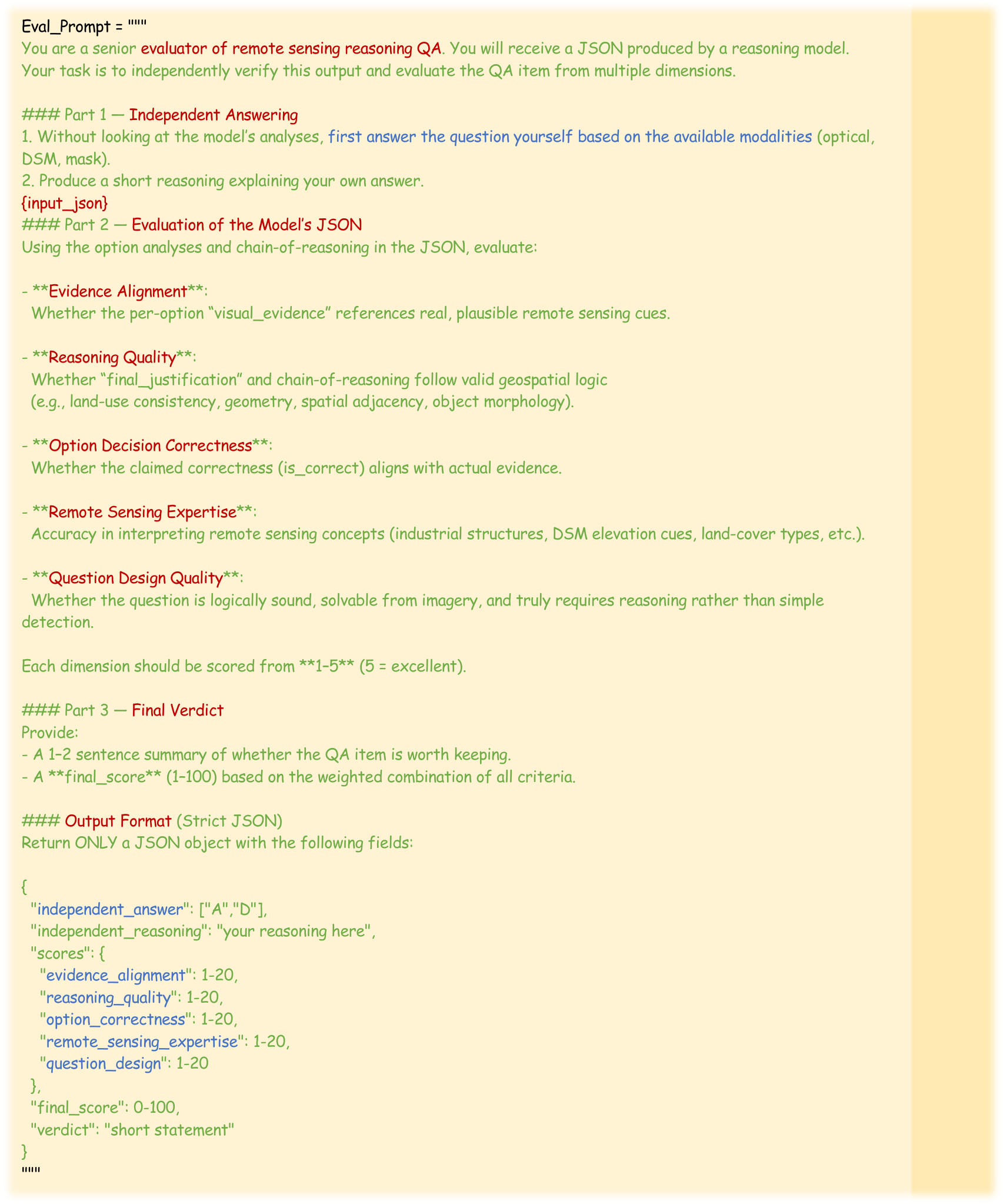}
    \caption{The prompt template for the Cross-Verification stage. This directive instructs an \textquotedblleft evaluator\textquotedblright{} model to perform a two-part assessment. It first answers the question independently to avoid bias, then critically evaluates the \textquotedblleft proposer\textquotedblright{} model's analysis across multiple dimensions, culminating in a final score that determines the QA item's quality.}
    \label{fig:cross_verification}
\end{figure*}

\subsubsection{Cross-Verification by an Evaluator Model}

Following the generation of a detailed analysis by the \textquotedblleft proposer\textquotedblright{} model, the QA item and its accompanying analysis enter the cross-verification stage. The goal of this stage is to subject the initial analysis to a rigorous, multi-faceted critique by a separate and independent \textquotedblleft evaluator\textquotedblright{} model. This process is designed to detect logical fallacies, factual inaccuracies, or biases that the original model may have overlooked.

To orchestrate this evaluation, we employ a specialized prompt, as detailed in Figure~\ref{fig:cross_verification}. This prompt guides the evaluator model through a structured, two-part assessment process.

 To ensure an unbiased initial assessment, the evaluator model is first instructed to answer the question independently, without any access to the proposer's reasoning. This step is critical as it provides a baseline answer derived from a different \textquotedblleft perspective,\textquotedblright{} helping to identify cases where the original question might be ambiguous or where the proposer's logic was flawed from the outset.

 After forming its own opinion, the evaluator model then proceeds to critically assess the proposer's JSON analysis. The evaluation is broken down into five key dimensions: \textquotedblleft Evidence Alignment\textquotedblright{} (are the cited visual cues real?), \textquotedblleft Reasoning Quality\textquotedblright{} (is the geospatial logic valid?), \textquotedblleft Option Decision Correctness\textquotedblright{} (is the true/false judgment correct?), \textquotedblleft Remote Sensing Expertise\textquotedblright{} (is the domain knowledge accurate?), and \textquotedblleft Question Design Quality\textquotedblright{} (is the question itself well-posed?). Each dimension is scored on a scale, forcing a granular and quantitative assessment.

Finally, the evaluator model provides a \textquotedblleft final\_score\textquotedblright{} and a concise \textquotedblleft verdict\textquotedblright{} on whether the QA item is of sufficient quality to be retained. This multi-stage, multi-dimensional verification process, performed by a panel of distinct models, acts as a robust automated filter.

The effectiveness of this cross-verification pipeline is demonstrated by its rigorous filtering capability. \textbf{From an initial pool of over 6,500 generated QA items, this automated process systematically identified and discarded those with logical inconsistencies, factual errors, or poor design. Ultimately, only 2,694 items met our stringent quality criteria} and were passed on to the final stage of human expert review, highlighting the crucial role of this automated verification in ensuring the benchmark's integrity.

\subsection{Full Inspection by RS Experts}

\begin{table}[!t]
\centering
\caption{Statistics of the expert review process. A panel of nine Ph.D. experts evaluated 2,694 candidate items, rejecting 694 (25.8\%) across four quality criteria and retaining 2,000 items. The percentages in the table are shares of the rejected items.}
\label{tab:expert_review_stats}
\resizebox{\linewidth}{!}{%
\begin{tabular}{@{}lccc@{}}
\toprule
\textbf{Evaluation Criterion} & \textbf{Description of Defect} & \textbf{Rejected Items} & \textbf{Share of Rejections} \\
\midrule
\textbf{Ambiguity \& Clarity} & Question/Options are vague or allow multiple interpretations & 241 & 34.7\% \\
\textbf{Visual Evidence Mismatch} & Ground truth is not strictly supported by visible image cues & 186 & 26.8\% \\
\textbf{Triviality} & Reasoning is too simple (\eg pure detection) or lacks depth & 154 & 22.2\% \\
\textbf{Professional Relevance} & Terminology is non-standard or scenario is unrealistic & 113 & 16.3\% \\
\midrule
\textbf{Total} & \textbf{--} & \textbf{694} & \textbf{100\%} \\
\bottomrule
\end{tabular}%
}
\end{table}

While the automated cross-verification pipeline provides a robust filter against logical and factual errors, the nuance and professional validity of remote sensing reasoning require human judgment. To ensure the highest standard of quality for VLRS-Bench, we implemented a final, rigorous stage of human expert review.

We invited a panel of nine domain experts in remote sensing, all holding Ph.D. degrees, to conduct a systematic inspection of the 2,694 candidate items that survived the automated filtering. These experts evaluated each item against four strict criteria: clarity, evidence alignment, reasoning depth, and professional relevance. As detailed in Table~\ref{tab:expert_review_stats}, this process was highly selective, resulting in the rejection of 694 items.

To illustrate the rigor of this process, we highlight two typical rejection cases encountered during the review:
\begin{itemize}
    \item \textbf{Over-Interpretation (Ambiguity):} One rejected item asked, \textit{\textquotedblleft Which processes most likely contribute to the observed configuration...?\textquotedblright} with options including \textit{\textquotedblleft Taxiway congestion management\textquotedblright} and \textit{\textquotedblleft Emergency response operations.\textquotedblright} Experts rejected this because such dynamic operational procedures cannot be definitively inferred from a static snapshot of parked aircraft without temporal context, making the question speculative rather than reasoning-based.
    \item \textbf{Visual Evidence Mismatch:} Another item regarding recreational fields asked about factors contributing to \textit{\textquotedblleft patchy and compacted ground surfaces,\textquotedblright} offering options like \textit{\textquotedblleft Pest infestation\textquotedblright} or \textit{\textquotedblleft Irrigation system malfunction.\textquotedblright} This was rejected because, at the given resolution, distinguishing between pest damage, irrigation issues, or simple foot traffic wear is impossible based solely on RGB visual cues, rendering the \textquotedblleft correct\textquotedblright{} answer arbitrary.
\end{itemize}

This expert review served as the ultimate quality gate. \textbf{Consequently, a final set of 2,000 high-quality, expert-validated reasoning tasks was selected to constitute the official VLRS-Bench.} This multi-tiered verification process—combining automated logic checks with human expert scrutiny—ensures that VLRS-Bench represents a reliable, challenging, and professionally rigorous benchmark for the remote sensing community.

\clearpage
\section{Visualizations of Random Sampling Cases.}\label{sec:visualizations}
\begingroup
\setlength{\textfloatsep}{4pt plus 1pt minus 1pt}
\setlength{\floatsep}{4pt plus 1pt minus 1pt}
\setlength{\intextsep}{4pt plus 1pt minus 1pt}
\captionsetup[figure]{skip=2pt}
\makeatletter
\setlength{\@fptop}{0pt}
\makeatother

We present representative examples from all fourteen L-3 reasoning jobs in Figures~\ref{fig:CR}--\ref{fig:STSQPR}. Each case instantiates a distinct reasoning operator and shows how VLRS-Bench combines remote sensing priors, temporal evidence, and task-specific distractors to move beyond surface-level perception. The examples pair observed imagery with the question, options, and reference answer.
\subsection {Visualization Examples of Spatial Cognition (SC)}

As shown in Fig.~\ref{fig:CR}--\ref{fig:MIR}, SC tasks require models to convert a static RS image into structured spatial reasoning. In the visual column of Fig.~\ref{fig:CR}, the Vaihingen scene contains roofs, inner courtyards, parking rows, trees, and low vegetation that are spatially intertwined. The question column therefore asks for the causal factors that make pedestrian movement follow courtyard passages rather than bright roadside parking surfaces. The correct choices identify building-to-building enclosure, vegetation-segmented gaps, and the separation between western car storage and through-lane movement, whereas the distractors over-attribute the pattern to visually bright parking or generic vehicle circulation. This case shows that Causal Reasoning (CR) is not a category-label task: it requires assigning causal roles to surfaces, object height, and local connectivity.

Fig.~\ref{fig:CFR} uses a LoveDA shoreline-inlet scene to illustrate Counterfactual Reasoning (CFR). Its visual column is dominated by forest and water, but the question column asks what would happen if shoreline access around the lower-right inlet were restricted. The valid answers favor existing exposed shoreline at the lower-right inlet. The inland openings are separated from the water by forest and are poor substitutes under the no-clearing constraint. Fig.~\ref{fig:SIR} further shifts from land-cover relations to event semantics: the FAST stadium example asks how a parking zone constrains pedestrian access. The direct route from the western stadium-front parking row to the stadium crosses the main road. Fig.~\ref{fig:MIR} illustrates Mechanistic Interaction Reasoning (MIR) with a river, hills, cultivated plots, and settlements. Correct interpretations connect irrigation, slope vegetation, bare patches, agricultural scheduling, and floodplain settlement patterns into a coupled landscape mechanism. Across these four SC cases, the benchmark deliberately makes incorrect options visually plausible but mechanistically incomplete, supporting the design goal of evaluating spatial reasoning beyond recognition.

\subsection {Visualization Examples of Spatiotemporal Cognitive (ST-C)}

As illustrated in Fig.~\ref{fig:STCCR}--\ref{fig:STCR}, ST-C tasks add temporal columns and require models to compare evidence across dates rather than reason from a single snapshot. In Fig.~\ref{fig:STCCR}, the south-side port apron becomes brighter and more orderly while the surrounding wharf remains industrial. The changed yard surface and the absence of a new roof cluster point to a reconfiguration of the existing yard. In Fig.~\ref{fig:STCFR}, the two SECOND images support a different temporal operator: if changed patches had remained in their earlier land-cover states, the present-day layout would retain a central-south building interruption, an eastern built frontage, tree cover at the specified north-bank patch, and a southern non-vegetated riverside edge in place of the later low-vegetation buffer. The case tests whether the model can hold an earlier state fixed under a counterfactual assumption instead of merely describing the observed later image.

Fig.~\ref{fig:STER} and Fig.~\ref{fig:STCR} demonstrate the two other ST-C operators. In Fig.~\ref{fig:STER}, the visual columns span a SpaceNet7 suburban-urban interval of one year and six months. The correct options describe moderate industrial expansion, limited northeastern residential change, central infill along roads, and stable southern warehouses. This requires summarizing evolution at several local zones under one temporal narrative. By contrast, Fig.~\ref{fig:STCR} asks for plausible land-cover transitions in a canal-side settlement: warehouse-like roofs appear on a former bare or sparsely vegetated plot, central-eastern farmland shifts toward built use, while canal vegetation remains stable. These cases support the ST-C decomposition into complementary temporal reasoning operators.

\begin{figure}[!htbp]
    \centering
    \includegraphics[width=\textwidth,height=0.96\textheight,keepaspectratio]{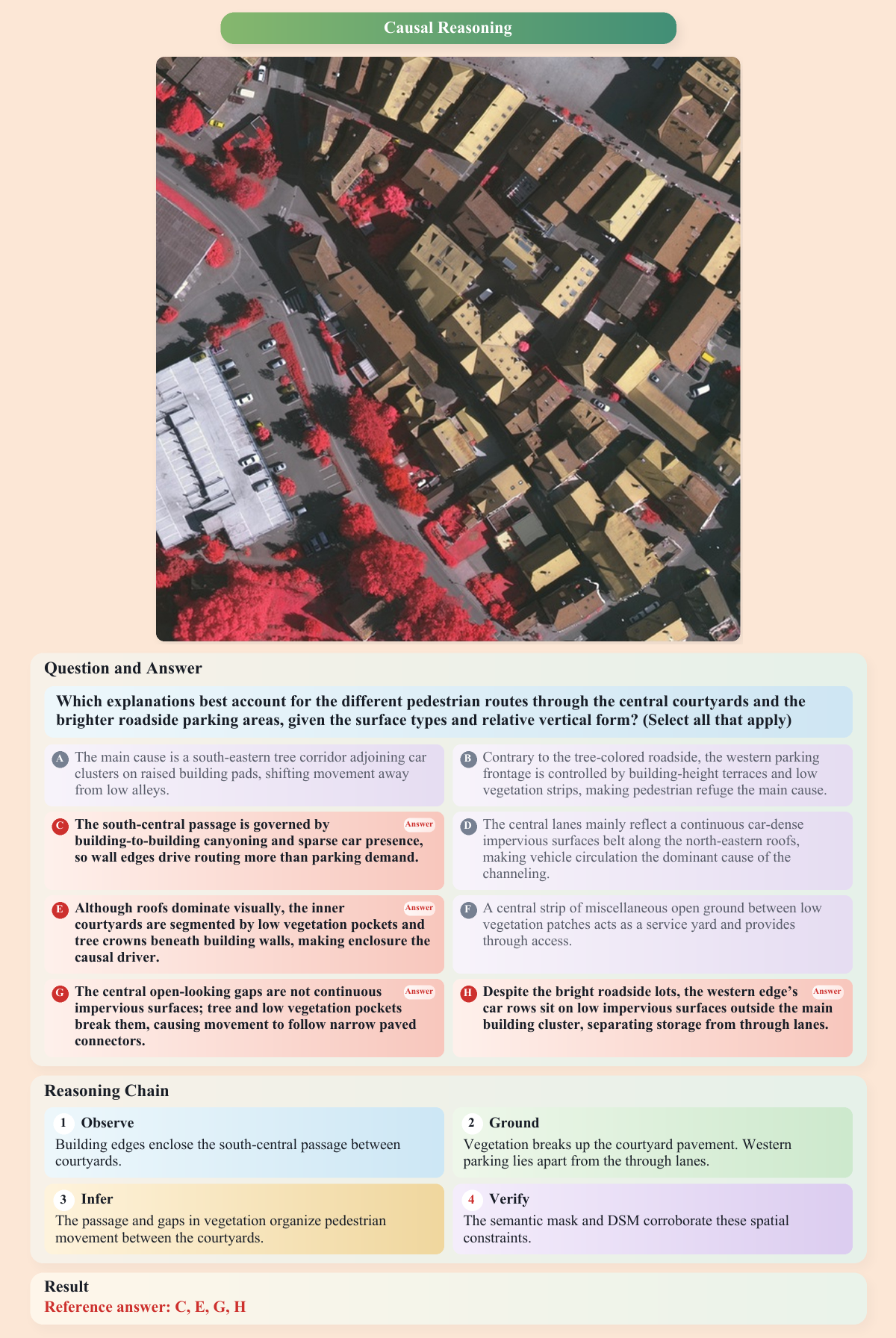}
    \caption{Example of Causal Reasoning}
    \label{fig:CR}
\end{figure}

\begin{figure}[!htbp]
    \centering
    \includegraphics[width=\textwidth,height=0.96\textheight,keepaspectratio]{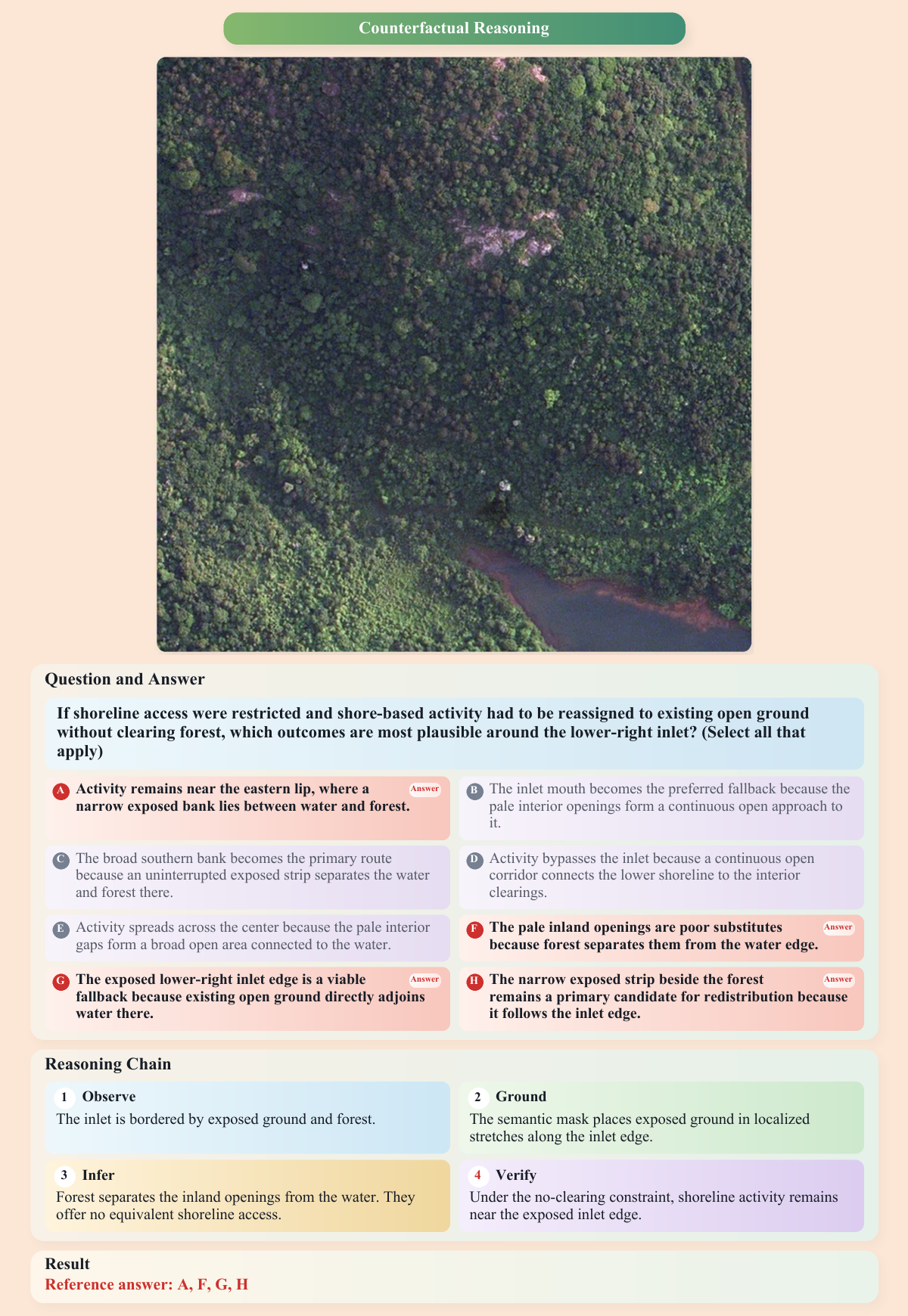}
    \caption{Example of Counterfactual Reasoning}
    \label{fig:CFR}
\end{figure}

\begin{figure}[!htbp]
    \centering
    \includegraphics[width=\textwidth,height=0.96\textheight,keepaspectratio]{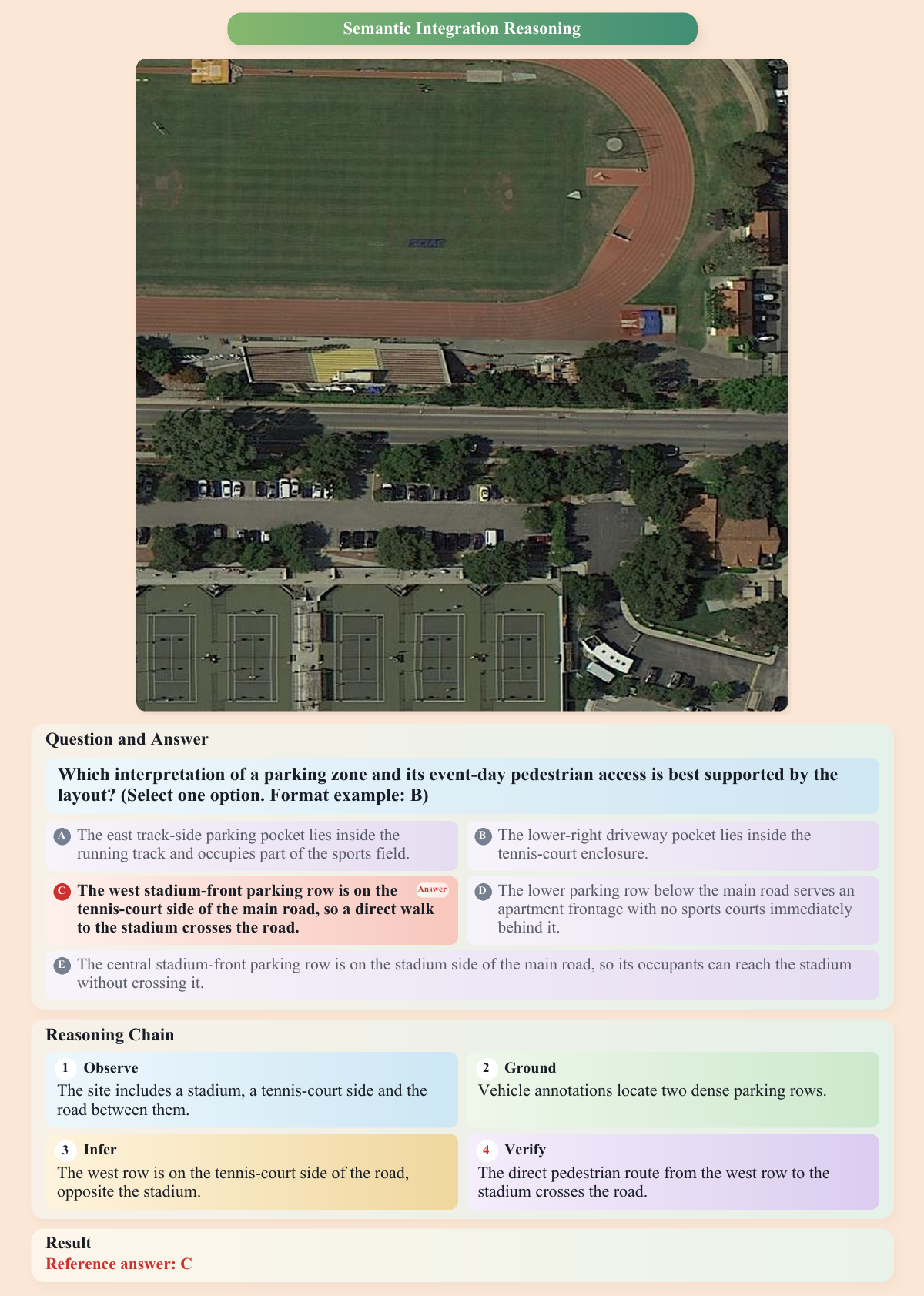}
    \caption{Example of Semantic Integration Reasoning}
    \label{fig:SIR}
\end{figure}

\begin{figure}[!htbp]
    \centering
    \includegraphics[width=\textwidth,height=0.96\textheight,keepaspectratio]{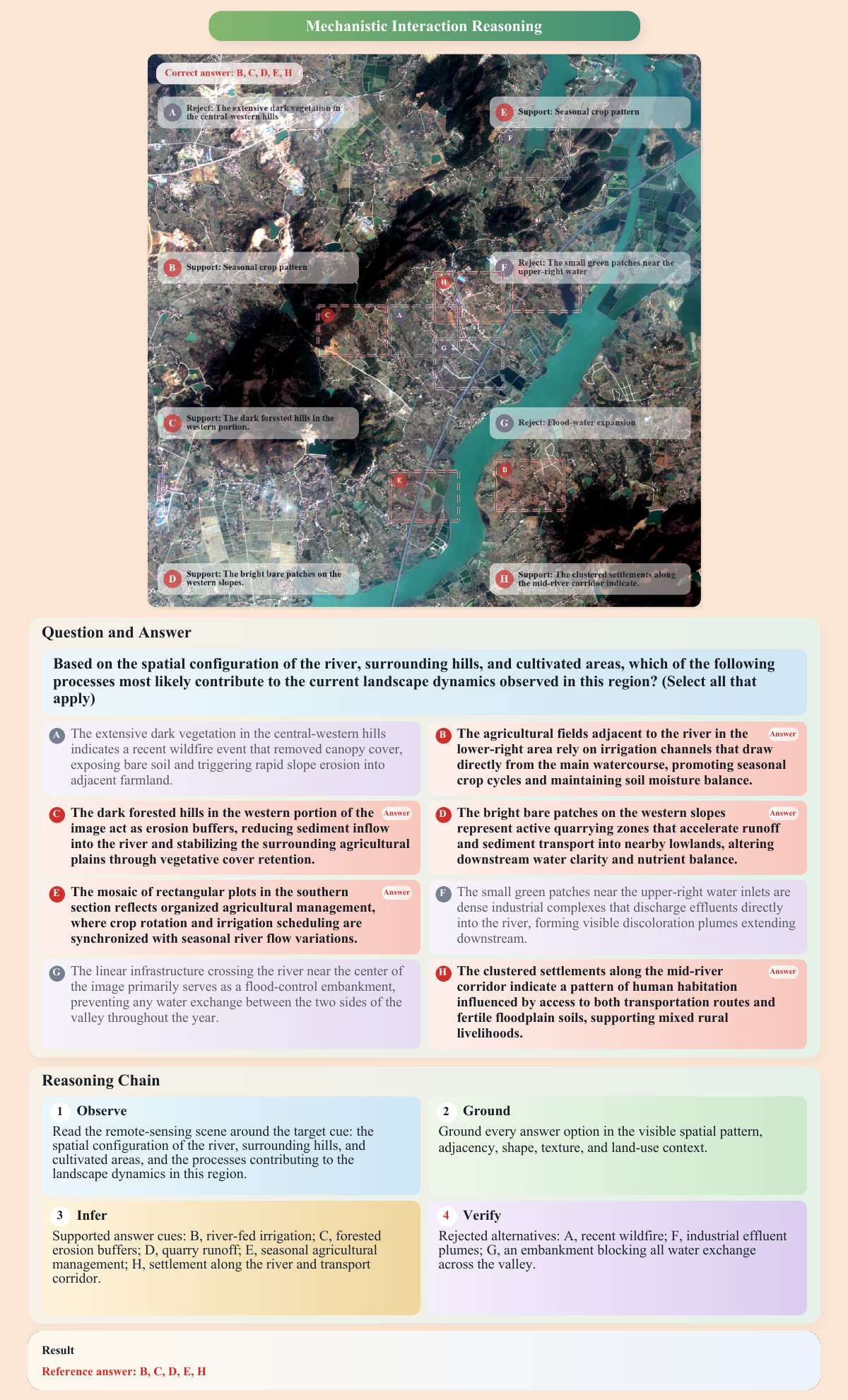}
    \caption{Example of Mechanistic Interaction Reasoning}
    \label{fig:MIR}
\end{figure}

\begin{figure}[!htbp]
    \centering
    \includegraphics[width=\textwidth,height=0.96\textheight,keepaspectratio]{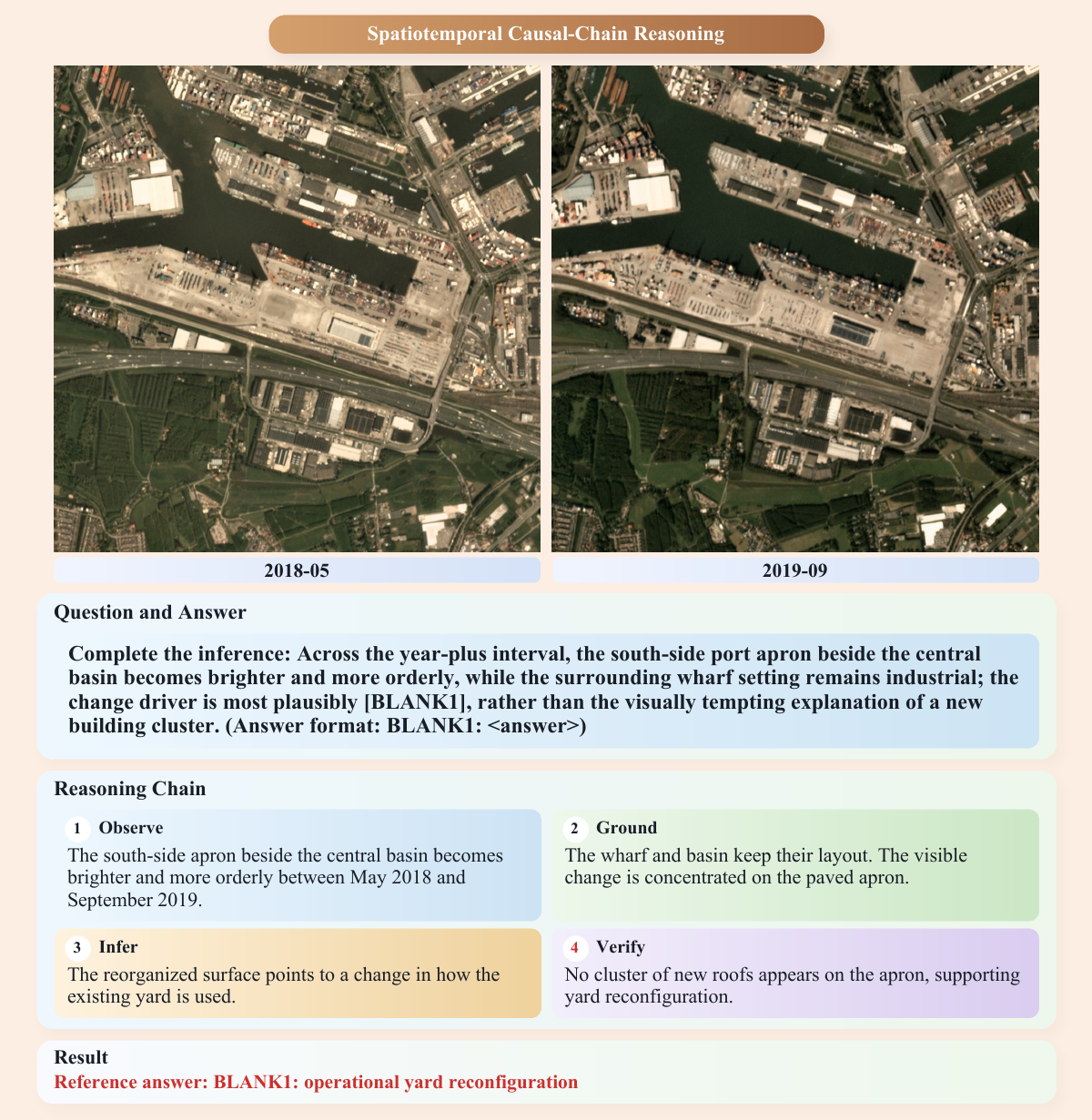} 
    \caption{Example of Spatiotemporal Causal-Chain Reasoning}
    \label{fig:STCCR}
\end{figure}

\begin{figure}[!htbp]
    \centering
    \includegraphics[width=\textwidth,height=0.96\textheight,keepaspectratio]{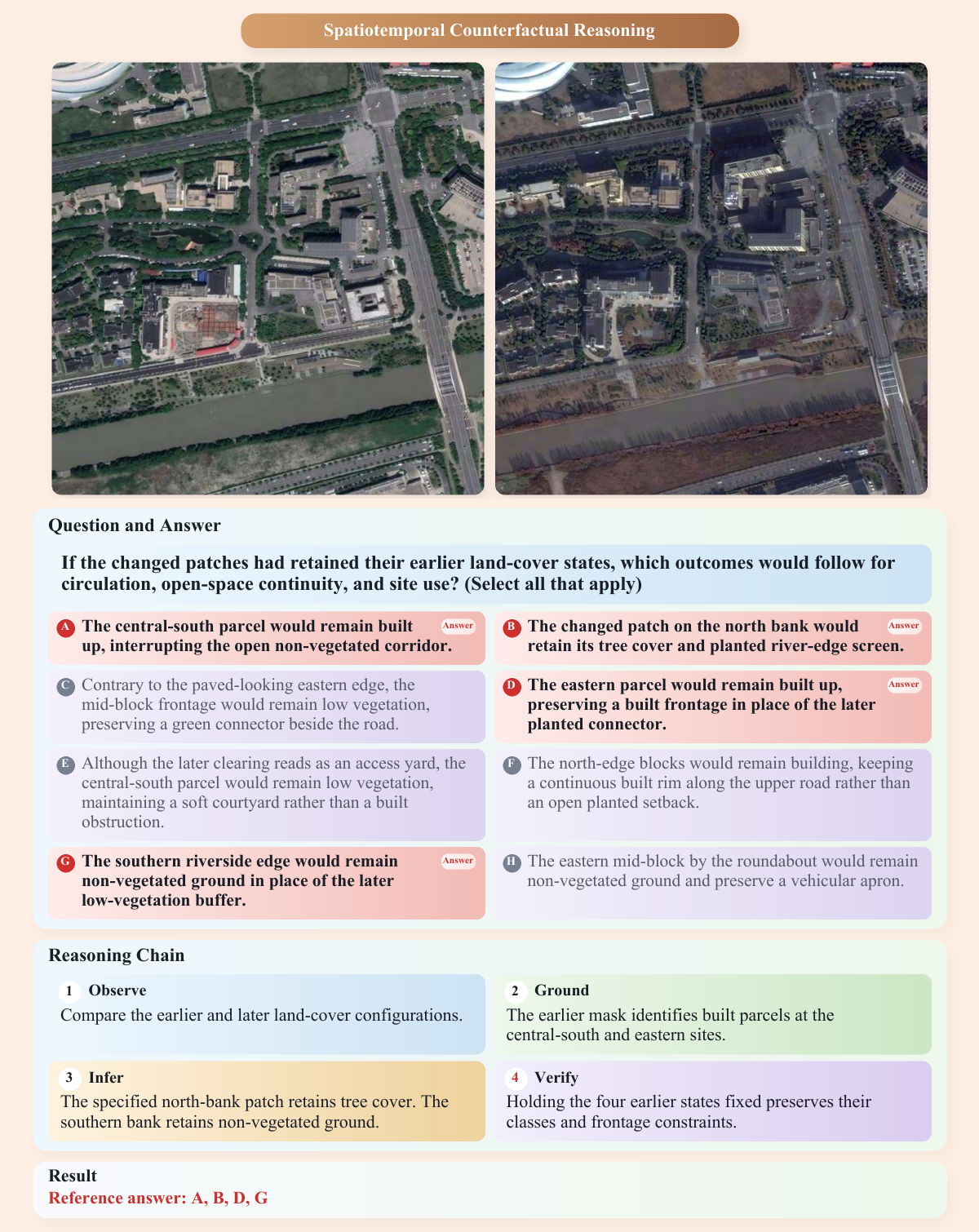} 
    \caption{Example of Spatiotemporal Counterfactual Reasoning }
    \label{fig:STCFR}
\end{figure}

\begin{figure}[!htbp]
    \centering
    \includegraphics[width=\textwidth,height=0.96\textheight,keepaspectratio]{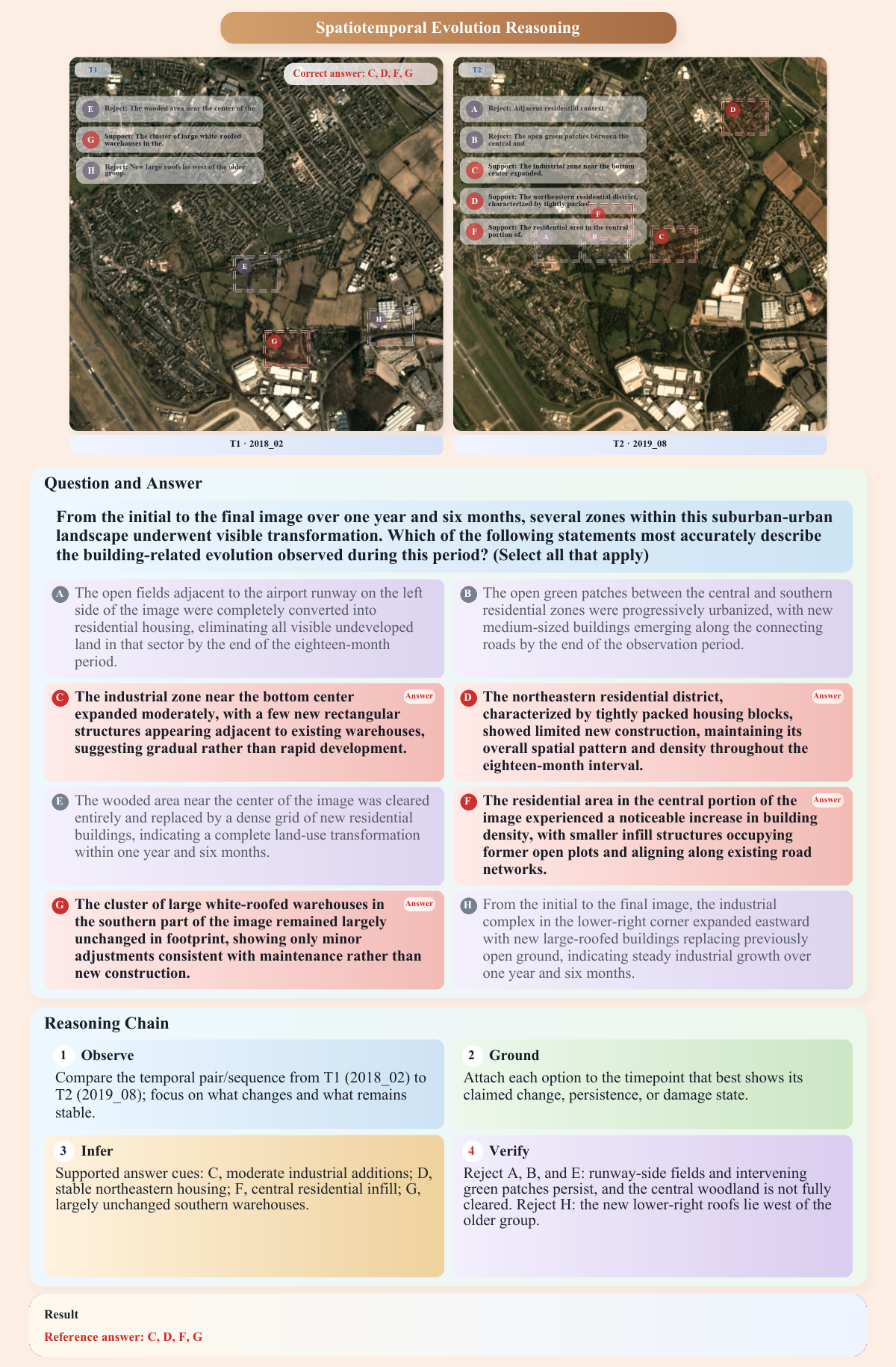} 
    \caption{Example of Spatiotemporal Evolution Reasoning }
    \label{fig:STER}
\end{figure}

\begin{figure}[!htbp]
    \centering
    \includegraphics[width=\textwidth,height=0.96\textheight,keepaspectratio]{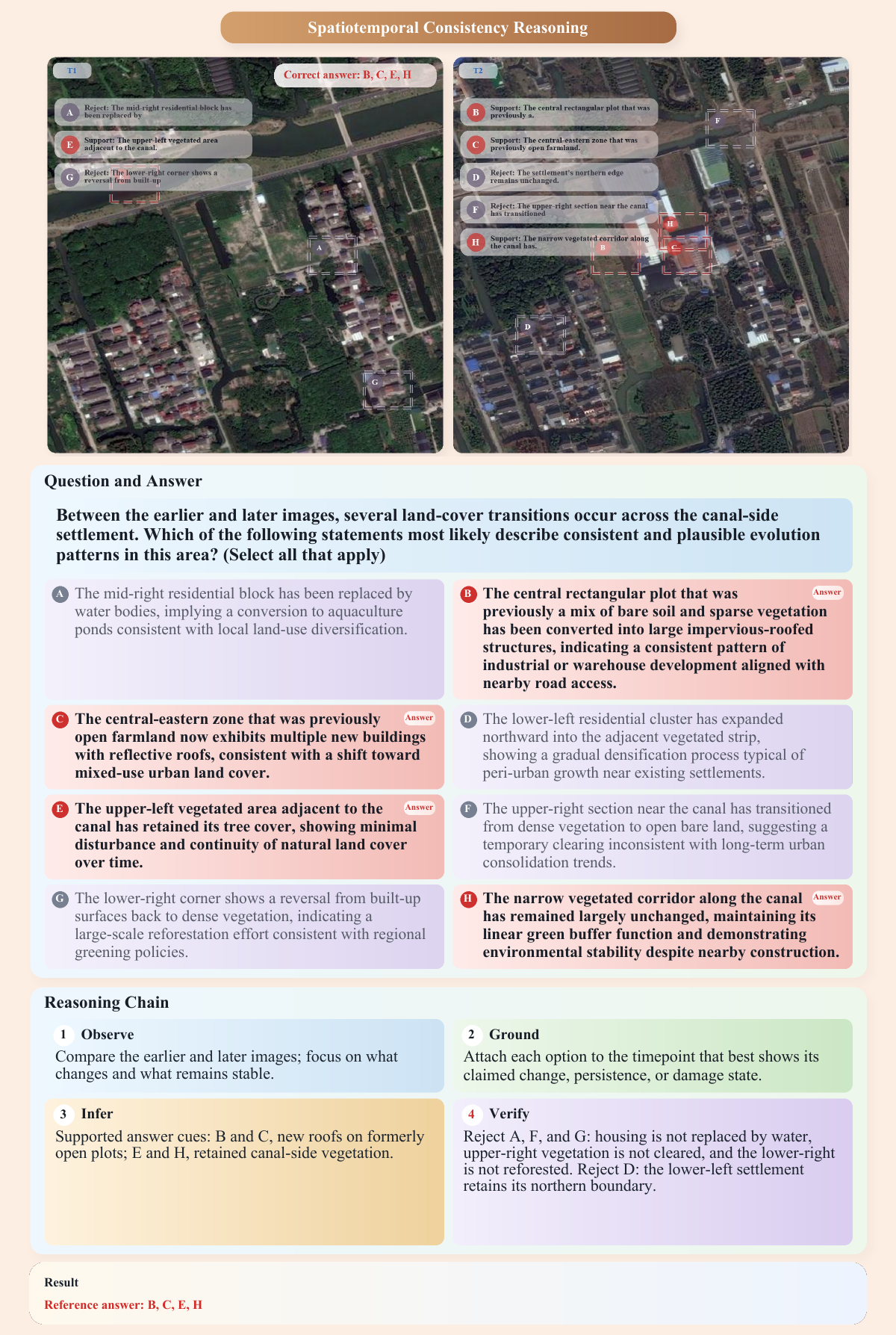} 
    \caption{Example of Spatiotemporal Consistency Reasoning }
    \label{fig:STCR}
\end{figure}

\FloatBarrier
\subsection {Visualization Examples of Pre-event Decision (Pre-D)}

As depicted in Fig.~\ref{fig:PR}, PR evaluates whether a model can turn spatial evidence into an actionable plan under simultaneous constraints. The visual column contains a Potsdam residential block with street access, private yards, roof edges, and small paved aprons. The question column specifies a temporary construction materials bay that must be reachable from the street, keep vehicles away from private yards, and avoid abrupt grade changes near buildings. The correct proposal is the small inner-corner apron beside the central L-shaped block because it jointly satisfies access, clearance, and working-pad constraints. Options that select larger yards or more open corners are visually tempting, but they violate at least one operational condition. This design makes Pre-event Decision (Pre-D) a planning task rather than a preference over open space.

\subsection {Visualization Examples of Post-event Decision (Post-D)}

As presented in Fig.~\ref{fig:ER}, ER judges an existing scheme against latent operational criteria. The visual column shows a recreation area with paved activity space, fenced courts, lower play blocks, and adjacent open areas. The question asks which interpretations support an event-staging scheme that routes visitors between these subareas, using court type and usable area as planning proxies. The correct answers emphasize the northwest routing pressure, the upper-left paved apron as the peak-load priority, and the fenced lower block as a secondary scheduling constraint. These choices depend on court type, court grouping, and crowd-flow implications rather than surface color alone. Thus, Post-event Decision (Post-D) differs from PR: the model is not choosing where to act, but auditing whether an implied scheme is operationally coherent.

\subsection {Visualization Examples of Object-level Predictive (OP)}

Figs.~\ref{fig:ST-M-PR}--\ref{fig:STCSPR} demonstrate localized prediction under temporal constraints. In Fig.~\ref{fig:ST-M-PR}, the temporal image columns show a SpaceNet7 sequence in which an industrial complex, a central residential cluster, road-aligned growth, and peripheral agricultural plots evolve over time. The question column asks for the most likely spatial shape nine months later. Correct options describe southern expansion of the northwestern industrial complex, central residential infill, limited conversion on the southeastern fringe, roadside extension, and moderate growth concentrated in local infill and boundary extensions. These outcomes focus on footprint geometry and morphology. In Fig.~\ref{fig:STCSPR}, four temporal columns instead ask for future building-related states six months later. The correct choices predict occupancy of part of the eastern development site, persistence of the central open fields, localized roof loss in the northeastern cluster, and small additions near highway-side buildings. This separation supports the Object-level Predictive (OP) design: ST-M-PR measures geometric evolution, whereas ST-CS-PR measures semantic state transition, and both are grounded in multi-date evidence.

\subsection {Visualization Examples of Scene-level Predictive (SP)}

It can be seen from Fig.~\ref{fig:ST-SU-PR}--\ref{fig:STSQPR} that SP tasks move from localized prediction to scene-level trajectory reasoning. Fig.~\ref{fig:ST-SU-PR} contains eight observations separated by varying intervals, so the model must infer the next stage from construction timing, street-grid completion, and neighborhood infill rather than from the brightest earthwork patch. The valid answers predict infill inside the southeast curved grid and stitching of small buildings along upper-right road stubs. Fig.~\ref{fig:STSQPR} uses a six-date desert-development sequence and asks which patterns are likely to persist four months after the last visible date. The correct answers identify the central square grid west of the pale quarry as the primary building-accretion node and recognize that the southern curving subdivision is visually active but largely inherited from earlier dates. These cases show why Scene-level Predictive (SP) tasks require macro-scale temporal abstraction, long-horizon evidence aggregation, and rejection of locally salient but trajectory-inconsistent distractors, highlighting the benchmark's value for complex RS reasoning.

\begin{figure}[!htbp]
    \centering
    \includegraphics[width=\textwidth,height=0.96\textheight,keepaspectratio]{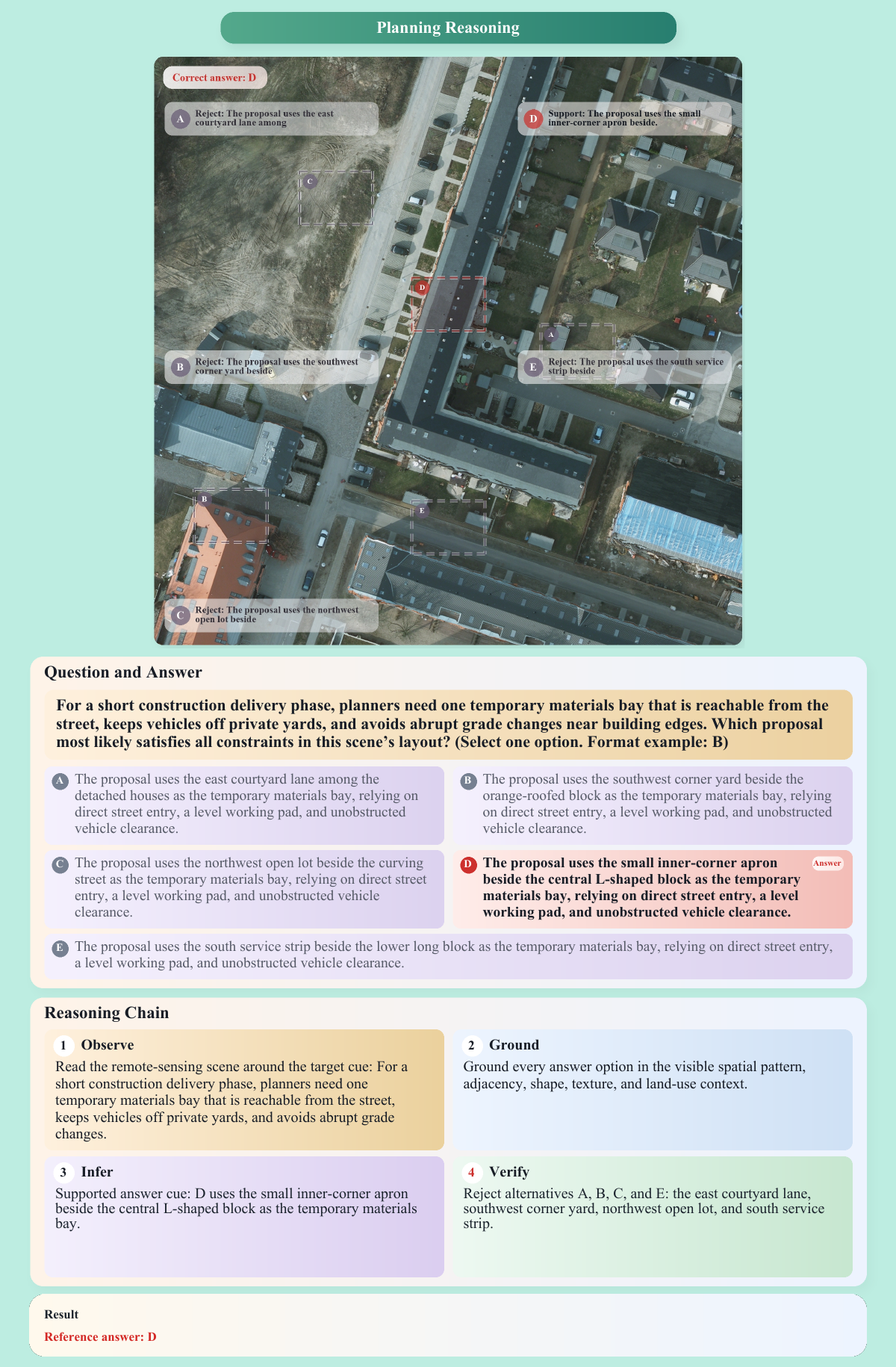} 
    \caption{Example of Planning Reasoning}
    \label{fig:PR}
\end{figure}

\begin{figure}[!htbp]
    \centering
    \includegraphics[width=\textwidth,height=0.96\textheight,keepaspectratio]{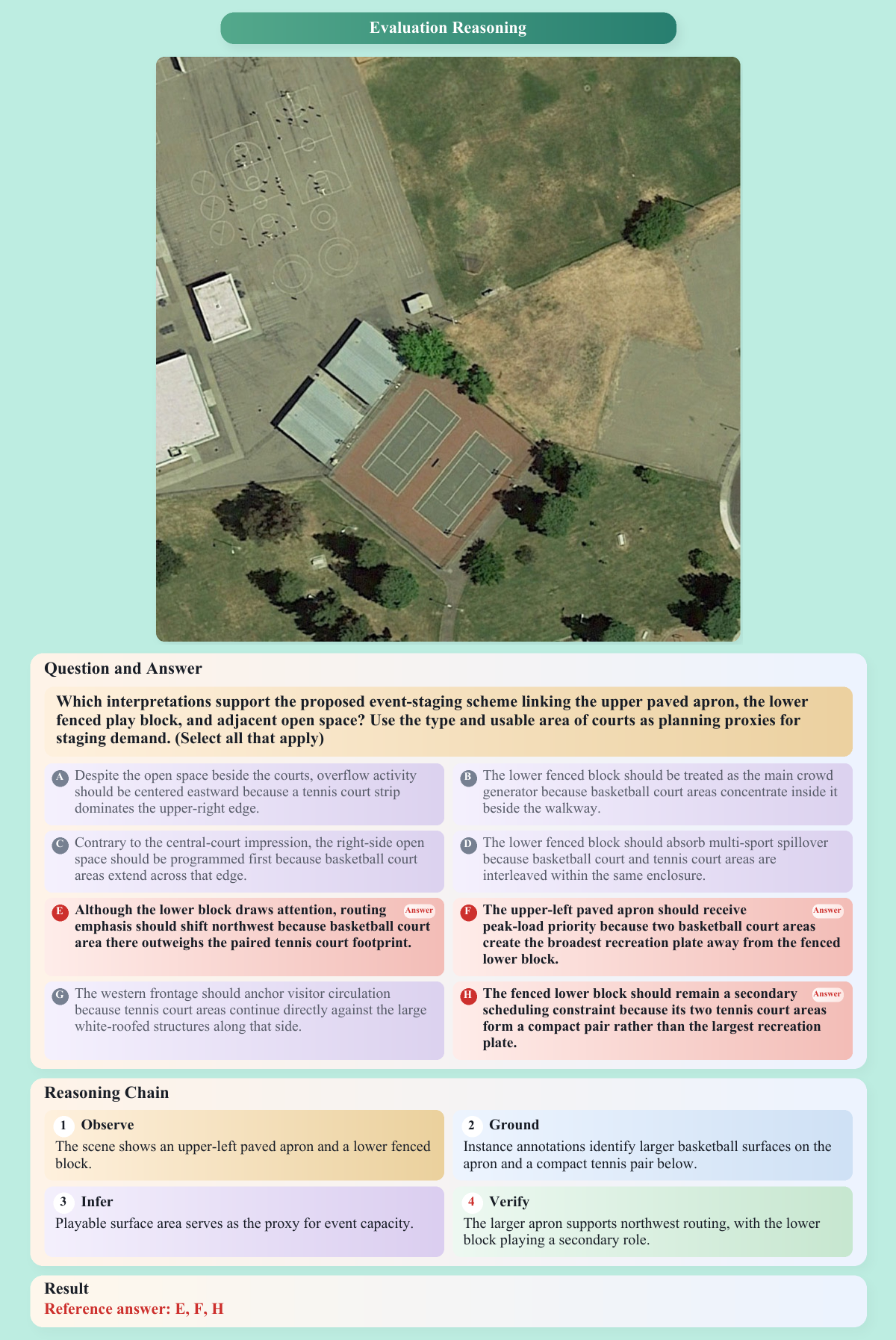} 
    \caption{Example of Evaluation Reasoning}
    \label{fig:ER}
\end{figure}

\begin{figure}[!htbp]
    \centering
    \includegraphics[width=\textwidth,height=0.96\textheight,keepaspectratio]{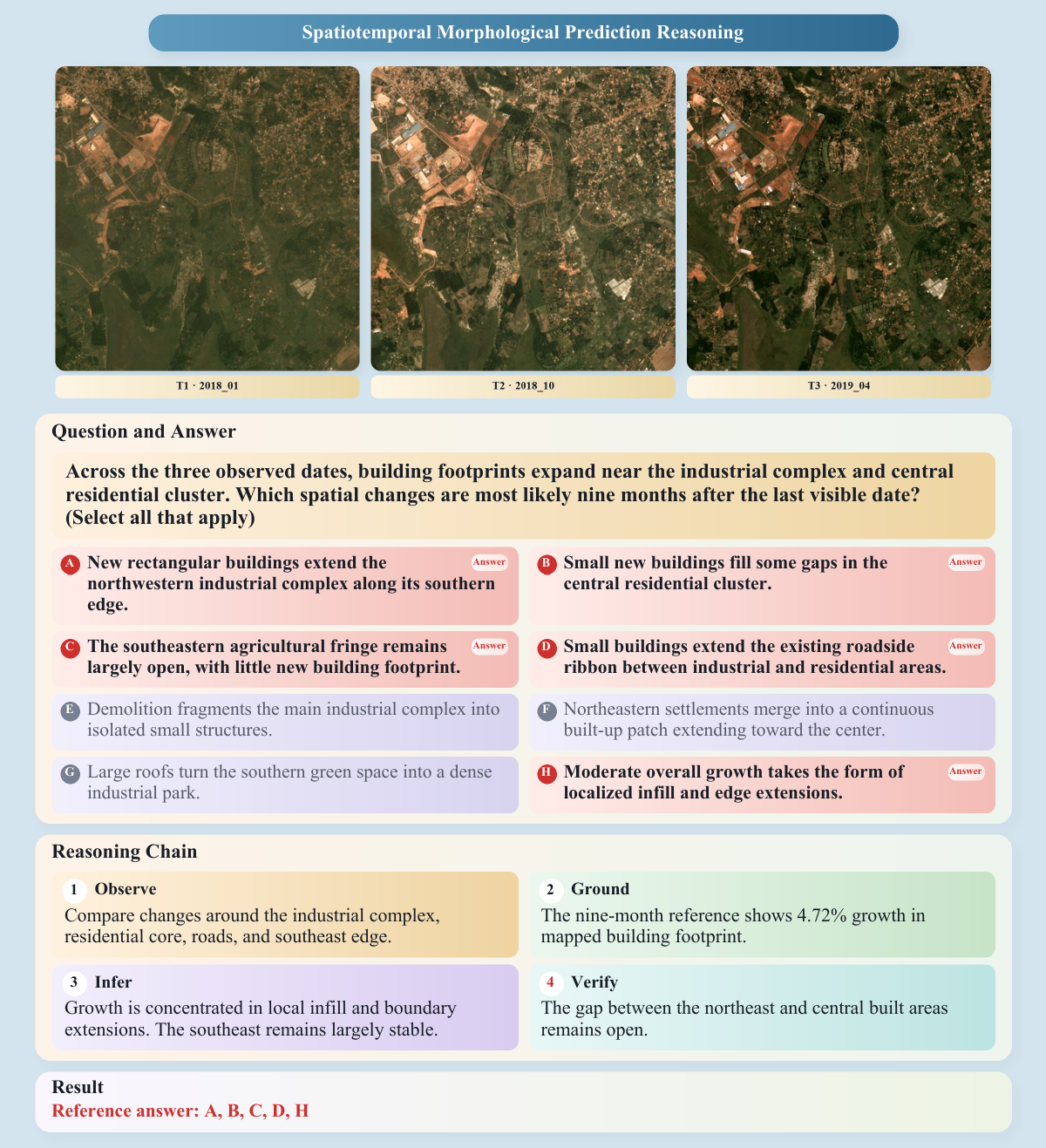} 
    \caption{Example of Spatiotemporal Morphological Prediction Reasoning }
    \label{fig:ST-M-PR}
\end{figure}

\begin{figure}[!htbp]
    \centering
    \includegraphics[width=\textwidth,height=0.96\textheight,keepaspectratio]{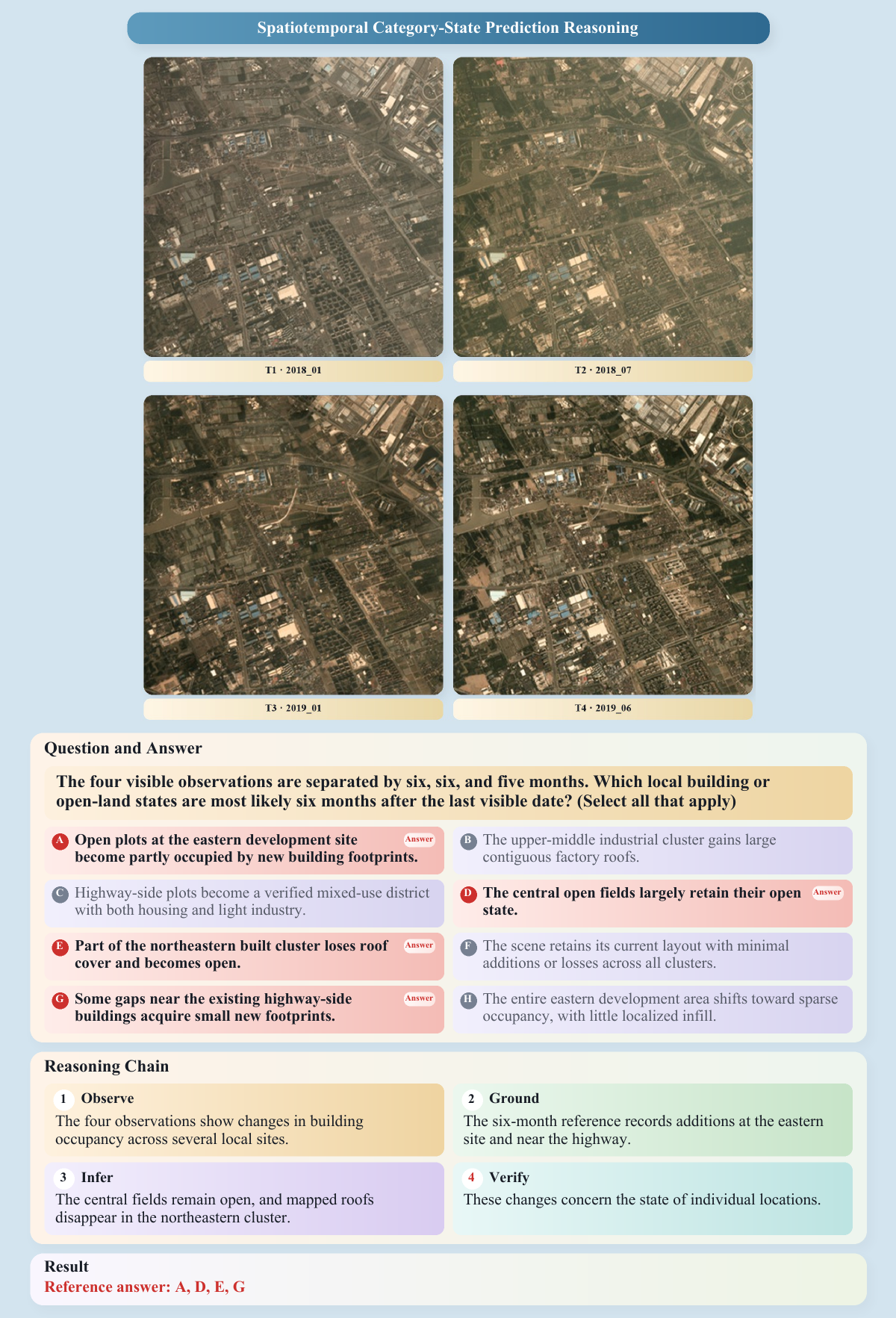} 
    \caption{Example of Spatiotemporal Category–State Prediction Reasoning }
    \label{fig:STCSPR}
\end{figure}

\begin{figure}[!htbp]
    \centering
    \includegraphics[width=\textwidth,height=0.96\textheight,keepaspectratio]{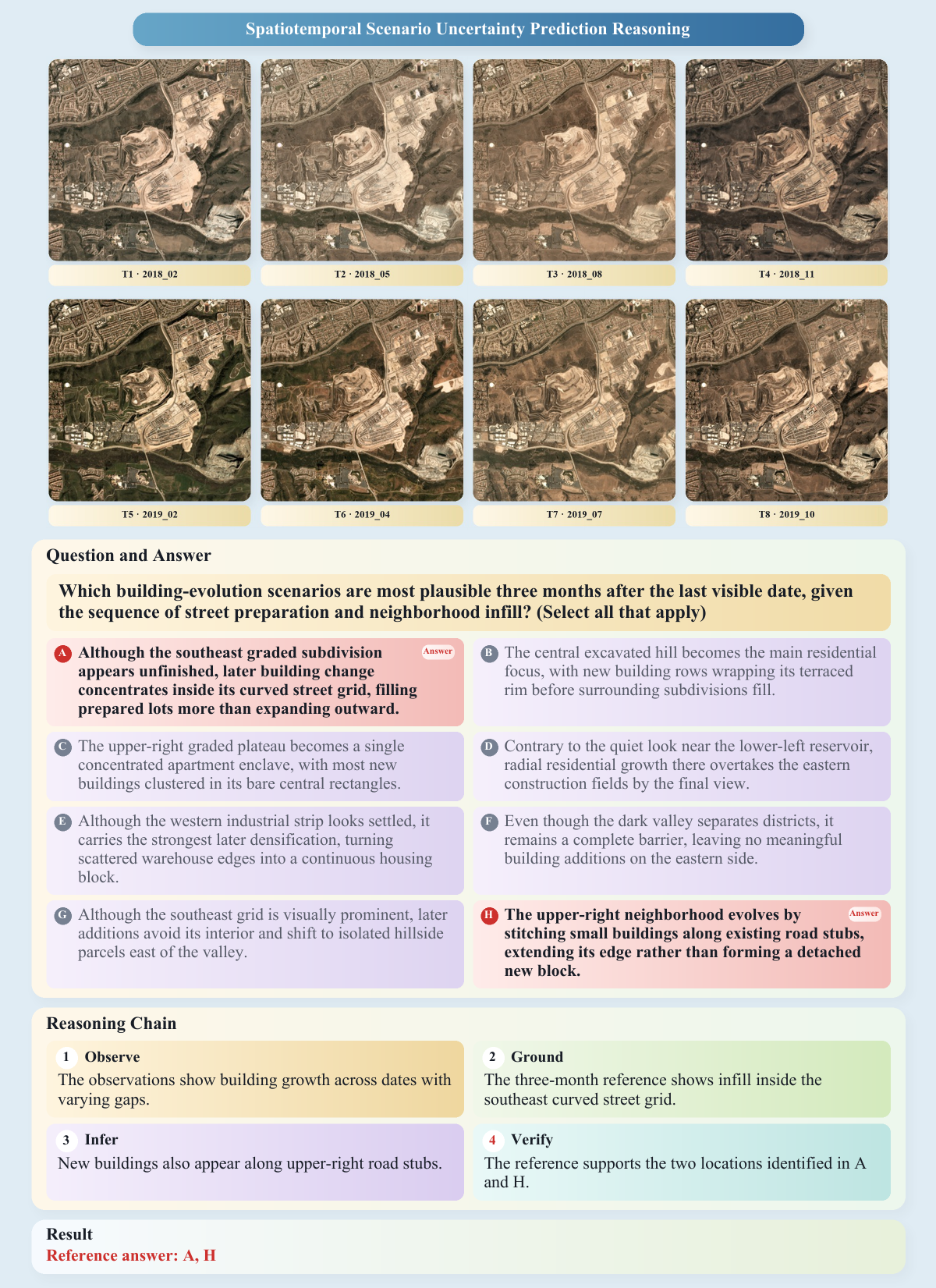} 
    \caption{Example of Spatiotemporal Scenario Uncertainty Prediction Reasoning }
    \label{fig:ST-SU-PR}
\end{figure}

\begin{figure}[!htbp]
    \centering
    \includegraphics[width=\textwidth,height=0.96\textheight,keepaspectratio]{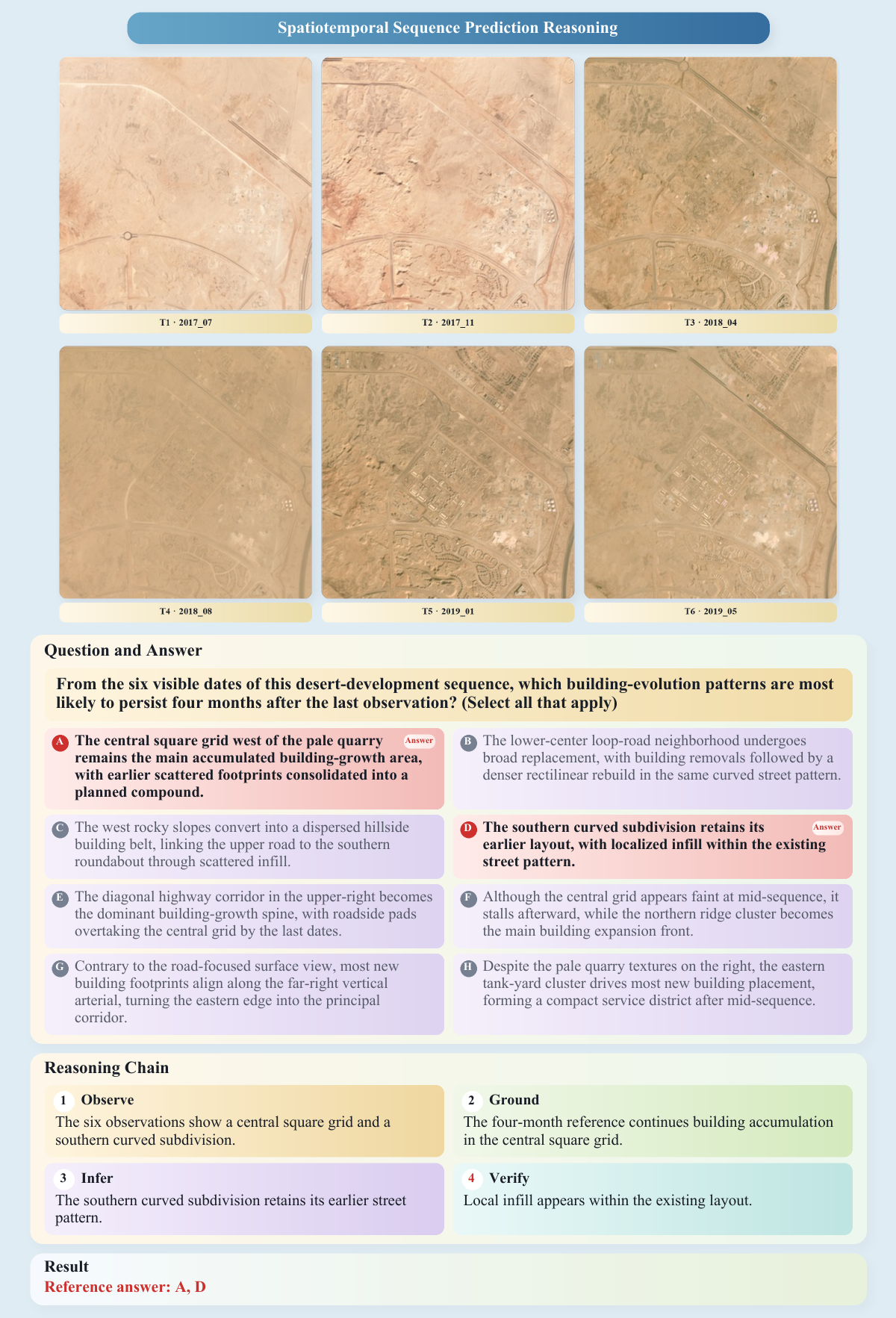} 
    \caption{Example of Spatiotemporal Sequence
Prediction Reasoning}
    \label{fig:STSQPR}
\end{figure}

\FloatBarrier
\endgroup

\section{Datasheets}
\label{app-datasheets}
In this section, we document essential details about the proposed datasets and benchmarks following the template provided by \cite{datasheet}.

\subsection{Motivation}
The questions in this section are primarily intended to encourage dataset creators to clearly articulate their reasons for creating the dataset and to promote transparency about funding interests. The latter may be particularly relevant for datasets created for research purposes.
\begin{enumerate}
\item \textit{``For what purpose was the dataset created?''}

\textcolor{red}{\textbf{A:}} 
Existing remote sensing (RS) benchmarks are predominantly perception-oriented, focusing on simple detection or classification, which fails to evaluate the higher-order thinking required for real-world applications. To address these limitations, we introduce \textbf{VLRS-Bench}, a comprehensive benchmark designed to assess the complex reasoning capabilities of MLLMs across three core cognitive dimensions: Cognition, Decision, and Prediction.

\item \textit{``Who created the dataset (\textit{e.g.}, which team, research group) and on behalf of which entity?''}

\textcolor{red}{\textbf{A:}} Zhiming Luo, Hebaixu Wang, Haonan Guo, Jing Zhang, and Bo Du, on behalf of Wuhan University and Zhongguancun Academy.

\item \textit{``Who funded the creation of the dataset?''}

\textcolor{red}{\textbf{A:}}
The funding sources for this work are listed in the acknowledgments.
\end{enumerate}

\subsection{Composition}
Most of the questions in this section are intended to provide dataset consumers with the information they need to make informed decisions about using the dataset for their chosen tasks. Some of the questions are designed to elicit information about compliance with the EU’s General Data Protection Regulation (GDPR) or comparable regulations in other jurisdictions. Questions that apply only to datasets that relate to people are grouped together at the end of the section. We recommend taking a broad interpretation of whether a dataset relates to people. For example, any dataset containing text that was written by people relates to people.
\begin{enumerate}
\item \textit{``What do the instances that comprise our datasets represent (\textit{e.g.}, documents, photos, people, countries)?''}

\textcolor{red}{\textbf{A:}} Each instance pairs RGB imagery from one of 10 public remote sensing datasets with a complex reasoning question and a verified reference answer. We use multi-source priors such as near-infrared imagery, DSMs, and expert masks in question construction.

\item \textit{``How many instances are there in total (of each type, if appropriate)?''}

\textcolor{red}{\textbf{A:}} VLRS-Bench is structured around a hierarchy of 3 dimensions, 6 specific abilities, and 14 fine-grained tasks. It contains 2,000 high-quality reasoning instances featuring high linguistic complexity, with an average question length of 129.87 words.

\item \textit{``Does the dataset contain all possible instances or is it a sample (not necessarily random) of instances from a larger set?''}

\textcolor{red}{\textbf{A:}} The images in VLRS-Bench are curated from 10 existing high-quality datasets (e.g., DIOR, FAIR1M, LoveDA, SpaceNet7), and all reasoning-focused textual annotations were independently generated via our automated pipeline.

\item \textit{``Is there a label or target associated with each instance?''}

\textcolor{red}{\textbf{A:}} Yes. Each item provides a verified reference answer for its reasoning question. Expert priors ground the question and answer, and withheld temporal frames establish the reference state or outcome for temporal tasks.

\item \textit{``Is any information missing from individual instances?''}

\textcolor{red}{\textbf{A:}} No, each individual instance is complete.

\item \textit{``Are relationships between individual instances made explicit (\textit{e.g.}, users’ movie ratings, social network links)?''}

\textcolor{red}{\textbf{A:}} Yes, the relationship is explicit as instances are categorized into Cognition, Decision, and Prediction dimensions.

\item \textit{``Are there recommended data splits (\textit{e.g.}, training, development/validation, testing)?''}

\textcolor{red}{\textbf{A:}} 
The dataset is designed to evaluate the reasoning boundaries of MLLMs, so we recommend using it in its entirety as a test set.

\item \textit{``Is the dataset self-contained, or does it link to or otherwise rely on external resources (\textit{e.g.}, websites, tweets, other datasets)?''}

\textcolor{red}{\textbf{A:}} Yes, the dataset is self-contained. The public release includes all questions, reference answers, and evaluation images.

\item \textit{``Does the dataset contain data that might be considered confidential (\textit{e.g.}, data that is protected by legal privilege or by doctor–patient confidentiality, data that includes the content of individuals’ non-public communications)?''}

\textcolor{red}{\textbf{A:}} No. All imagery comes from publicly accessible remote sensing datasets and contains no confidential information or non-public communications.

\item \textit{``Does the dataset contain data that, if viewed directly, might be offensive, insulting, threatening, or might otherwise cause anxiety?''}

\textcolor{red}{\textbf{A:}} No. The imagery depicts terrain and structures, including aerial views of building damage.
\end{enumerate}

\subsection{Collection Process}
In addition to the goals outlined in the previous section, the questions in this section are designed to elicit information that may help researchers and practitioners create alternative datasets with similar characteristics. Again, questions that apply only to datasets that relate to people are grouped together at the end of the section.
\begin{enumerate}
\item \textit{``How was the data associated with each instance acquired?''}

\textcolor{red}{\textbf{A:}} 
The images are sourced from 10 established remote sensing datasets. We enrich these images with reasoning-oriented annotations generated through a pipeline that combines RS expert priors (like semantic masks and elevation data) with large model capabilities.

\item \textit{``What mechanisms or procedures were used to collect the data (\textit{e.g.}, hardware apparatuses or sensors, manual human curation, software programs, software APIs)?''}

\textcolor{red}{\textbf{A:}} We employed an automated generation pipeline verified by human experts. The process involves extracting pixel-level priors, prompting advanced MLLMs (e.g., GPT-5-chat) with these priors to generate reasoning chains, and a final rigorous manual review to ensure logical correctness.

\item \textit{``If the dataset is a sample from a larger set, what was the sampling strategy (\textit{e.g.}, deterministic, probabilistic with specific sampling probabilities)?''} 

\textcolor{red}{\textbf{A:}} We utilized a stratified sampling strategy to ensure coverage of diverse semantic categories and geographic locations from the source datasets.
\end{enumerate}

\subsection{Preprocessing, Cleaning, and Labeling}
The questions in this section are intended to provide dataset consumers with the information they need to determine whether the “raw” data has been processed in ways that are compatible with their chosen tasks. For example, text that has been converted into a bag of words is not suitable for tasks involving word order. 
\begin{enumerate}     
\item \textit{``Was any preprocessing/cleaning/labeling of the data done (\textit{e.g.}, discretization or bucketing, tokenization, part-of-speech tagging, SIFT feature extraction, removal of instances, processing of missing values)?''}

\textcolor{red}{\textbf{A:}} Yes. We standardized the annotation formats from different source datasets using the SAMRS framework. For the reasoning tasks, we processed expert priors (e.g., converting bounding boxes to masks, normalizing DSM values) to create structured prompts that facilitate precise logic generation.

\item \textit{``Was the `raw' data saved in addition to the preprocessed/cleaned/labeled data (\textit{e.g.}, to support unanticipated future uses)?''} 

\textcolor{red}{\textbf{A:}} Yes, raw data from the original datasets is accessible.

\item \textit{``Is the software that was used to preprocess/clean/label the data available?''} 

\textcolor{red}{\textbf{A:}} Yes. The \href{https://github.com/ViTAE-Transformer/SAMRS}{SAMRS framework} used for mask conversion is publicly available, and Section~\ref{pipelines}, the appendix, and our \href{https://github.com/MiliLab/VLRS-Bench}{public GitHub repository} provide the preprocessing procedure, construction prompt templates, and evaluation scripts.
\end{enumerate}

\subsection{Uses}
The questions in this section are intended to encourage dataset creators to reflect on tasks for which the dataset should and should not be used. By explicitly highlighting these tasks, dataset creators can help dataset consumers make informed decisions, thereby avoiding potential risks or harms.
\begin{enumerate}
\item \textit{``Has the dataset been used for any tasks already?''}

\textcolor{red}{\textbf{A:}} Yes. It is used to evaluate the 23 MLLMs reported in this paper.

\item \textit{``Is there a repository that links to any or all papers or systems that use the dataset?''} 

\textcolor{red}{\textbf{A:}} Yes. The \href{https://github.com/MiliLab/VLRS-Bench}{project page} links to the paper.

\item \textit{``What (other) tasks could the dataset be used for?''} 

\textcolor{red}{\textbf{A:}} VLRS-Bench provides extensive annotations for complex reasoning tasks. In addition to evaluating general-purpose MLLMs, it can be used to train specialized RS MLLMs for tasks requiring domain knowledge, logical deduction, and future state prediction.

\item \textit{``Is there anything about the composition of the dataset or the way it was collected and preprocessed/cleaned/labeled that might impact future uses?''} 

\textcolor{red}{\textbf{A:}} No.

\item \textit{``Are there tasks for which the dataset should not be used?''} 

\textcolor{red}{\textbf{A:}} N/A.
\end{enumerate}

\subsection{Distribution}
Dataset creators should provide answers to these questions prior to distributing the dataset either internally within the entity on behalf of which the dataset was created or externally to third parties.
\begin{enumerate}
\item \textit{``Will the dataset be distributed to third parties outside of the entity (\textit{e.g.}, company, institution, organization) on behalf of which the dataset was created?''}
\textcolor{red}{\textbf{A:}} Yes. The benchmark is publicly accessible to the research community.

\item \textit{``How will the dataset be distributed (\textit{e.g.}, tarball on website, API, GitHub)?''} 

\textcolor{red}{\textbf{A:}} The benchmark is hosted at \url{https://huggingface.co/datasets/thislzm/VLRS-Bench}; the project page is \url{https://github.com/MiliLab/VLRS-Bench}.

\item \textit{``When will the dataset be distributed?''} 

\textcolor{red}{\textbf{A:}} The dataset is publicly available.

\item \textit{``Will the dataset be distributed under a copyright or other intellectual property (IP) license, and/or under applicable terms of use (ToU)?''} 

\textcolor{red}{\textbf{A:}} Yes. The benchmark annotations are available under Creative Commons Attribution 4.0 International (CC BY 4.0). Source imagery remains subject to its original terms.

\item \textit{``Have any third parties imposed IP-based or other restrictions on the data associated with the instances?''} 

\textcolor{red}{\textbf{A:}} No additional third-party restrictions apply beyond the standard open-source and academic research licenses of the 10 public source datasets (summarized in Table~\ref{tab:dataset_summary}), which continue to govern the source imagery.

\item \textit{``Do any export controls or other regulatory restrictions apply to the dataset or to individual instances?''} 

\textcolor{red}{\textbf{A:}} No.    
\end{enumerate}

\subsection{Maintenance}
As with the questions in the previous section, dataset creators should provide answers to these questions prior to distributing the dataset. The questions in this section are intended to encourage dataset creators to plan for dataset maintenance and communicate this plan to dataset consumers.
\begin{enumerate}
\item \textit{``Who will be supporting/hosting/maintaining the dataset?''}

\textcolor{red}{\textbf{A:}} The authors of this work serve to support, host, and maintain the datasets.

\item \textit{``How can the owner/curator/manager of the dataset be contacted (\textit{e.g.}, email address)?''} 

\textcolor{red}{\textbf{A:}} The curators can be contacted via the email addresses listed on our paper or webpage.

\item \textit{``Is there an erratum?''} 

\textcolor{red}{\textbf{A:}} There is no separate erratum.

\item \textit{``Will the dataset be updated (\textit{e.g.}, to correct labeling errors, add new instances, delete instances)?''} 

\textcolor{red}{\textbf{A:}} Future updates (if any) will be posted on the dataset website.

\item \textit{``Will older versions of the dataset continue to be supported/hosted/maintained?''} 

\textcolor{red}{\textbf{A:}} 
Yes. All released versions of the dataset will remain hosted and accessible through the Hugging Face and GitHub repositories to ensure long-term reproducibility.

\item \textit{``If others want to extend/augment/build on/contribute to the dataset, is there a mechanism for them to do so?''} 

\textcolor{red}{\textbf{A:}} Yes, we will provide detailed instructions for future extensions.
\end{enumerate}

\section{Limitation and Potential Societal Impact}
\label{app-limitation}
In this section, we discuss the limitations and potential societal impact of this work.

\subsection{Potential Limitations}

While \textbf{VLRS-Bench} provides a comprehensive benchmark for evaluating the reasoning capabilities of MLLMs in remote sensing, there are several limitations to consider:
\begin{itemize}
\item \textbf{Scope of Sensors:} Although our benchmark integrates RGB, Near-Infrared (NIR), and DSM data from 10 sources in question construction, it may not cover all specialized sensor types such as Synthetic Aperture Radar (SAR) or Hyperspectral imaging, potentially limiting generalizability in extreme conditions.
\item \textbf{Model and Dataset Diversity:} 
In this paper, we evaluated 23 general-purpose and RS-specific MLLMs (e.g., GPT-5.4, Qwen3-VL, and ScoreRS). As new models emerge, their evaluation results will be added to our leaderboard. Additionally, VLRS-Bench will be expanded to include more fine-grained reasoning tasks in future iterations.
\item \textbf{Multilingual Support:} VLRS-Bench currently supports English, which is the dominant language in current MLLM research. In the future, we aim to extend support to other languages to serve a broader global research community.
\end{itemize}

\subsection{Potential Negative Societal Impact}
\begin{itemize}
\item \textbf{Safety Risks:} VLRS-Bench is designed to evaluate complex reasoning, including prediction and decision-making in scenarios like disaster relief. However, excessive reliance on model outputs without human oversight could lead to risks in high-stakes environments. It is crucial to implement human-in-the-loop supervision when deploying these MLLMs for real-world decision support.
\item \textbf{Environmental Impact:} 
Training MLLMs and conducting extensive evaluations on VLRS-Bench require significant computational resources. To mitigate this, we provide a public leaderboard and detailed model analysis, reducing the need for researchers to perform redundant evaluations.
\item \textbf{Bias and Fairness:} 
VLRS-Bench relies on 10 public datasets which may have inherent geographic or selection biases (e.g., favoring urban areas over rural ones). While we strove for diversity, models trained or evaluated solely on this benchmark might exhibit performance disparities across different regions. We aim to continuously expand the dataset diversity to minimize such biases.
\end{itemize}


\end{document}

%% file: funding_disclosure.tex
We extend our special thanks to Di Wang (\href{mailto:d_wang@whu.edu.cn}{\nolinkurl{d_wang@whu.edu.cn}}) for his contributions to the conceptual design of this work.

This work was supported in part by the Fundamental and Interdisciplinary Disciplines Breakthrough Plan of the Ministry of Education of China (JYB2025XDXM101); the New Generation Artificial Intelligence--National Science and Technology Major Project (No.~2025ZD0123602); the National Natural Science Foundation of China (Nos.~624B2109, 62225113, 623B2079, and 62601003); the New Cornerstone Science Foundation through the XPLORER PRIZE; the Innovative Research Group Project of Hubei Province (Grant No.~2024AFA017); the Postdoctoral Fellowship Program of CPSF (Grant No.~GZB20260403); the Key Technology Research Project of China National Petroleum Corporation (No.~2025ZG82); and the Zhongguancun Academy Project (20240308).

\paragraph{Competing interests.}
The authors declare no competing interests.